\documentclass[twocolumn]{article}

\usepackage{arxiv_two_col}

\usepackage[utf8]{inputenc}
\usepackage[T1]{fontenc}
\usepackage{hyperref}
\usepackage{booktabs}
\usepackage{amsfonts}
\usepackage{nicefrac}
\usepackage{microtype}
\usepackage{lipsum}
\usepackage{graphicx}
\usepackage[numbers]{natbib}
\usepackage{doi}

\usepackage{wrapfig}
\usepackage{subcaption}
\usepackage{bbm}
\usepackage{amsmath}
\usepackage{xurl}
\usepackage{pgfplots}
\pgfplotsset{compat=newest}

\graphicspath{ {graphics/} }

\usepackage{mathtools}
\usepackage[most]{tcolorbox}
\usepackage{multirow}
\usepackage{amssymb}

\usepackage{newfloat}
\tcbuselibrary{skins,breakable}

\DeclareFloatingEnvironment[
  fileext=lob,
  listname={List of Boxes},
  name=Box,
  placement=tbp,
  within=none,
]{textbox}

\newtcolorbox{promptbox}[2][]{
  enhanced,
  colback=gray!8,
  colframe=gray!60,
  boxrule=0.4pt,
  arc=2mm,
  left=3mm,right=3mm,top=2mm,bottom=2mm,
  fonttitle=\bfseries\sffamily,
  fontupper=\sffamily\fontsize{7.8pt}{10pt}\selectfont,
  title=#2,
  #1
}

\newcommand{\ie}{\textit{i}.\textit{e}.}
\newcommand{\eg}{\textit{e.g.}}

\newcommand{\ul}[1]{\underline{\text{#1}}}

\title{Why Do LLMs Struggle in Strategic Play?\\Broken Links Between Observations, Beliefs, and Actions}

\date{}

\author{
    Jan Sobotka\footnotemark[1]\;\,\footnotemark[2]\\
    EPFL\\
    \And
    Mustafa O. Karabag\\
    The University of Texas at Austin\\
    \And
    Ufuk Topcu\\
    The University of Texas at Austin\\
}

\renewcommand{\headeright}{}
\renewcommand{\undertitle}{}
\renewcommand{\shorttitle}{Why Do LLMs Struggle in Strategic Play? Broken Links Between Observations, Beliefs, and Actions}

\hypersetup{
    pdftitle={Why Do LLMs Struggle in Strategic Play? Broken Links Between Observations, Beliefs, and Actions},
    pdfauthor={Jan Sobotka, Mustafa O. Karabag, Ufuk Topcu},
    pdfkeywords={Large language models, Strategic decision-making, Artificial intelligence, Game theory},
}

\begin{document}
\twocolumn[\maketitle]
\renewcommand{\thefootnote}{\fnsymbol{footnote}}
\footnotetext[1]{Corresponding author: \texttt{jan.sobotka@epfl.ch}}
\footnotetext[2]{This work was completed while J.S. was with The University of Texas at Austin.}

\begin{abstract}
Large language models (LLMs) are increasingly tasked with strategic decision-making under incomplete information, such as in negotiation and policymaking. While LLMs can excel at many such tasks, they also fail in ways that are poorly understood. We shed light on these failures by uncovering two fundamental gaps in the internal mechanisms underlying the decision-making of LLMs in incomplete-information games, supported by experiments with open-weight models Llama 3.1, Qwen3, and gpt-oss. First, an observation-belief gap: LLMs' internal representations of latent game states are substantially more accurate than their own verbal reports. However, these representations, which we call internal beliefs following game-theoretic terminology, are brittle. In particular, the belief accuracy degrades with multi-hop reasoning, exhibits primacy and recency biases, and drifts away from Bayesian coherence over extended interactions. Second, a belief-action gap: The implicit conversion of internal beliefs into actions is weaker than that of the beliefs externalized in the prompt, yet neither belief-conditioning consistently achieves higher game payoffs. Moreover, acting optimally on the decoded beliefs would improve payoffs in about 95\% of games, pointing to a bottleneck in the belief-to-action conversion. These results show how analyzing LLMs' internal processes can expose systematic vulnerabilities that warrant caution before deploying LLMs in strategic domains without robust guardrails.
\end{abstract}

\section{Introduction}
\label{sec:introduction}
Large language models (LLMs) are increasingly used for decision-making in strategic domains requiring reasoning under uncertainty, such as negotiation~\cite{bianchi2024how,abdelnabi2024negotiation,priya2025argue,kwon2025astra,kwon2025evaluating}, coordination~\cite{abdelnabi2024negotiation,su2026endrewardengineeringllms}, and policymaking~\cite{coz2025what,pan2025urban,ziegler2025,chen2025}. These settings require agents to construct accurate beliefs about latent variables that are not directly observable, such as opponents' strategies, private cards, or hidden roles. While the general capabilities of LLMs have improved rapidly, our understanding of the decision-making processes they employ in these strategic contexts --- and where those processes diverge from rational behavior --- has not kept pace~\cite{abdelnabi2024negotiation,scott2024,goktas2025strategic,kempinski2025got,lore2024}. As a result, it remains difficult to accurately predict, debug, or improve the reliability and safety of LLMs in strategic settings~\cite{song2026large}.

\begin{figure}[t]
    \centering
    \includegraphics[width=\linewidth]{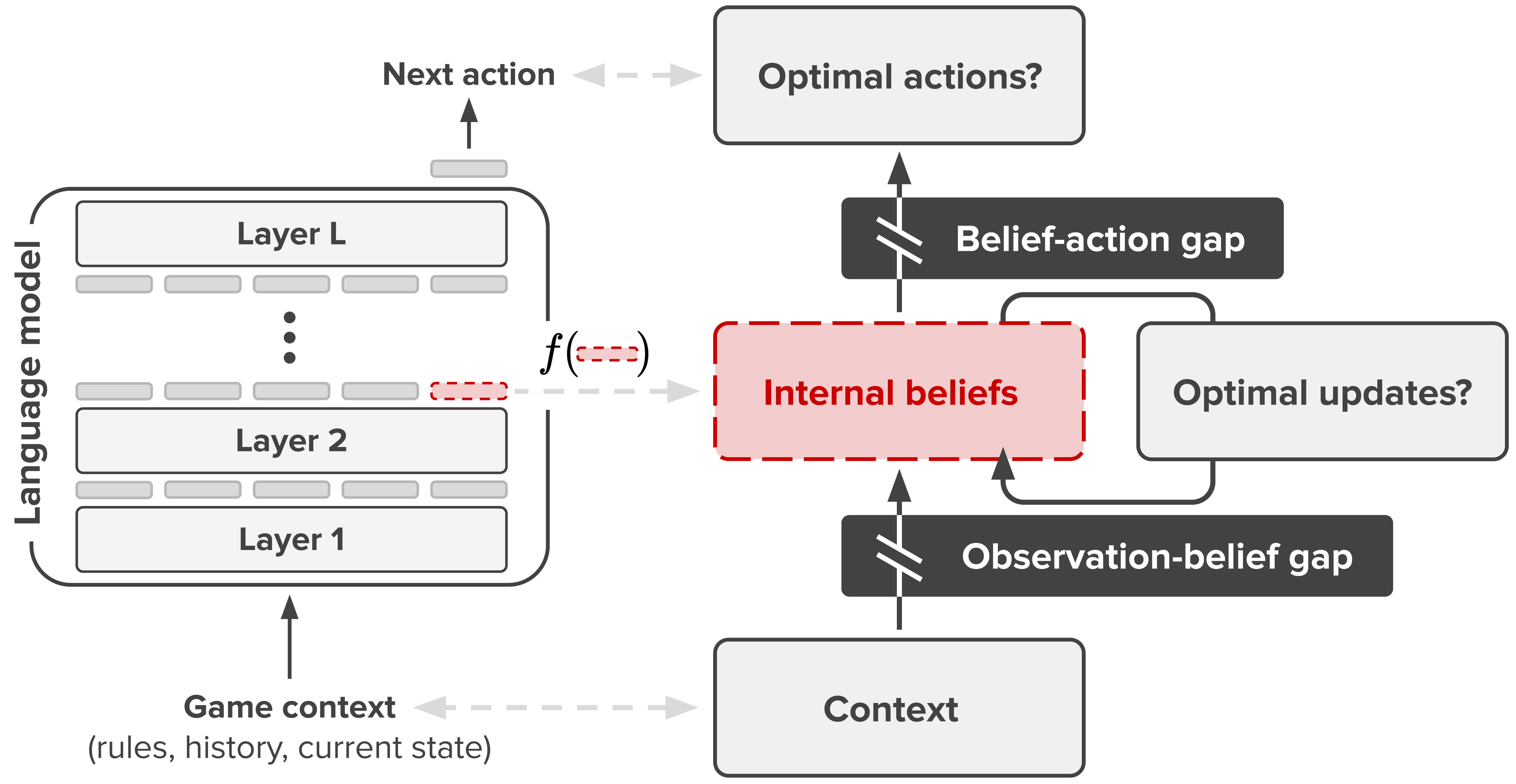}
    \caption{Aspects of strategic gameplay studied in this manuscript. $f$ refers to an \textit{internal probe} which decodes the contents of the LLM's internal representations.}
    \label{fig:overview}
\end{figure}

We employ game theory and tools from mechanistic interpretability to identify and understand the strengths and weaknesses of LLM agents' strategic decision-making. While game theory provides formal models of optimal reasoning under uncertainty, mechanistic interpretability methods enable inspection of the internal states and mechanisms that support such reasoning in these models. This joint approach allows a structured analysis of how LLMs form, update, and act on their representations of strategic contexts, which we call (internal) beliefs following game-theoretic terminology. In particular, probing and activation steering techniques from mechanistic interpretability allow us to decode LLM agents' \textit{internal representations} into game-theoretic concepts and to manipulate these representations through causal interventions (\autoref{fig:overview}) to achieve specific game outcomes. In modern transformer-based LLMs, these internal representations, sometimes called \textit{hidden states}, refer to continuous-valued vectors produced by intermediate layers of LLMs, which provide a window into their internal processing of inputs. 

Existing literature proposes various methods for defining and extracting LLM beliefs~\cite{herrmann2024standards}. In game-theoretic settings, however, beliefs are well-defined: they refer to distributions over latent game variables (\eg, opponent's strategy type) that are sufficient for optimal decision-making~\cite{harsanyi1967games,kreps1982,astrom1965,sondik1978}. We therefore extract an agent's internal beliefs using linear probes~\cite{belinkov2022probing,gurneefinding} trained to predict these ground-truth latent game variables. This approach yields estimators that maximize predictive accuracy with respect to the ground-truth variables, which amounts to interpreting internal beliefs as the most accurate information linearly decodable at the point of action selection. A question then arises whether the model utilizes these representations optimally when selecting actions. We remark that these beliefs need not coincide with beliefs in the sense used in philosophy of mind and cognitive science~\cite{schwitzgebel2024belief,dennett1989intentional}, which is neither unanimously defined nor tied to an externally specified ground truth. We further discuss the interpretation of beliefs in \autoref{app:internal_belief_interpretation} in the Appendix.

Our focus on internal belief processes contrasts with prior work~\cite{bianchi2024how,abdelnabi2024negotiation,kwon2025evaluating,coz2025what,scott2024,lore2024,akata2025playing,guertler2025textarena}, which analyzed the external behavior (final outputs) of LLM agents, leaving the decision-making mechanisms underlying their (sub)optimal performance largely unexplored. The joint analysis of internal representations and external behavior allows us to reveal two fundamental gaps in LLMs' internal mechanisms.

First, we identify an \textbf{observation-belief gap}, defined as the discrepancy between a ground-truth latent game variable and the belief recoverable by the agent. While LLMs form internal beliefs that are consistently more accurate reflections of the ground truth compared to their verbalized counterparts, this process remains a bottleneck. Specifically, belief formation is brittle: its accuracy declines as the compositional complexity increases, and the internal representations of interaction history exhibit a bias that prioritizes the first and most recent observations, mirroring primacy and recency biases in humans~\cite{murdock1962serial,yoo2025kq}. Furthermore, we demonstrate that the internal consistency of these beliefs deteriorates over time, as the models' belief updates drift away from Bayesian optimality in repeated interaction.

Second, we discover a \textbf{belief-action gap}, in which the model fails to reliably translate internal and verbal beliefs into optimal strategic choices. In particular, LLMs (1) frequently fail to select payoff-maximizing actions even when such actions are implied by the beliefs they represent or report, (2) exhibit consistent action-letter biases which partially explain their suboptimality, (3) respond only weakly in their final actions when their hidden states are steered toward specific beliefs, and (4) do not play consistently better even when explicitly prompted with these internal beliefs.

We verify these gaps across three families of open-weight LLMs (Llama-3.1, Qwen3, gpt-oss) and three incomplete-information games (repeated normal-form games, Generalized Kuhn Poker~\cite{ganzfried2019mistakes}, and The Chameleon~\cite{chameleon}).

These findings carry practical implications for deploying LLM agents in strategic settings. Because internal beliefs can be substantially more accurate than verbalized ones, evaluations that rely on stated probabilities or explanations may be misleadingly optimistic or pessimistic about an agent's true representational competence. Similarly, as LLM beliefs lose coherence over extended interactions (despite the short prompt context length), short-horizon rollouts can hide late-stage failures. More broadly, practitioners should anticipate systematic vulnerabilities (primacy/recency anchoring, under-reaction over time, exploitability via mid-trajectory deviations) and mitigate them with design choices that make beliefs explicit and coherent (\eg, adding internal beliefs in context, enforcing explicit reasoning steps, performing continual belief consistency checks), rather than assuming that prompting alone will elicit robust strategic behavior.

\section{Experimental Setup}
\label{sec:exp_setup}
\begin{figure*}[t]
    \centering
    \includegraphics[width=\linewidth]{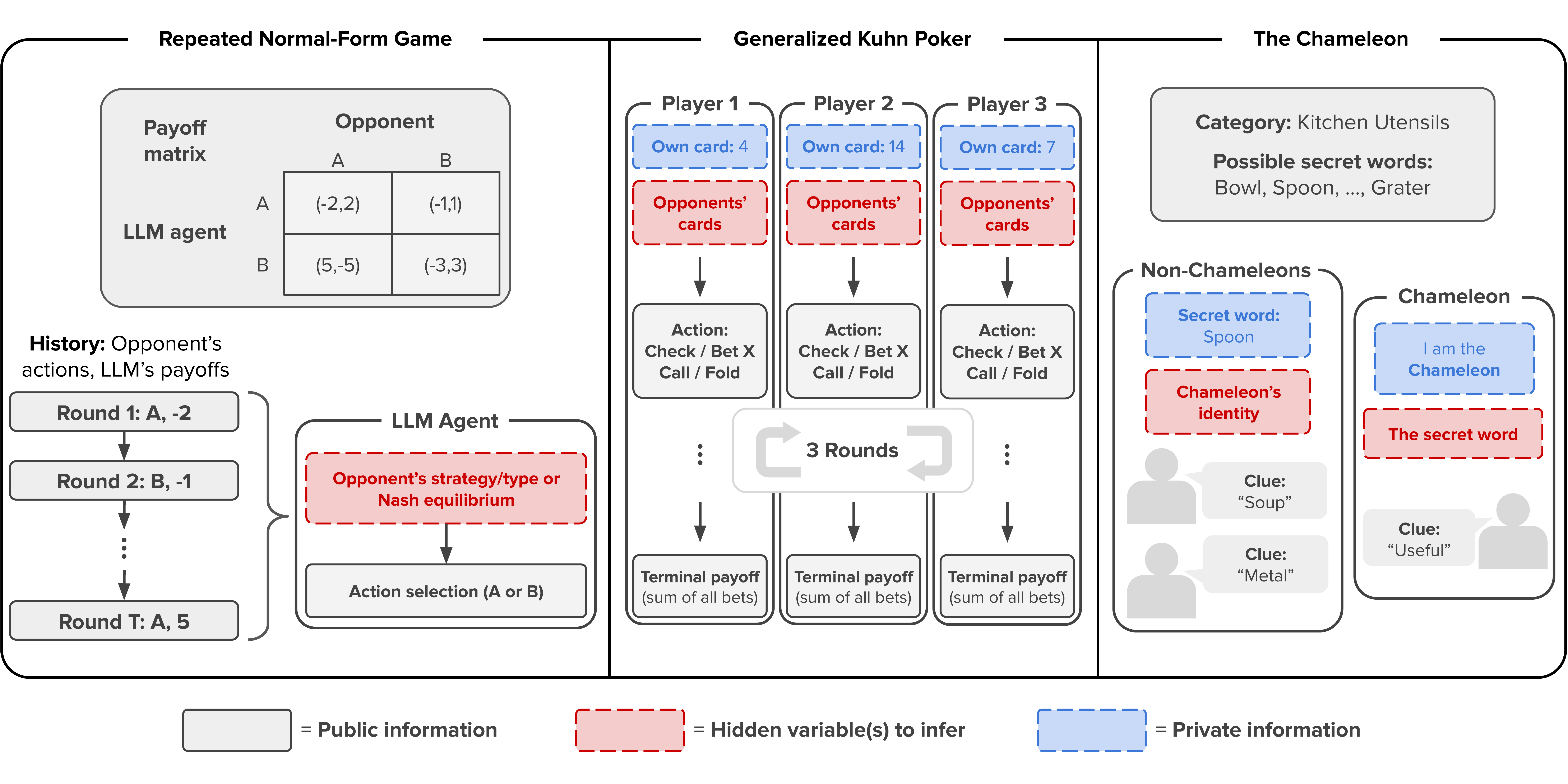}
    \caption{Schematic overview of the games used in this study. In Generalized Kuhn Poker and The Chameleon, separate LLM agents take the role of the players.}
    \label{fig:games}
\end{figure*}

\subsection{Models}
We use open-weight, instruction-tuned LLMs Llama 3.1 70B~\cite{llama3}, Qwen3 32B~\cite{qwen3}, and gpt-oss 20B~\cite{agarwal2025gpt}. The selection is made to include the largest model variants below 100B across widely used model families. For computational and memory efficiency reasons, we use quantized variants of models with more than 30B parameters. In \autoref{app:additional_experimental_results} in the Appendix, we show that the non-quantized and base (\ie, not instruction-tuned) counterparts exhibit qualitatively similar phenomena, which are in most cases also quantitatively close to the models studied in the main text. The focus on open-weight models is motivated by the fact that many of our experiments rely on access to the LLMs' internal representations, which APIs of closed-source models such as GPT-5~\cite{openai_gpt5} do not provide. Nevertheless, where hidden states are not required, we evaluate frontier API-only LLMs, including reasoning models, in \autoref{app:additional_experimental_results} in the Appendix.

\subsection{Games}
We study three classes of games that involve distinct forms of strategic inference under incomplete information: repeated normal-form games (matrix representation of common finite-action scenarios), Generalized Kuhn Poker~\cite{ganzfried2019mistakes,ganzfried2022} (an imperfect-information card game), and The Chameleon~\cite{chameleon} (a hidden-identity, social deduction game). These games span increasing levels of inferential complexity while retaining clear objectives, precise control over latent variables, and, in some cases, computable optimal strategies. An overview of the rules and the information flow in these games is shown in \autoref{fig:games} and described in detail below.

\textit{Repeated normal-form games.}
In repeated normal-form games, an LLM agent plays a two-player matrix game against a fixed opponent for $T$ rounds. We consider $2\times2$ normal-form games: In each round, both players simultaneously select an action from the set $\{A,B\}$ and receive payoffs determined by a payoff matrix. 

To analyze belief formation in LLMs in a sufficiently complex setting, the payoff matrix and the complete action history are observable to both players, but the opponent's underlying memoryless fixed strategy (probability distribution over actions) or type (stochastic policy class or payoff matrix) is not directly observed. The LLM must therefore infer these latent opponent properties from past actions and use them to guide future decisions. The hidden variables of interest in this setting are the parameters of the opponent's strategy, the opponent's type, or the unique Nash equilibrium (as specified in individual experiments). Note that we use ``incomplete information'' here in the broad sense of uncertainty about the opponent's behavioral policy, rather than only in the standard Harsanyi sense of uncertainty about payoff-relevant types~\cite{harsanyi1967games}. The opponent-type experiments correspond more closely to the latter, in particular the three-hop setting, in which the candidate types differ by their payoff matrices. The Nash equilibrium is a further special case, since it is not a random variable from the players' perspective but a fixed game-theoretic quantity computable from the remaining game information.

\textit{Generalized Kuhn Poker.}
Similar to \cite{ganzfried2022}, we generalize the standard two-player, zero-sum, imperfect-information card game called Kuhn Poker~\cite{kuhn1950simplified} into its larger variant. While the original Kuhn Poker is restricted to two players, a deck of three cards, and a single betting round, our generalized version is a natural extension to $N \ge 2$ players, a deck of $D \ge 3$ numerically ranked cards, and $R \ge 1$ rounds of betting ($N=3, R=3, D=20$ in our experiments). At the start of each game trial, referred to as ``hand'' in poker, each player is dealt one private card from a known deck. Players then proceed through up to $R$ betting rounds, taking turns to act and observing the full public betting history but never the opponents' private cards. Available actions include checking, betting an allowed amount, folding in response to a bet, or calling a bet, subject to the player's individual stack constraints. At the end of the hand --- either after all betting rounds conclude or earlier if all but one player folds --- the player holding the highest private card among the remaining players wins the pot.

From the perspective of any single agent, the latent variables are the private cards of the other players. Optimal play, therefore, requires maintaining and updating a belief distribution over opponents' private cards conditioned on observed actions and selecting actions accordingly. Generalized Kuhn Poker thus serves as a suitable testbed for evaluating whether LLMs translate inferred beliefs about hidden private information into approximately optimal betting behavior.

\textit{The Chameleon.}
The Chameleon~\cite{chameleon} is an $N$-player hidden-identity game that requires agents to strategically reveal, conceal, and infer information~\cite{karabag2026do}. At the start of the game, a category and a finite set of possible secret words are publicly revealed. One word is uniformly randomly selected as the \emph{secret word} and privately disclosed to all non-chameleon players, while a single chameleon player does not observe it. Player identities (chameleon vs.\ non-chameleon) are private. Players then respond sequentially, each providing a single-word clue after observing all prior clues. After all clues are given, players simultaneously vote to identify the chameleon. If the chameleon is correctly identified, it receives a second chance to guess the secret word; the chameleon wins if it either avoids identification or guesses the secret word correctly, and otherwise the non-chameleons win.

For our analysis, the key latent game variables are the secret word and the chameleon's identity. Non-chameleon agents must generate clues that are informative enough to signal shared knowledge to other non-chameleons while remaining sufficiently ambiguous to conceal the secret word from the chameleon. Conversely, the chameleon must infer the secret word and blend in despite lacking privileged information. The Chameleon game, therefore, provides a natural language-based setting in which to study the formation of internal beliefs and their translation into action.

\paragraph{\textit{Information-flow perspective.}}
Across all three games, the core structure consists of publicly observable rules and action histories, agent-private information, and latent variables that are never directly observed but must be inferred. \autoref{fig:games} makes this structure explicit: grey denotes public information and rules, red hidden variables to be inferred, and blue an agent's private information. This shared information-flow structure enables a unified analysis of belief formation, belief updating, and belief-to-action conversion across qualitatively different strategic environments.

\begin{textbox}[t]
\centering
\begin{promptbox}{Prompt: Repeated normal-form games}
You are playing a game repeatedly with 1 other player, Player 1. There are 2 possible actions in each round (action A, action B). Players make their actions simultaneously, and the outcome is determined by their actions in the given round. All the players, including you, want to maximize their payoff (number of points). Players may adapt their strategies as they play more rounds.
\vspace{\the\baselineskip}

Here are the rules of the game:

\hspace{2ex} If you play action A and the other player plays action A, you

\hspace{2ex} get 4.2 points and they get 6.5 points.

\hspace{2ex} If you play action A and the other player plays action B, you

\hspace{2ex} get 4.4 points and they get 8.9 points.

\hspace{2ex} If you play action B and the other player plays action A, you

\hspace{2ex} get 9.6 points and they get 3.8 points.

\hspace{2ex} If you play action B and the other player plays action B, you

\hspace{2ex} get 7.9 points and they get 5.3 points.
  
\vspace{\the\baselineskip}

Here is the history of the game so far:

\hspace{2ex} Round 1: You played action B, and Player 1 played action B.

\hspace{2ex} Your payoff for this round was 7.9 points.

\hspace{2ex} Round 2: You played action A, and Player 1 played action A.

\hspace{2ex} Your payoff for this round was 4.2 points.

\hspace{2ex} \dots

\hspace{2ex} Round 8: You played action A, and Player 1 played action A.

\hspace{2ex} Your payoff for this round was 4.2 points.

\vspace{\the\baselineskip}

Given the history and the rules of the game, please provide your action for the next round. Respond with a single line that contains only the letter of your chosen action, do not say anything else.
\end{promptbox}
\caption{The prompt used for asking LLMs for their next action in the repeated normal-form games. Rounds 3 to 7 are omitted for presentation purposes.}
\label{fig:matrix_prompt}
\end{textbox}

\subsection{Probing}

Strategic decision-making under incomplete information requires agents to construct accurate beliefs about latent variables that are not directly observable, such as opponents' strategies, private cards, or hidden roles. To evaluate whether LLMs can extract such latent game information, we adopt a mechanistic interpretability approach based on probing~\cite{belinkov2022probing,gurneefinding,voita2020information,elazar2021amnesicprobing}. Specifically, we train simple linear probes (logistic or linear regression models) to predict ground-truth latent variables from the LLM's internal representations, while keeping the base LLM's parameters frozen. We use these \textit{internal probes} merely as exogenous sensors to analyze the internal representations --- their predictions are not provided to the base LLM unless stated otherwise (\autoref{subsec:bag_analysis}). In parallel, we evaluate \textit{verbal probes}, where the LLM is explicitly asked in natural language to infer the same latent variables from the identical context. The ability of such probes to reconstruct unobserved environment variables tells us about the representational and inferential capacity of current LLMs. 

We deliberately restrict our internal probes to linear models, as the goal is not to improve the inference capabilities of the base LLM, but rather to analyze how the pre-existing systems operate~\cite{belinkov2022probing,voita2020information,elazar2021amnesicprobing}. Intuitively, such probes search the LLM's hidden state space to identify a linear direction that best represents the latent environment variable, thereby uncovering the LLM's inference capabilities. We refer to these inferred variables as \textit{internal beliefs}.

In each game, we prompt the LLMs with the publicly available, full interaction history and ask them to select their next action. An example prompt for the repeated normal-form games is given in Box~\ref{fig:matrix_prompt} (other prompts in the Appendix). We generate such independent game trajectories for all games, each with randomly initialized hidden variables (\eg, opponent strategies, uniformly sampled payoff matrices, private cards, etc.). Internal probes are trained on hidden states extracted from the intermediate layers of the LLM while it is processing the final token of the prompt prior to action generation. The choice of middle layers is motivated by prior work, which found that these layers contain most of the high-level semantic information~\cite{skean2025layer,geva2021transformer,nostalgebraist2020interpreting}. We use the hidden states from the last token position because in the decoder-only Transformer architecture~\cite{vaswani2017attention,radford2018improving}, information relevant for next-token prediction (action selection) must be represented in this final state to causally influence the next-token probabilities. We split the data into disjoint subsets for the probe's training, hyperparameter selection, and held-out evaluation, ensuring that the reported probe performance reflects genuine generalization rather than memorization. Further details on gameplay and probes are provided in the Appendix, including an extended discussion of our game-theoretic definition of beliefs, its relation to prior usage of the term, and what it implies for our claims about the belief-action gap.

\section{Results}
\label{sec:results}
\begin{figure*}[t]
    \centering
    \includegraphics[width=0.95\linewidth]{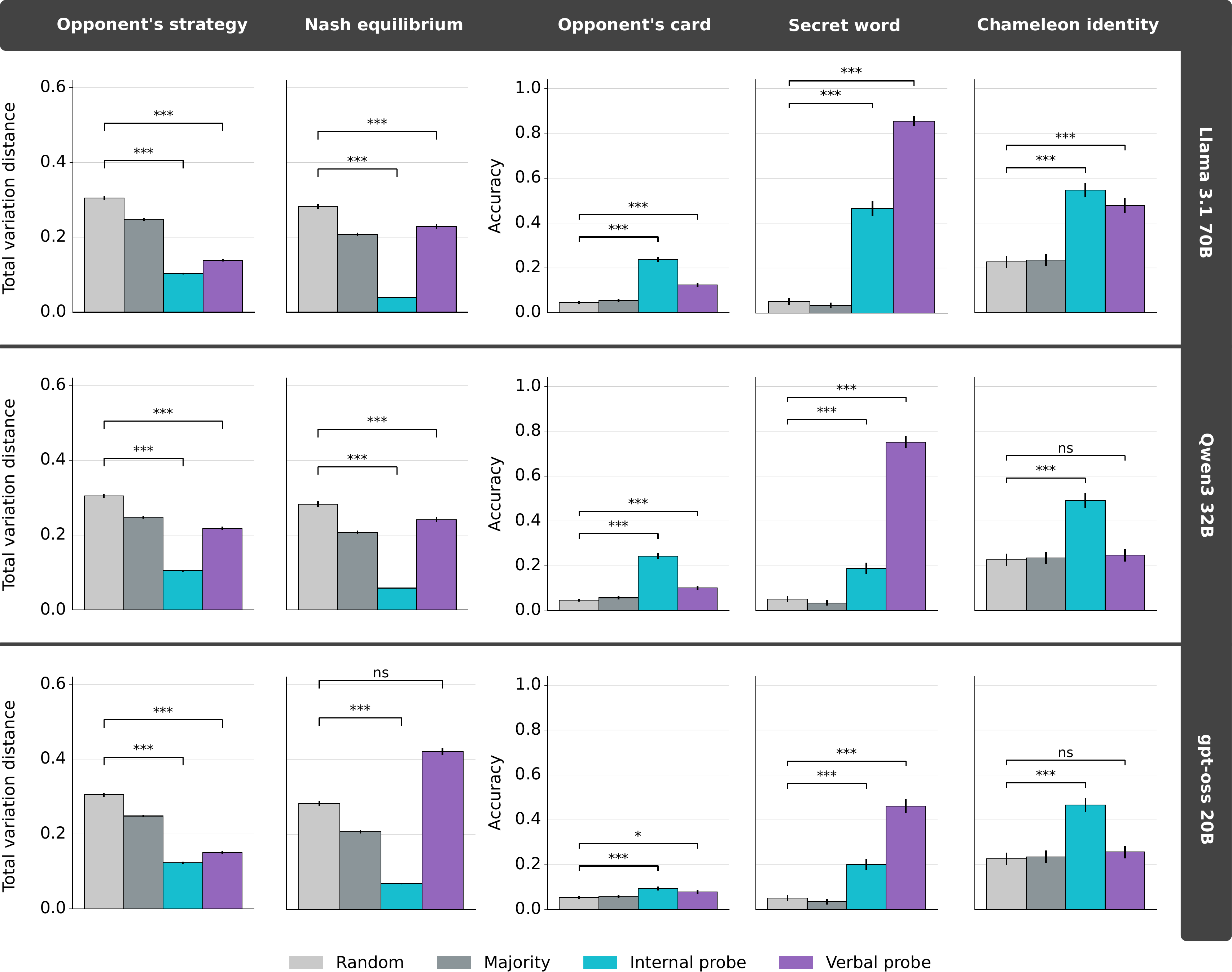}
    \caption{Inference of latent information in the selected strategy games. Shown are the means and their standard errors obtained from 800 (repeated normal-form games), 400 (Generalized Kuhn Poker), and 230 (The Chameleon) game trials. Asterisks denote statistical significance (one-sided Wilcoxon signed-rank test for distribution-based targets - two leftmost plots, McNemar's test for classification tasks - three rightmost plots; * $p<0.05$; ** $p<0.01$; *** $p<0.001$). Total variation distance: lower is better; Accuracy: higher is better. Plots from left to right: \textbf{(i)} \textit{Repeated normal-form games:} Opponent's fixed strategy (probability distribution over actions). \textbf{(ii)} \textit{Repeated normal-form games:} Mixed strategies Nash equilibrium (probability distribution over actions for both players; zero-sum payoff structure; $T = 0$). \textbf{(iii)} \textit{Generalized Kuhn Poker:} Secret card of the opponent. \textbf{(iv)} \textit{The Chameleon:} Secret word. \textbf{(v)} \textit{The Chameleon:} Chameleon identity.}
    \label{fig:hidden_info}
\end{figure*}

\subsection{Belief Formation: Inferring Latent Game Information}\label{subsec:belief_formation}
We examine how successful LLMs are in inferring latent game information given a history of actions. We perform this analysis for all three games and provide experimental details in \autoref{app:experimental_design} (Appendix). We compare internal and verbal probes against two baselines: a \textit{random} baseline that samples predictions from a uniform distribution over latent variables, and a \textit{majority} baseline that always predicts the most frequent latent value (or the average for distribution-based variables) in the internal probe's training data.
 
\paragraph{\textit{Internal beliefs are more accurate than verbal beliefs.}}
\autoref{fig:hidden_info} summarizes the performance of the probes across all games and latent variables. In all settings, internal probes substantially outperform both baselines, achieving significantly lower total variation distance when predicting distributions (two plots on the left) and higher accuracy when predicting discrete hidden variables (three plots on the right). In contrast, verbal probes perform markedly worse, sometimes only marginally improving over the majority baseline and, in 4 out of 5 cases, underperforming internal probes.

The only qualitatively opposite result occurs when inferring the secret word in the Chameleon, where internal probes underperform verbal ones. The explanation behind this reversal is the way the internal probes are constructed in this scenario. Specifically, for secret-word probing, we cannot use standard multinomial logistic regression linear probes because the secret words are not ordered, unlike targets such as possible opponents' private cards in the Generalized Kuhn Poker. Predicting the secret words directly would require substantially more expressive probing techniques, increasing model complexity and hindering interpretability. In light of these limitations, we use projections of the LLM's internal representations onto the secret-word embeddings of the same LLM. Normalizing the projected scores of each potential secret word gives us a measure of alignment between the current internal belief and the set of possible inferred beliefs~\cite{park2024the,elhage2022superposition,meng2022locating}.

To verify that the probes read out inference performed by the model rather than performing it themselves, we compare them against a predictor of the same linear class trained directly on the raw observable game information stated in the prompt (Appendix, \autoref{app:additional_experimental_results}). This baseline approaches the internal probe only for the opponent's strategy, which is close to an empirical action frequency, and remains near chance for the Nash equilibrium and the opponent's type, both of which are nonlinear functions of the observations. The hidden states therefore encode a transformed estimate of the latent variables rather than a trivial re-encoding of the raw observations.

These findings also hold under a realistic framing of the games, in which the repeated normal-form games are presented as a financial decision-making scenario~\cite{lore2024} and The Chameleon as a shared-channel scenario with a corporate spy. Additionally, the results are stable under changes to the surface form of the prompt, including action-label and prompt-order variations (Appendix, \autoref{app:additional_experimental_results}).

The results demonstrate that LLMs form accurate implicit beliefs about latent game information that are not faithfully reflected in their explicit verbalizations. Crucially, the strong performance of linear probes indicates that this information is represented in a linearly accessible form at the point of LLMs' action selection, implying that belief inference itself might not be the primary source of LLM suboptimality in these strategic settings. Instead, these findings motivate our later analysis of how (and to what extent) these relatively accurate internal beliefs causally translate into optimal action choices. Additionally, the dissociation between internal and verbal beliefs has practical consequences for evaluation and alignment. Benchmarks that rely on eliciting explicit probability judgments or explanations may substantially underestimate the representational competence of LLM agents, as accurate beliefs can be present but not faithfully verbalized. Improving reliability in strategic domains may therefore require either methods that expose internal beliefs or training objectives that more tightly couple internal representations with external reports.

\subsection{Belief Formation: Effects of Side Information and Intermediate Reasoning Hops}

We next examine how belief accuracy changes with auxiliary information and an increasing number of required intermediate reasoning hops. We focus on repeated normal-form games in which the prompt specifies that the opponent belongs to one of two possible types. A \textit{reasoning hop} is defined as a distinct latent computation that must be correctly performed and composed in order to identify the opponent's type from the game history.

We consider three settings with increasing hop counts. In the \textit{one-hop} setting, opponent types differ only by their stationary mixed strategies, which we explicitly specify in the prompt. Correct inference requires estimating the opponent's empirical action distribution from the history and matching it to the closer of the two candidate strategies. In the \textit{two-hop} setting, each opponent type specifies distinct strategies for each of two round types (\textit{blue} and \textit{red}). Here, the LLM is explicitly told the round type at each timestep and must estimate a separate opponent's strategy for each round type from the action history, before matching the resulting pair of strategies to the correct opponent type. In the \textit{three-hop} setting, opponent types differ by their payoff matrices. In this case, the LLM must first infer the mixed-strategy Nash equilibrium implied by each payoff matrix, then estimate the opponent's empirical strategy from observed actions, and finally map this estimate to the closer equilibrium strategy. In the one- and two-hop settings, we withhold the payoff matrices so that the models must use the strategies to infer the opponent's type (prompts in the Appendix). We note that these three conditions change the form of the task --- from matching a stationary strategy to estimating context-conditional strategies, to deriving and comparing equilibrium behavior from payoff matrices --- each requiring an additional reasoning hop from the previous. The conditions therefore vary the difficulty of identifying the opponent's type, of which hop count is one component alongside the intrinsic structure of each task.

\begin{figure}[t]
    \centering
    \includegraphics[width=\linewidth]{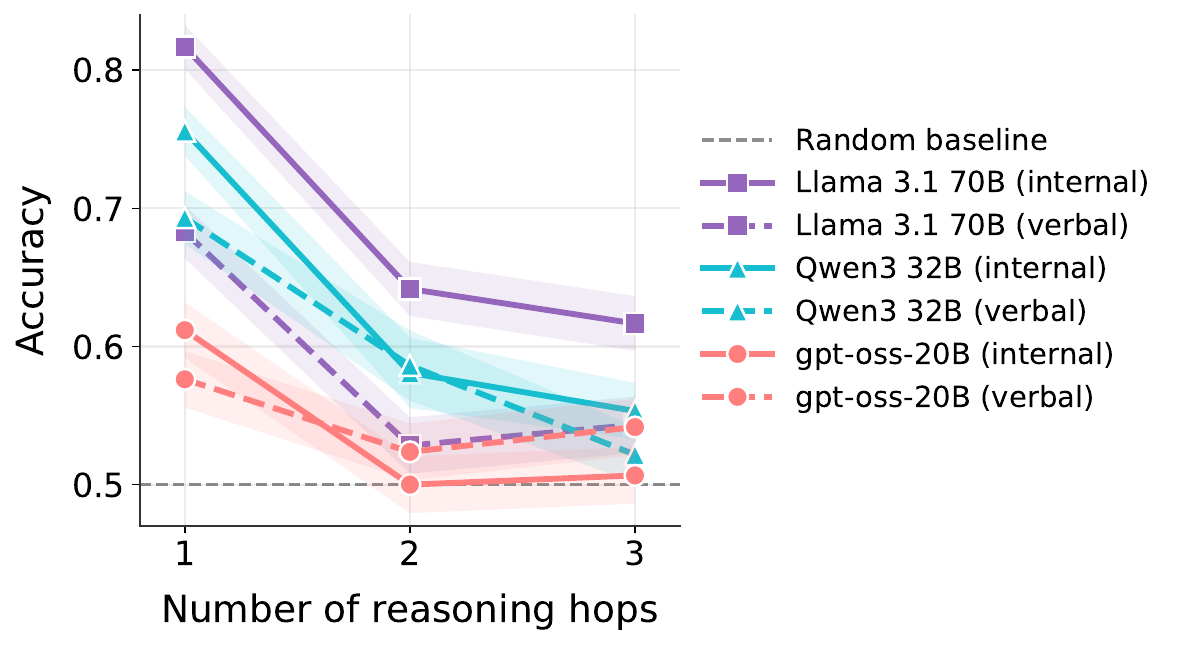}
    \caption{Inference of opponent's type in repeated normal-form games. Number of reasoning hops corresponds to: \textbf{1:} Opponent's type by strategy. \textbf{2:} Opponent's type by strategy and round type. \textbf{3:} Opponent's type by payoff matrix. Shown are the means and their standard errors obtained from 800 repeated normal-form games. \textit{Internal} and \textit{verbal} correspond to internal and verbal probes, respectively.}
    \label{fig:hidden_info_types}
\end{figure}

\begin{figure*}[h!]
    \centering
    \includegraphics[width=\linewidth]{pca_r15_annot.pdf}
    \caption{PCA of the internal representations of Llama 3.1 70B in round 15 of repeated normal-form games. Shown are the projections onto the first three principal components (PC) and the associated variances explained by these PCs. Both the left and right panels depict identical down-projected internal representations, but with color labeling by different game information. Data collected from 1,000 game trials.}
    \label{fig:pca_past_actions}
\end{figure*}

\paragraph{\textit{Belief formation is successful for each individual reasoning hop, while unsuccessful for combined hops.}} As shown in \autoref{fig:hidden_info_types}, belief accuracy declines with hop count. In the one-hop setting, probes with the larger models achieve accuracies above 0.75, significantly outperforming the \textit{random} baseline. In the two-hop setting, accuracy drops to below 0.65 and then either continues to degrade or plateaus in the three-hop setting, sometimes reaching near-random levels.

Together, these results demonstrate that LLMs can accurately represent latent aspects of the game in isolation, but struggle when these representations must be composed across multiple heterogeneous reasoning steps: Belief formation in terms of linear decodability and explicit verbal identification becomes increasingly brittle as the inferential difficulty of composing these steps grows. In particular, \autoref{fig:hidden_info} showed that both opponent strategy and the mixed-strategy Nash equilibrium are individually inferable, while their composition fails under multi-hop inference (\autoref{fig:hidden_info_types}).

As additional experiments in \autoref{app:additional_experimental_results} show, this degradation also holds when the side information about the opponent's type is placed before rather than after the action history, and under the alternative action labels and the realistic framing noted earlier.

In real-world settings such as negotiation or policymaking, correct behavior often depends on chaining multiple conditional inferences. Our results imply that the belief formation in LLM agents may perform reliably when reasoning steps are isolated, yet fail when these steps must be integrated. The degradation in inferring latent information motivates the paradigm of verbal Chain-of-Thought reasoning~\cite{wei2022chain,wang2024chainofthought,besta2025reasoning} or methods for strengthening compositional internal inference~\cite{saunshi2025reasoning,hao2025training,chen2025reasoning}.

\paragraph{\textit{Action optimality does not track belief accuracy across hop conditions.}}
We also measure the rate at which the LLM selects the best response to the true opponent type in each condition, with the payoff matrices stated in the prompt throughout so that a best response is well-defined. The best-response rate given in \autoref{app:additional_experimental_results} declines from one to two hops, tracking the decline in decodability, but then rises sharply at three hops for the larger models, above their one-hop values and at the point where decodability is lowest.

While counterintuitive, this rise may be attributable to the fact that the three conditions differ in how much policy-relevant information the prompt contains: from two candidate opponent strategies in the one-hop setting to four in the two-hop setting to none in the three-hop setting. The ordering of action optimality thus runs opposite to the amount of strategic side information supplied, being highest in the three-hop condition and lowest in the two-hop condition. One potential explanation is that in the three-hop condition the models' actions are driven directly by the observed action history and the stated payoff matrices, without an intermediate representation of the opponent's type, which would account for action optimality being highest where type decodability is lowest. This suggests that providing more policy-relevant side information need not improve the strategic decision-making of LLMs, and that representations and behavior are worth evaluating separately.

\subsection{Attention to Opponent's Past Actions}

\paragraph{\textit{Interaction history dominates representational variance relative to side information.}}
In \autoref{fig:pca_past_actions}, we examine how information about (i) specific past opponent actions and (ii) the opponent's type is organized in Llama 3.1's internal representations during repeated normal-form games. We perform principal component analysis (PCA) on the model's hidden states at the action-selection step and visualize the resulting down-projected states, labeled by the opponent's first-round action or the opponent's type. The two possible types of the opponent differ in their strategies, and the experimental setup corresponds to one reasoning hop in \autoref{fig:hidden_info_types}.

\autoref{fig:pca_past_actions} shows that even late in the game (round 15 is shown), variation associated with the opponent's round-1 action aligns with higher-variance principal components than variation associated with type labels. While PCA variance does not by itself establish causal influence on the final policy, this result indicates that the model's state space continues to allocate substantial representational capacity to early interaction evidence, relative to the (potentially decision-relevant) type information provided in the prompt.

\begin{figure}[t]
    \centering
    \includegraphics[width=\linewidth]{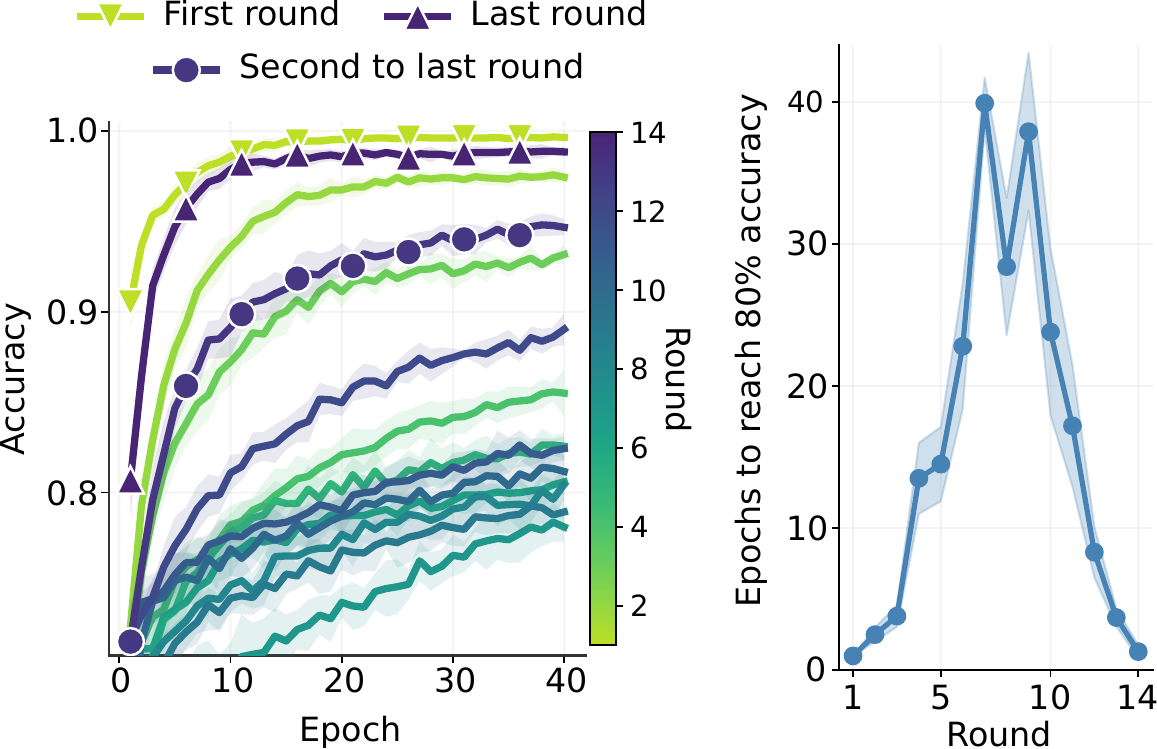}
    \vspace{-8pt}
    \caption{Probing past opponent's actions from internal representations of Llama 3.1 70B in round 15 of repeated normal-form games. Left: The validation accuracy of the internal probe over the probe's training epochs, colored by the round of the opponent's action that is being probed. Right: The number of the probe's training epochs (y-axis) after which the probe achieves 80\% validation accuracy in decoding the opponent's actions from previous rounds (x-axis). Each line represents the mean $\pm$ std from 10 training runs of the internal probe.}
    \label{fig:per_round_probing}
\end{figure}

\paragraph{\textit{Recency and first-action bias in internal representations.}}
The PCA visualization from \autoref{fig:pca_past_actions} suggests that the LLM's decision-making is a result of, among other things, a memory of past opponent actions. To understand the contents of this memory, we analyze how easily individual past opponent's actions can be extracted from the LLM's internal representations. As shown in \autoref{fig:per_round_probing}, the actions from the first and the last rounds are the most easily extractable --- the internal probe requires just around 20 training epochs to reach near-perfect decoding. In contrast, the extractability of actions from the intermediate rounds follows a U-shaped curve, with the opponent's actions in the middle rounds being the least decodable. The attention to the first and last opponent's actions mirrors recency bias observed in humans~\cite{murdock1962serial,yoo2025kq} and positional biases of LLMs observed in prior studies~\cite{laurito2025aiai,knipper2025bias,wu2025on}.

In practical applications, such U-shaped memory of past interactions could systematically bias strategic responses toward early commitments and recent deviations, making the LLM predictable and potentially exploitable. For example, an adversary could first signal cooperation or defection to anchor the LLM's internal state, then strategically deviate in intermediate rounds without much long-term penalty, and then realign behavior near the end of the interaction to influence the LLM's final decisions.

\begin{figure*}[t]
    \centering
    \includegraphics[width=\linewidth]{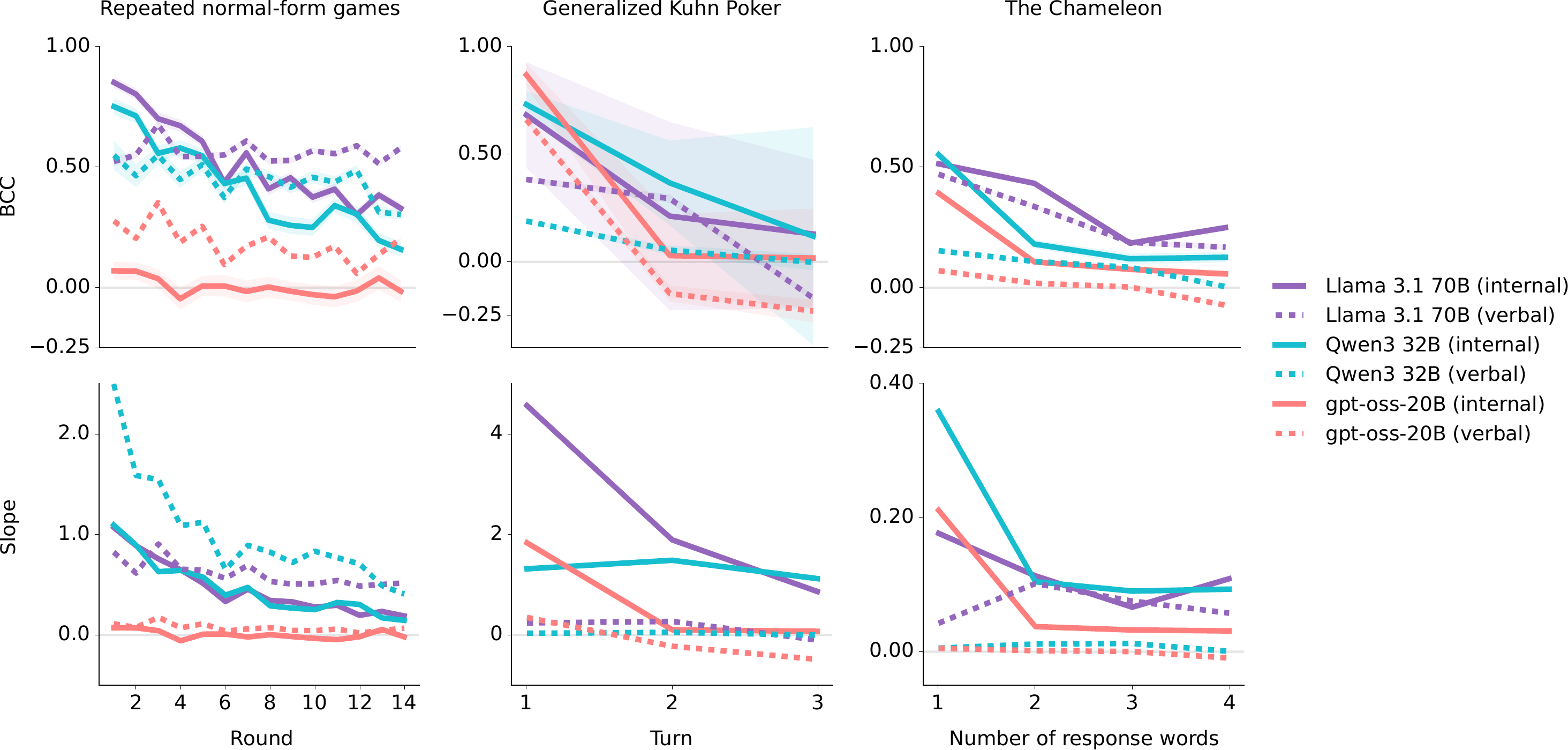}
    \vspace{-8pt}
    \caption{Progression of the Bayesian Coherence Coefficient (BCC; top) and the slope of its associated regression line (bottom) with increasing length of interaction. BCC is measured with respect to internal and verbal beliefs about the opponent's type (repeated normal-form games; 800 game trials), opponent's secret card (Generalized Kuhn Poker; 2,000 game trials), and the chameleon identity (The Chameleon; 1,168 game trials). Details in \autoref{app:bcc} in the Appendix.}
    \label{fig:bcc_joint}
\end{figure*}

\subsection{Belief Updating}

In multi-step strategic interactions with incomplete information, belief inference requires continual updating of beliefs in response to new evidence. The setting, therefore, naturally lends itself to the Bayesian framework: Robust Bayesian updating enables optimal control in partially observable environments~\cite{astrom1965,sondik1978}. Moreover, since people are often modeled as --- and model others as --- Bayesian reasoners~\cite{knill2004bayesian,baker2009inverse,griffiths2024bayesian}, LLMs that deviate from Bayesian updating become less predictable and interpretable to the humans who oversee or interact with them.

To evaluate whether current LLMs update beliefs in a Bayesian manner, we measure the Pearson correlation coefficient between the observed LLM's belief updates and optimal, self-consistent Bayesian updates. This evaluation metric has been formalized as the Bayesian Coherence Coefficient (BCC)~\cite{imran2025are}, which we extend from output-token probabilities to internal and verbal beliefs.

Formally, given an interaction history $h_{t}$ up to timestep $t$ and LLM's belief distribution $\hat b_t\coloneq P(Z=z \mid h_t)$ over some latent variable $Z$ (\eg, opponent's strategy type), we compute the belief update as:
\begin{equation}
\widehat{\Delta}_t(z,z')
\;\coloneq\;
\log\frac{\hat b_t(z)}{\hat b_t(z')}
-
\log\frac{\hat b_{t-1}(z)}{\hat b_{t-1}(z')}.
\end{equation}
Correspondingly, we compute the expected (Bayes) updates $\Lambda_t(z,z')$ based on a likelihood model of the observables that the LLM obtains at each timestep. In repeated normal-form games, for example, the LLM's observables are the opponent's actions $a_t\in\{A,B\}$ sampled from the action distribution $\pi_z$ given by the opponent's type $z\in\{1,2\}$. In this case, the Bayes-predicted updates become:
\begin{equation}
\Lambda_t(z,z')
=
\log
\frac{\pi_z\!\left(a_t\right)}{\pi_{z'}\!\left(a_t\right)}.
\end{equation}

Given these belief updates over randomly sampled pairs of latent variables, $\mathcal{Z} = \{(z,z') \in \mathrm{dom}(Z) \times \mathrm{dom}(Z)\}$, we compute the BCC as follows:
\begin{equation}
\mathrm{BCC}_t
\;\coloneq\;
\rho\!\left(
\mathrm{vec}\big(\{\widehat{\boldsymbol{\Delta}}_t\}\big),
\mathrm{vec}\big(\{\boldsymbol{\Lambda}_t\}\big)
\right),
\end{equation}
where $\widehat{\boldsymbol{\Delta}}_t \in \mathbb{R}^{|\mathcal{Z}|}$ and
$\boldsymbol{\Lambda}_t \in \mathbb{R}^{|\mathcal{Z}|}$ are the observed and the expected belief updates, respectively, $\rho(\cdot,\cdot)$ is the Pearson correlation coefficient, and $\mathrm{vec}(\cdot)$ denotes concatenation over game trials.

By correlating the observed updates in LLMs' beliefs with the ground-truth expected updates, BCC evaluates the extent to which LLMs remain consistent with the optimal Bayesian belief updating. Further details, including the likelihood models for other games, are included in \autoref{app:bcc} in the Appendix.

\paragraph{\textit{Bayesian belief updating deteriorates over extended interactions.}}
As shown in \autoref{fig:bcc_joint} on the top, internal belief updates exhibit substantially higher Bayesian optimality than verbal belief updates in early rounds. However, the BCC of both internal and verbal beliefs decays over time. Internal beliefs are especially susceptible to this deterioration: their BCC falls below half of the initial value by round 10 in normal-form games and by turn 3 in more complex games of Generalized Kuhn Poker and The Chameleon.

When we analyze the slope of the fitted regression line between the observed and expected updates of Llama 3.1 70B in repeated normal-form games (\autoref{fig:bcc_joint}, bottom; the underlying round-by-round scatter plots presented in \autoref{app:round_by_round_bcc}), we find that internal belief updates transition from early updating at correct magnitude (slope $\approx$ 1) to later under-updating (slope $<$ 1), providing further evidence for a systematic drift away from Bayesian optimality over long horizons. We observe a similar decrease in the magnitude of belief updates in other games and models (\autoref{fig:bcc_joint}, bottom).

These results show that while LLMs are capable of approximately Bayesian belief updating, this capability deteriorates over extended interactions. In high-stakes domains requiring long-horizon decision-making, such as multi-step planning or negotiation, such drift may lead to underreaction to new information or persistence in outdated hypotheses. Moreover, as the model's updates become less aligned with Bayesian expectations, it may become less predictable and understandable to agents with Bayesian world models (\eg, humans), highlighting an area for improvement.

\subsection{Belief-to-Action Conversion: Steering Internal Belief Representations}

To evaluate the extent to which the chosen internal beliefs about latent game variables are causally involved in decision-making, we intervene directly on the LLMs' hidden states. Specifically, we add scaled \textit{steering vectors} to the LLMs' hidden states at inference time. Similar to prior studies~\cite{li2023iti,marks2024the}, we use the learned weights of the linear (internal) probes as the steering vectors. The layer where we additively inject these vectors is based on the initial hyperparameter search of the internal probe (Appendix, Table~\ref{tab:appendix:internal_probe_hyperparameters}). By using the probe's weights directly, we bias the models' internal representations toward different opponents' strategies (repeated normal-form games), toward higher opponents' card values (Generalized Kuhn Poker), or toward different secret words (The Chameleon) without modifying the prompt or model parameters. This steering experiment builds on the idea of the linear representation hypothesis for LLMs~\cite{park2024the}, which posits that concepts in LLMs are represented along linear directions in the representation space. Our activation steering is inspired by prior work that similarly linearly shifted internal representations of linguistic and higher-level semantic concepts to make LLMs safer~\cite{bhattacharjee2024towards,batenburg2026steering}, more truthful~\cite{marks2024the,ravfogel2025emergence}, or less sycophantic~\cite{rimsky2023reducing}. Unlike these studies, however, we do not propose steering as a method for improving gameplay. We use it as a causal sensitivity analysis, testing whether the direction identified by the internal probe contributes to action selection.

To quantify the strength of the causal pathway from internal beliefs to actions, we compare the action distributions of the steered LLMs with those produced in a counterfactual setting where the targeted latent variable is actually true. For example, in Generalized Kuhn Poker, we steer the LLMs toward believing that the opponent holds a stronger private card and compare the resulting action distribution to one observed when the opponent indeed holds such a card and plays accordingly. We refer to this counterfactual setting as the \textit{contrast}, following prior work~\cite{turner2023steering,mallen2024eliciting}. We quantify the gap between the steered and counterfactual action distributions using the total variation distance. We use the same metric to select the highest-performing multiplication factor on 50 held-out trials. The considered search space is $\{1, 5, 10, 15, 20\}$. While this experiment tests how internal beliefs causally mediate actions, additional analyses evaluating statistical (correlational) mediation and layer-wise decodability at the action-selection token are provided in the Appendix.

\begin{figure}[t]
    \centering    \includegraphics[width=0.92\linewidth]{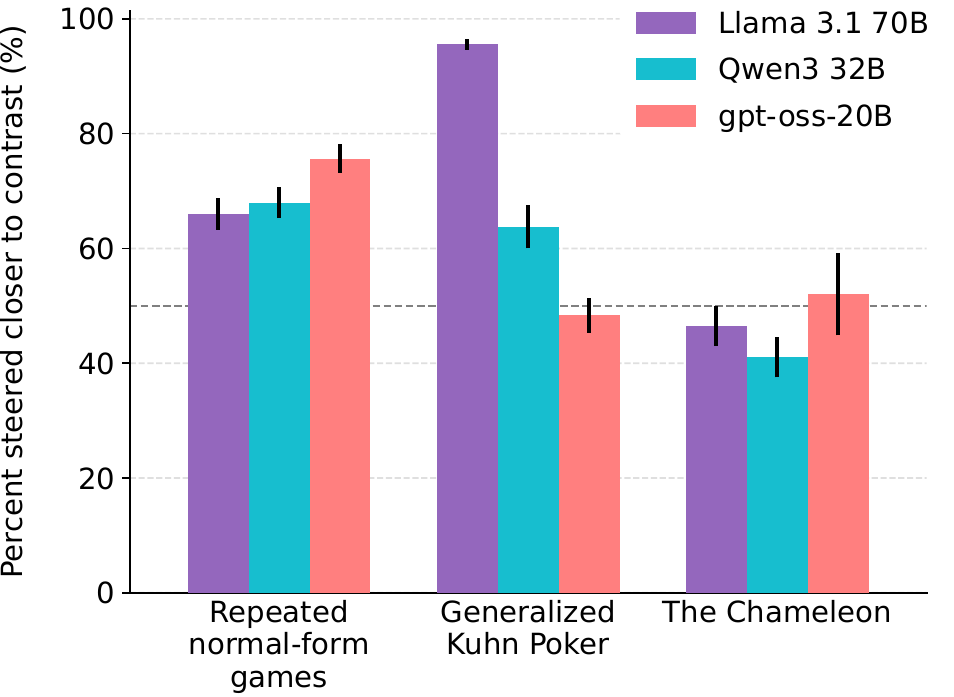}
    \caption{Steering internal representations toward different opponents' strategies (repeated normal-form games), toward higher opponents' secret cards (Generalized Kuhn Poker), and toward different secret words (The Chameleon). Shown are the means and their standard errors from 300 (repeated normal-form games) and 200 (Generalized Kuhn Poker, The Chameleon) game trials. Random chance is 50\%.}
    \label{fig:steering}
\end{figure}

\paragraph{\textit{Internal beliefs have a weak causal influence on action selection.}}
As shown in \autoref{fig:steering}, steering produces consistent but limited behavioral effects. In repeated normal-form games, steering internal beliefs toward different opponents' strategies (distributions over actions) brings the LLMs' strategies closer to the target counterfactuals in around 70\% of trials. For gpt-oss-20B in Generalized Kuhn Poker, this belief-to-action conversion reaches just around the 50\% chance level. The chance-level steering effects in The Chameleon are likely attributable to the construction of the steering vector in that setting. Because we steer with the same readout we probe with, the secret-word steering vector is a projection onto the model's own secret-word embeddings (\autoref{subsec:belief_formation}) rather than a trained linear-probe direction as in the other settings, and is therefore not optimized for decoding accuracy. The weak steering effects are consistent with the weak performance of this readout in the probing task (\autoref{fig:hidden_info}).

These results establish a causal pathway from the identified internal belief representations to action selection, but also demonstrate that this pathway is underutilized. Even when beliefs are manually nudged in the correct direction (\ie, toward beliefs consistent with the ground-truth), the resulting strategy changes are unreliable. This raises the question of how much of the models' suboptimal play is attributable to this incomplete routing from internal beliefs to actions, and how much better LLMs could play if these beliefs were used optimally.

\subsection{Belief-to-Action Conversion: Optimality of Actions}
To state the (under)utilization of internal beliefs in outcome-relevant units, we quantify the payoff that is achievable from the decoded beliefs but not realized by the model. For repeated normal-form games, we compare best responses (BR) to the internal and verbal beliefs $\hat{b}$, $\text{arg max}_a\mathbb{E}_{a'\sim\hat{b}}[u(a,a')]$, to the actions played by the LLM, along with a uniform random baseline and a ceiling expected payoff obtainable by BR with respect to the ground-truth opponent strategy. This comparison isolates the payoff that the extracted belief supports from the model's ability to act on it.

\begin{figure*}[t]
    \centering
    \includegraphics[width=\linewidth]{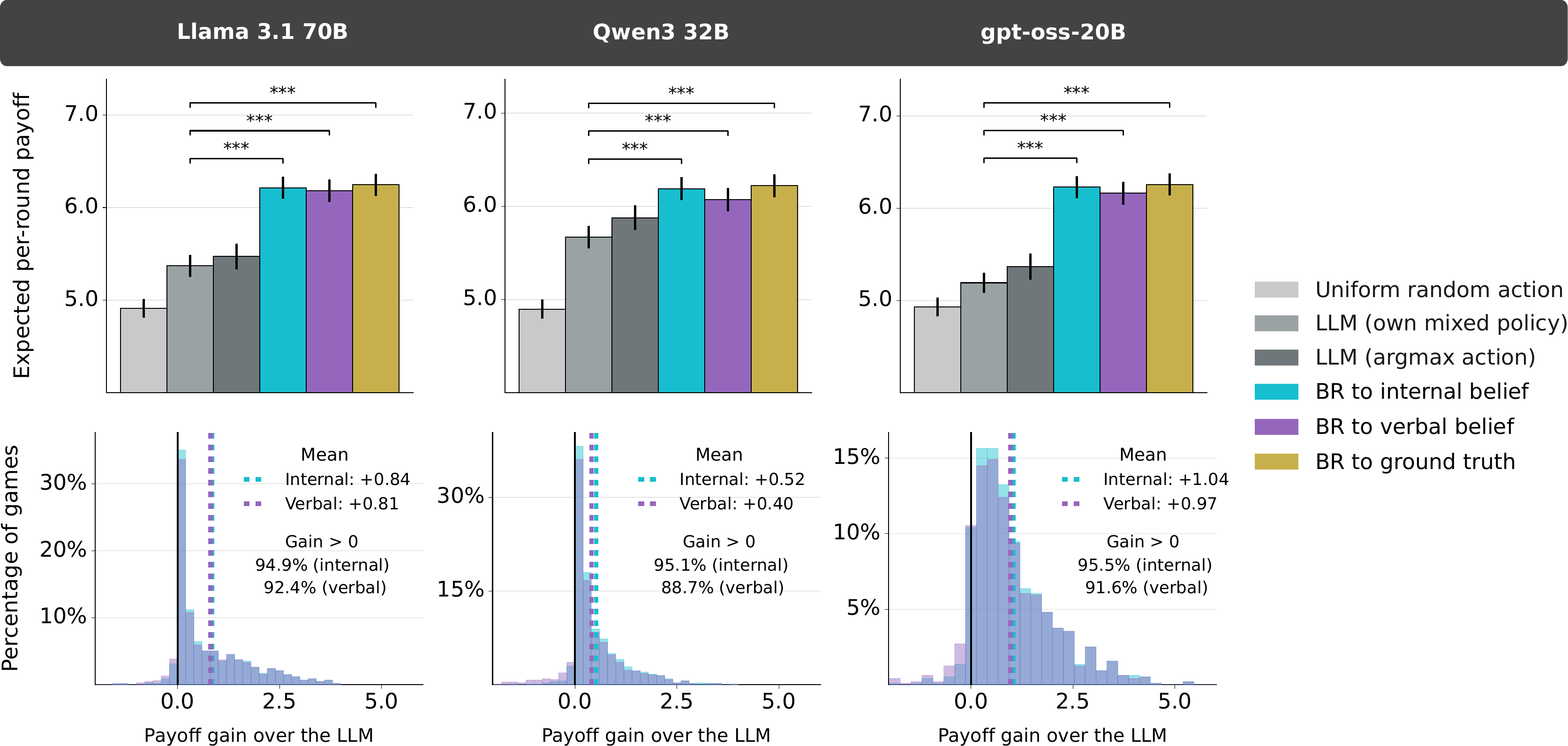}
    \vspace{-8pt}
    \caption{Payoffs obtained by LLMs vs. by best responses (BR) to the probed beliefs and ground-truth opponent's strategies in repeated normal-form games. \textbf{Top:} Mean expected per-round payoff of each decision rule (Random, LLM, BR). Mixed policy refers to the probability distribution over actions. BR to the ground-truth opponent strategy is the ceiling attainable with a perfectly accurate belief. Error bars denote 95\% confidence intervals of the mean, and asterisks denote statistical significance of the paired difference from the LLM's own mixed policy (two-sided Wilcoxon signed-rank test; * $p<0.05$; ** $p<0.01$; *** $p<0.001$). \textbf{Bottom:} Distribution over games of the per-game payoff gain of BR to the probed belief relative to the LLM's own mixed policy. Shown are results obtained from 1,000 game trials with payoff matrices sampled from $\mathcal{U}(0,10)$.}
    \label{fig:payoffs_wrt_br}
\end{figure*}

\begin{figure*}[h!]
    \centering
    \includegraphics[width=0.96\linewidth]{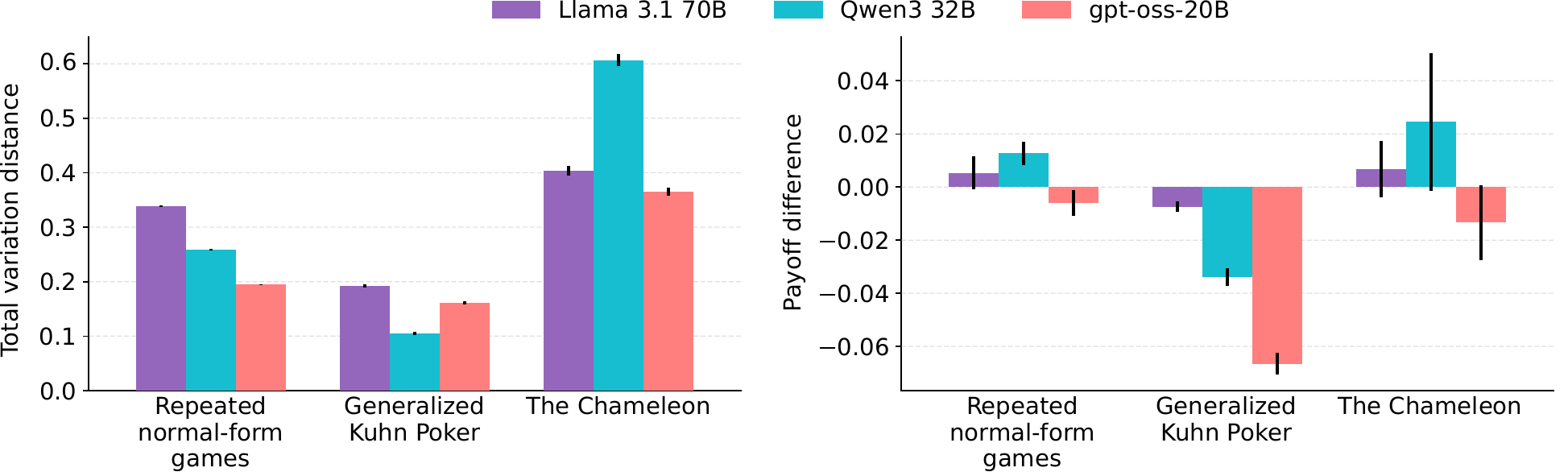}
    \caption{Differences in the gameplay of LLMs under implicit vs. explicit (prompt-based) conditioning by internal beliefs. Left: Total variation distance between the belief-conditioned action distributions of LLMs. Right: Payoff differences between the two types of conditioning. Shown are the means and their standard errors obtained from 2,000 (repeated normal-form games and Generalized Kuhn Poker) and 300 (The Chameleon) game trials.}
    \label{fig:belief_action_gap}
\end{figure*}

\paragraph{\textit{Conversion of internal beliefs into actions is suboptimal.}}
We find that across all three models, best-responding to the model's own internal belief yields a significantly higher payoff than the model's own play, with a gain in about 95\% of games (\autoref{fig:payoffs_wrt_br}). The gap to the ground-truth ceiling is small: for Llama 3.1 70B, the decoded belief already supports 96.2\% of the total achievable improvement over the model's own play. By contrast, the models themselves close only a fraction of the distance between uniform action selection and optimal play (34.4\% for Llama 3.1 70B), leaving a significant payoff increase unrealized. Substituting the model's temperature-zero argmax action for its full action distribution (\ie, policy) does not change this conclusion. Furthermore, the same pattern of belief underutilization holds for verbal beliefs.

This result establishes the belief-action gap: models contain easily accessible information about important latent game states (\eg, opponent's strategy) that would allow for more optimal gameplay than the models themselves achieve when relying on this information implicitly. This conclusion does not require the decoded belief to be the representation that fully captures the model's policy, since the comparison involves no intervention on the representations: whatever the relationship between the probe direction and the model's internal computation, the information recoverable at the action-selection step supports substantially better play than the model achieves. The same holds for verbal beliefs, where no internal probe direction is involved. Given this weak information routing in the model, a natural question arises whether explicit (verbal) conditioning on the internal beliefs improves the belief-to-action conversion and, in turn, closes the belief-action gap.

\subsection{Belief-to-Action Conversion: Implicit vs. Explicit Conditioning on Internal Beliefs}\label{subsec:bag_analysis}
We next quantify the differences between LLMs' gameplay when they play only according to their internal beliefs (implicit action conditioning) vs. when their internal beliefs are externalized in a prompt (explicit action conditioning). We measure this difference using two metrics: (i) the total variation distance between the internal-belief conditioned action distribution and the action distribution conditioned on the model's internal beliefs through prompt, and (ii) the expected difference in payoffs between these two types of conditioning.

\paragraph{\textit{Externalized beliefs have a strong causal influence on action selection, but are unreliable in improving gameplay.}}
As shown in \autoref{fig:belief_action_gap} on the left, there is a large mismatch between the LLMs' action distributions when conditioned implicitly vs. explicitly through prompts. Moreover, as shown on the right in \autoref{fig:belief_action_gap}, neither implicit nor explicit action conditioning by beliefs consistently achieves better payoffs. These results demonstrate that while LLMs are more sensitive to externalized beliefs than purely internal ones, neither is consistently better at improving expected game payoffs.

\begin{figure}[t!]
    \centering
    \includegraphics[width=0.9\linewidth]{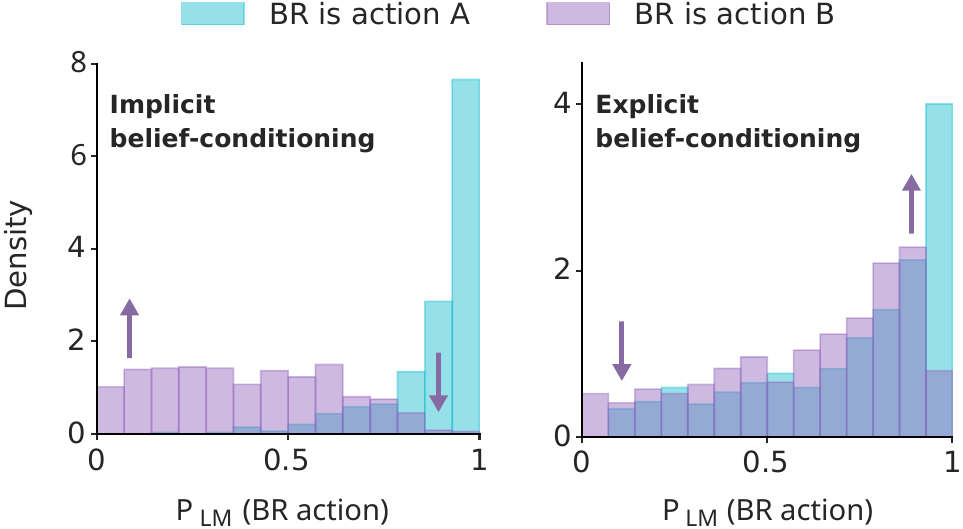}
    \caption{Effect of the action-letter bias on selecting the best response (BR) action by the LLM. The plot on the right illustrates a scenario where verbalizing internal beliefs about the opponent's strategy partially mitigates this bias. Shown is the histogram of next-action probabilities (x-axis) assigned by the LLM across 1,000 repeated normal-form games.}
    \label{fig:first_action_bias}
\end{figure}

\paragraph{\textit{Action-letter bias.}}
To further characterize failures in belief-to-action conversion, we analyze a systematic action-letter bias in repeated normal-form games. Specifically, we measure the probability that the LLM selects the best-response (BR) action implied by the internal belief about the opponent's strategy when that action corresponds to letter ``A'' versus the letter ``B''.

As shown in \autoref{fig:first_action_bias} on the left, when the BR corresponds to the action ``A'', the LLM assigns it high probability mass, often exceeding $0.9$. However, when the BR corresponds to the  action ``B'', the assigned probability is substantially lower and broadly distributed, even when internal beliefs clearly favor that action. This asymmetry reveals a strong token bias that interferes with belief-consistent rational decision-making. Permuting the action labels confirms this interpretation: the inflated probability mass follows the letter A rather than the first-listed option (Appendix, \autoref{app:additional_experimental_results}).

Interestingly, verbalizing the LLM's internal belief in the prompt partially mitigates this effect. When beliefs about the opponent's strategy are explicitly stated in the prompt, the distribution over action probabilities becomes more balanced between the two actions, with the probability of selecting the correct BR action increasing when it is labeled ``B''.

Taken together, these results reinforce the conclusion that even if LLMs possessed fully accurate (internal) beliefs, they do not act on them reliably or optimally. While explicit prompt conditioning does not improve gameplay in general, it can at least partially mitigate biases exhibited by LLMs.

\section{Discussion}
\label{sec:discuission}
LLMs are already being deployed in strategic decision-making contexts, from automated negotiation~\cite{bianchi2024how,abdelnabi2024negotiation,priya2025argue,kwon2025astra,kwon2025evaluating} to policy advisory systems~\cite{coz2025what,pan2025urban,ziegler2025,chen2025}, where their outputs can carry serious real-world consequences~\cite{abdelnabi2024negotiation,goktas2025strategic}. This has been reflected in a stream of scientific studies analyzing and improving LLMs as strategic agents~\cite{bianchi2024how,abdelnabi2024negotiation,priya2025argue,kwon2025astra,su2026endrewardengineeringllms,coz2025what,ziegler2025,chen2025,scott2024,lore2024,akata2025playing,guertler2025textarena}. We advance this line of work by opening the black box: using mechanistic interpretability tools to expose the internal mechanisms driving LLM decision-making and tracing where they diverge from game-theoretic rationality.

Across three classes of incomplete-information games, we find that LLM agents often contain useful latent game information, but struggle to maintain it stably over time and to use it reliably for rational action selection. We separate this into two broken links. 

First, the \textbf{observation-belief gap}: internal probes recover opponent strategies, Nash equilibria, and other latent game information significantly better than verbal probes, implying that models form linearly accessible implicit beliefs during decision-making that are not faithfully reported in text. However, belief formation is brittle: accuracy degrades as the compositional complexity increases, belief consistency deteriorates after repeated interactions, and the representation of interaction history exhibits a U-shaped bias toward the first and most recent observations. 

Second, the \textbf{belief-action gap}: although accurate internal beliefs are present at the action-selection step, their conversion into optimal actions is weak and prone to action-letter biases, and neither their implicit use nor their externalized prompt variants achieves consistent improvements in game payoffs. The same gap holds for verbal beliefs, which are the model's own reported estimates rather than probe-decoded ones. In frontier API-only models, it narrows, but does not disappear: for GPT-5.4 mini, the gap largely closes in the two-action games and reopens once the action space grows to three (Appendix, \autoref{app:additional_experimental_results}).

The weak effects of steering on internal beliefs leave open where in the computation the failure occurs. One possibility is that the models hold an accurate policy-relevant belief and convert it into actions poorly. Another is that action selection is influenced by a richer, distributed representation of which the probe direction is one component, so that the accurate information is present but incompletely routed to the decision. We regard this incomplete routing as a candidate mechanism for the poor belief-to-action conversion rather than an alternative to it. A linear readout sits at the lower end of the expressiveness available to the model itself, since every component that reads the residual stream, including the attention query, key, and value projections, the MLP up-projections, and the unembedding, begins by applying a learned linear map to that state. Information that linear probes recover is therefore within reach of the first operation of any downstream component, and the residual computation needed to act on it is a single expected-payoff comparison over two actions using a payoff matrix stated in the prompt. Localizing the failure within the forward pass, for instance through richer, context-dependent interventions, is a natural direction for future work.

These findings disentangle ``bad play'' into separate components, which behavior-only evaluation might conflate. Practically, the fact that accurate internal beliefs are often linearly decodable implies there is a strong substrate to build on: better interfaces for extracting beliefs, better training signals that bind beliefs to optimal actions, and better agent architectures that separate state estimation from decision could yield gains without requiring completely new capabilities. Our work also serves as a proof of concept for how mechanistic interpretability can identify intrinsic failure modes of LLM agents, ranging from action-letter biases and recency/primacy weighting to drifts in belief updating over long horizons. For example, the degradation of Bayesian coherence of beliefs over extended interactions suggests that even when models begin with approximately optimal updates, they can gradually under-react to new evidence, creating predictable vulnerabilities in negotiation, repeated coordination, or adversarial settings. Additionally, the U-shaped memory of past interactions implies systematic exploitability: an opponent can anchor early impressions of cooperation, strategically hide deviations in the middle stages of interaction, and steer final outcomes by showing cooperative moves in late stages.

Looking forward, our findings suggest concrete research directions. Reducing the observation-belief gap requires improving compositional latent inference and long-horizon belief maintenance (\eg, structured intermediate belief states, Chain-of-Thought reasoning, or training that directly targets multi-hop latent computations). Reducing the belief-action gap requires strengthening the causal link from beliefs to optimal decisions (\eg, calibration of action selection against belief-implied best responses, or debiasing action-letter effects). As LLMs continue to be deployed in strategic domains, our results suggest that guardrails and human oversight remain essential --- not as a general precaution, but because the specific vulnerabilities we identify are systematic, predictable, and currently unmitigated.

\section*{Acknowledgements}
This work was supported in part by the Army Research Office (ARO) under grant number W911NF-23-1-0317 and the Office of Naval Research (ONR) under grant number N00014-24-1-2432.

\bibliographystyle{unsrtnat}
\bibliography{references}

\clearpage
\appendix
\section{Appendix}
\subsection{Interpretation of Internal Beliefs}\label{app:internal_belief_interpretation}
Throughout the paper, we refer to the quantities recovered by the internal probes as \textit{internal beliefs}. Because this term carries different meanings in different literatures, we make our usage explicit here, explain why we adopt it, and discuss what our results do and do not establish about the role of these representations in the models' decision-making.
 
\paragraph{An operational definition grounded in game theory.}
Criteria for defining the beliefs of an LLM are not unique~\cite{herrmann2024standards}. Reasons such as the lack of ground truth and label asymmetries in datasets motivate the use of criteria including coherence, uniformity, and use. These challenges, which are common in settings studied in philosophy and economics, are not present in the game models that we consider. Game models allow identification of ground-truth variables that are sufficient for optimal decision-making~\cite{harsanyi1967games,kreps1982,astrom1965,sondik1978}, and the interaction history given in the context provides the evidence from which these variables can be inferred. We therefore use the accuracy of beliefs as the main criterion for belief extraction, and define an internal belief as the estimate of such a variable obtained by a linear readout of the representation that feeds into action selection. This definition lets us break the decision-making chain into independent pieces: extracting the most accurate and readily accessible contained information about the variables of interest, and analyzing whether this contained information is utilized for the downstream task of choosing actions. The separation mirrors that made in belief-based approaches in game theory~\cite{kreps1982,astrom1965,sondik1978}, which distinguish the inference of beliefs from observations and the conversion of beliefs into actions.
 
We stress that this definition does not presuppose that the model uses these representations for decision-making, let alone optimally. On the contrary, quantifying the extent to which the model does or does not exploit the decodable information is the purpose of our belief-action gap analysis. We also do not intend the cognitive-science reading of the term, which typically concerns settings in which beliefs are neither unanimously defined nor tied to an externally specified ground truth~\cite{schwitzgebel2024belief,dennett1989intentional}.
 
\paragraph{Relation to prior usage.}
Referring to probe-decoded latent-state estimates as beliefs or knowledge follows established usage. Prior work has trained probes on internal representations to predict ground-truth external variables in question answering~\cite{schouten2025,liu2024universal} and to predict the correctness of a model's own output~\cite{yang2025}, describing the recovered quantity in these terms. Where our setting differs from that literature is in the availability of the target. Approaches such as CCS~\cite{burns2023discovering} search for truth-like directions without supervision precisely because reliable ground truth is often unavailable for open-ended natural-language claims, and this unavailability is what motivates the alternative criteria discussed above~\cite{herrmann2024standards}. In games, the decision-relevant latent variables are specified by the environment, so supervised probing is available to us and gives a target that is unambiguous by construction.
 
\paragraph{Why we restrict the readout to be linear.}
Two considerations motivate the restriction. First, the goal is to quantify the inferential capability already present in the model rather than to add expressive power to it~\cite{belinkov2022probing,elazar2021amnesicprobing}, and more expressive probes risk learning the target function themselves instead of reading out what the model has computed~\cite{hewitt2019,voita2020information}. We report a nonlinear probing ablation and a minimum description length analysis in \autoref{app:alternative_probing}, both of which support the conclusions drawn from the linear probes.
 
Second, linear accessibility is the decision-relevant notion in our setting. Up to layer normalization, the components of a transformer read the residual stream through linear projections, and the next-token logits are a linear function of the final hidden state, so at the action-selection step information can influence the action only to the extent that it is linearly exposed. A linear readout therefore sits at the lower end of the expressiveness available to the model itself: every component that reads the residual stream, including the attention query, key, and value projections, the up-projections of each MLP, and the unembedding, begins by applying a learned linear map to that state. Whatever our probes recover linearly is within reach of the first operation of any downstream component, so the information is present in the form that the model reads most directly.
 
\paragraph{What the definition implies for the belief-action gap.}
A natural objection to our terminology is that the probe may capture only one component of a richer and more distributed representation that governs action, in which case a weak effect of intervening on the probe direction would indicate a poor approximation to the operative representation rather than a weak conversion of internal beliefs into actions. Under our definition, these two readings largely coincide. Suppose the objection is correct and the representation governing action is distributed and only partially aligned with the probe direction. The accurate estimate is nonetheless present at the decision point, in the form the model reads most directly. If the action-governing representation does not fully exploit this accurate and readily accessible estimate, then the model fails to convert available belief information into optimal actions, which is what we call the belief-action gap. A poorer approximation of the probe direction to the operative representation is therefore not an alternative to weak conversion, but one candidate mechanism by which the conversion fails, in this case through routing or integration rather than through a final decision rule.
 
Two further points are worth stating. First, our central evidence for the gap does not depend on this interpretation at all. The payoff analysis in the main text compares the payoff obtainable by best-responding to the extracted belief with the payoff the model realizes, and it involves no intervention on the representations, so it holds whatever the relationship between the probe direction and the model's internal computation. The same gap holds for verbal beliefs, where the model has already externalized the estimate, and no internal direction is involved. Second, the accuracy of our probes with respect to the ground truth and their adequacy as a measure of the operative representation are separate quantities, so a loss of adequacy in the more compositional conditions cannot by itself explain the loss of accuracy that we report in the main text.
 
\paragraph{Verbal beliefs are not a measure of operativeness.}
Finally, we caution against reading the internal-verbal comparison as a measure of how operative the internal beliefs are. LLMs are well documented to produce verbal reports that are plausible yet causally disconnected from their internal computation~\cite{turpin2023,chen2025reasoningmodelsdontsay}. Verbalization and the operativeness of internal-belief representations therefore vary separately, and a narrowing gap on a decoding metric is not evidence that the verbal probe better tracks the operative representation in more compositional settings.
 
\paragraph{What remains open.}
Our definition and results leave open where in the computation the conversion failure occurs, and hence to which stages a routing failure could be localized. The layer-wise analysis in \autoref{app:layerwise_tracing} is a first step: the latent variable remains decodable in the layers where the action is formed, which indicates a failure of utilization rather than of availability. Localizing the failure further, for instance with richer and context-dependent interventions, is a natural direction for future work.

\subsection{Experimental Design}\label{app:experimental_design}

\paragraph{Interaction with LLMs.}
We used instruction-tuned variants of Qwen3 32B (\nolinkurl{Qwen/Qwen3-32B-AWQ}, quantized), Llama 3.1 70B (\nolinkurl{hugging-quants/Meta-Llama-3.1-70B-Instruct-AWQ-INT4}, quantized), and gpt-oss 20B (\nolinkurl{openai/gpt-oss-20b}) from HuggingFace. The selection of these models was based on computational constraints, their established popularity and performance~\cite{llama3,qwen3,bi2025gptoss}, and the observed inability of smaller models to adhere to game rules or formatting requirements. To ensure reproducibility and facilitate precise hidden-state analysis, all models generated their outputs with a temperature set to zero.

\paragraph{Belief formation.}
Experiments on belief formation used the following setup. For probing the opponent's strategy in repeated normal-form games, we sampled payoff matrix entries from a uniform distribution $\mathcal{U}(0,10)$ and used two random samples from the Dirichlet distribution ($\alpha=(1.0,1.0)$) as the policies (probability distributions over actions) of the LLM's rollout and the opponent. We then simulated play between the rollout and opponent policies for $T \thicksim \mathcal{U}(0,30)$ rounds. We recorded the LLM's internal representations while it generated the action letter (``A'' or ``B'') for the subsequent round. For Nash equilibrium inference, we set $T=0$ and constructed zero-sum payoff matrices by sampling one player's payoffs from $\mathcal{U}(0,10)$ and setting the opponent's to the negative. This $T=0$ constraint ensured that inference relied solely on the game structure (payoff matrices) rather than an empirical estimate of the equilibrium strategies from the action history.

In trials of Generalized Kuhn Poker, we let three LLMs play against each other for three betting rounds. Private cards were sampled uniformly from a deck of 20 cards. Initial stacks were set to 100, with allowed bet sizes of [1, 3, 5, 10, 15, 20, 50] chips. The prompt used for all LLMs is given in \autoref{fig:kuhn_poker_prompt}.

For probing the secret word and chameleon identity in the Chameleon game, we collected game trials of four LLMs playing against each other. We use the original categories and secret words from The Chameleon~\cite{chameleon}, combined with additional categories and secret words generated by prompting GPT-4o (\verb|gpt-4o-2024-08-06|). In total, there are 20 original and 54 LLM-generated categories, where each category has 16 potential secret words. The prompt for asking GPT-4o to generate new categories and possible secret words is shown in \autoref{fig:chameleon_new_game_cards_prompt}, and the prompts used for trials of The Chameleon game itself are displayed in Boxes~\ref{fig:chameleon_base_prompt}, \ref{fig:chameleon_nonchameleon_prompt}, and \ref{fig:chameleon_chameleon_prompt}.

\subsection{Probing}\label{app:probing}
\paragraph{Internal probes.}
For inferring target latent variables (\eg, the opponent's strategy), we parametrize internal probes as linear models $f_\theta: \mathbb{R}^{d} \rightarrow \mathbb{R}^{z}, \ f_\theta(\mathbf{h}) = \mathbf{W} \mathbf{h} + \mathbf{b}$ where $\theta = \{\mathbf{W} \in \mathbb{R}^{z,d},\ \mathbf{b} \in \mathbb{R}^z\}$ are the internal probe's parameters, $d$ is the dimension of the LLM's internal representations $\mathbf{h} \in \mathbb{R}^d$ and $z$ is the dimension of the latent variable to infer (\eg, $z=2$ for opponent's strategy). Since the opponent's strategy and the two mixed strategies in the Nash equilibrium are probability distributions, we apply softmax at the outputs of the corresponding internal probes. Similarly, we apply softmax when predicting probability distributions over discrete target latent variables (\eg, opponent's secret card).

We train all internal probes to minimize the cross-entropy loss using the Adam optimizer~\cite{kingma2014adam}. We use weight decay to regularize the optimization problem. For repeated normal-form games, we collect 4,000 independent game trials and split them into training (65\%), validation (15\%), and test sets (20\%). We perform the same 65/15/20 split for 2,000 independent game trials of Generalized Kuhn Poker and all categories and secret words in The Chameleon (73 categories with 16 secret words each, for a total of 1,168 trials). Validation sets are used to search for the best hyperparameters for training the internal probes. The hyperparameter search space and selected values are provided in \autoref{tab:appendix:internal_probe_hyperparameters}.

\paragraph{Verbal probes.}
Verbal probes used the same data for evaluation as the internal probes. However, verbal probes did not involve any additional training or hyperparameter search. Instead, we prompted the LLM with the original game context suffixed with a question about the latent variable. For classification tasks, we used the normalized (log) probabilities assigned by the LLM to tokens corresponding to the possible target values (\eg, numbers 1 to 20 for the opponent's secret card)~\cite{si2023prompting,petroni2019language,peng2022copen,imran2025are}. For tasks of inferring probability distributions (\eg, the opponent's strategy), we asked the LLM to output the full distribution in JSON format, similar to prior studies~\cite{wang2025observer,suh2025language}. Prompts are provided in Boxes \ref{fig:chameleon_nonchameleon_prompt}, \ref{fig:matrix_vp}, \ref{fig:kuhn_poker_vp}, and \ref{fig:chameleon_secret_word_vp}.

We did not observe any failures in parsing LLM outputs, as we caught corner cases in the initial stages of the study. In particular, when probing probability distributions in repeated normal-form games, models sometimes output probabilities in fractional form (\eg, \verb|{"A": 4/6, "B": 2/6}|).

\begin{table*}[t]
    \centering
    \renewcommand{\arraystretch}{1.16}
    \setlength{\tabcolsep}{6pt}
    \caption{Hyperparameter search space for internal probes, with final selected values \underline{underlined}.}
    \vspace{0.15cm}
    \resizebox{0.98\textwidth}{!}{
    \begin{tabular}{lcccccc}
        \toprule
         \textbf{Target variable} & \textbf{Learning rate} & \textbf{Weight decay} & \textbf{Num. of epochs} & \textbf{Batch size} & \textbf{Layer number} \\
        \midrule 
        
        \multicolumn{6}{@{}l}{\textbf{Repeated normal-form games}}\\
        
        \textsc{Opponent's strategy (Llama 3.1 70B)}
            & $\{\text{1e-3, 1e-4, } \ul{1e-5}\}$
            & $\{\ul{0}, \text{1e-3, 1e-5}\}$
            & $\{\text{20, 50, 100, 200, 400, } \ul{600}\}$
            & $\{\ul{32}, \text{128}\}$
            & $\{\ul{10}, \text{20, 30, 40, 50, 60}\}$\\
        
        \textsc{Opponent's strategy (Qwen3 32B)}
            & $\{\text{1e-3, 1e-4, } \ul{1e-5}\}$
            & $\{\ul{0}, \text{1e-3, 1e-5}\}$
            & $\{\text{20, 50, 100, 200, 400, } \ul{600}\}$
            & $\{\ul{32}, \text{128}\}$
            & $\{\ul{10}, \text{20, 30, 40, 50, 60}\}$\\

        \textsc{Opponent's strategy (gpt-oss 20B)}
            & $\{\text{1e-3, 1e-4, } \ul{1e-5}\}$
            & $\{\text{0, } \ul{1e-3}, \text{1e-5}\}$
            & $\{\text{20, 50, 100, 200, 400, } \ul{600}\}$
            & $\{\ul{32}, \text{128}\}$
            & $\{\text{5, 10, 15, } \ul{20}\}$\\

        \textsc{Nash equilibrium (Llama 3.1 70B)}
            & $\{\text{1e-3, } \ul{1e-4}, \text{1e-5}\}$
            & $\{\text{0, } \ul{1e-3}, \text{1e-5}\}$
            & $\{\text{20, 50, 100, 200, } \ul{400}, \text{600}\}$
            & $\{\text{32, } \ul{128}\}$
            & $\{\text{10, 20, 30, } \ul{40}, \text{50, 60}\}$\\
        
        \textsc{Nash equilibrium (Qwen3 32B)}
            & $\{\text{1e-3, 1e-4, } \ul{1e-5}\}$
            & $\{\text{0, } \ul{1e-3}, \text{1e-5}\}$
            & $\{\text{20, 50, 100, 200, 400, } \ul{600}\}$
            & $\{\text{32, } \ul{128}\}$
            & $\{\text{10, 20, 30, 40, } \ul{50}, \text{60}\}$\\

        \textsc{Nash equilibrium (gpt-oss 20B)}
            & $\{\text{1e-3, 1e-4, } \ul{1e-5}\}$
            & $\{\text{0, } \ul{1e-3}, \text{1e-5}\}$
            & $\{\text{20, 50, 100, 200, 400, } \ul{600}\}$
            & $\{\ul{32}, \text{128}\}$
            & $\{\text{5, 10, 15, } \ul{20}\}$\\

        \textsc{Opp. type by strategy (Llama 3.1 70B)}
            & $\{\text{1e-3, 1e-4, } \ul{1e-5}\}$
            & $\{\ul{0}, \text{1e-3, 1e-5}\}$
            & $\{\text{20, 50, 100, 200, 400, } \ul{600}\}$
            & $\{\ul{32}, \text{128}\}$
            & $\{\text{10, 20, 30, 40, 50, } \ul{60}\}$\\
        
        \textsc{Opp. type by strategy (Qwen3 32B)}
            & $\{\ul{1e-3}, \text{1e-4, 1e-5}\}$
            & $\{\ul{0}, \text{1e-3, 1e-5}\}$
            & $\{\text{20, 50, 100, 200, } \ul{400}, \text{600}\}$
            & $\{\ul{32}, \text{128}\}$
            & $\{\text{10, 20, 30, 40, } \ul{50}, \text{60}\}$\\

        \textsc{Opp. type by strategy (gpt-oss 20B)}
            & $\{\ul{1e-3}, \text{1e-4, 1e-5}\}$
            & $\{\text{0, } \ul{1e-3}, \text{1e-5}\}$
            & $\{\text{20, 50, 100, 200, } \ul{400}, \text{600}\}$
            & $\{\ul{32}, \text{128}\}$
            & $\{\text{5, 10, 15, } \ul{20}\}$\\

        \textsc{Opp. type by strategy and round (Llama 3.1 70B)}
            & $\{\ul{1e-3}, \text{1e-4, 1e-5}\}$
            & $\{\ul{0}, \text{1e-3, 1e-5}\}$
            & $\{\text{20, 50, 100, 200, } \ul{400}, \text{600}\}$
            & $\{\ul{32}, \text{128}\}$
            & $\{\text{10, 20, } \ul{30}, \text{40, 50, 60}\}$\\
        
        \textsc{Opp. type by strategy and round (Qwen3 32B)}
            & $\{\text{1e-3, 1e-4, } \ul{1e-5}\}$
            & $\{\ul{0}, \text{1e-3, 1e-5}\}$
            & $\{\text{20, 50, } \ul{100}, \text{200, 400, 600}\}$
            & $\{\ul{32}, \text{128}\}$
            & $\{\text{10, 20, 30, 40, } \ul{50}, \text{60}\}$\\

        \textsc{Opp. type by strategy and round (gpt-oss 20B)}
            & $\{\text{1e-3, } \ul{1e-4}, \text{1e-5}\}$
            & $\{\ul{0}, \text{1e-3, 1e-5}\}$
            & $\{\text{20, 50, 100, } \ul{200}, \text{400, 600}\}$
            & $\{\ul{32}, \text{128}\}$
            & $\{\ul{5}, \text{10, 15, 20}\}$\\

        \textsc{Opp. type by payoffs (Llama 3.1 70B)}
            & $\{\ul{1e-3}, \text{1e-4, 1e-5}\}$
            & $\{\text{0, 1e-3, } \ul{1e-5}\}$
            & $\{\text{20, 50, 100, } \ul{200}, \text{400, 600}\}$
            & $\{\ul{32}, \text{128}\}$
            & $\{\text{10, 20, 30, } \ul{40}, \text{50, 60}\}$\\
        
        \textsc{Opp. type by payoffs (Qwen3 32B)}
            & $\{\ul{1e-3}, \text{1e-4, 1e-5}\}$
            & $\{\ul{0}, \text{1e-3, 1e-5}\}$
            & $\{\text{20, 50, 100, } \ul{200}, \text{400, 600}\}$
            & $\{\ul{32}, \text{128}\}$
            & $\{\text{10, 20, 30, 40, } \ul{50}, \text{60}\}$\\

        \textsc{Opp. type by payoffs (gpt-oss 20B)}
            & $\{\text{1e-3, 1e-4, } \ul{1e-5}\}$
            & $\{\ul{0}, \text{1e-3, 1e-5}\}$
            & $\{\text{20, } \ul{50}, \text{100, 200, 400, 600}\}$
            & $\{\text{32, } \ul{128}\}$
            & $\{\text{5, 10, 15, } \ul{20}\}$\\

        \multicolumn{6}{@{}l}{\textbf{Generalized Kuhn Poker}}\\
        
        \textsc{Opponent's secret card (Llama 3.1 70B)}             
            & $\{\ul{1e-3}, \text{1e-4, 1e-5}\}$
            & $\{\text{0, 1e-3, } \ul{1e-5}\}$
            & $\{\text{20, 50, 100, } \ul{200}, \text{400}\}$
            & $\{\ul{32}, \text{128}\}$
            & $\{\text{10, } \ul{20}, \text{30, 40, 50, 60}\}$\\

        \textsc{Opponent's secret card (Qwen3 32B)}
            & $\{\ul{1e-3}, \text{1e-4, 1e-5}\}$
            & $\{\text{0, 1e-3, } \ul{1e-5}\}$
            & $\{\text{20, 50, 100, 200, } \ul{400}\}$
            & $\{\text{32, } \ul{128}\}$
            & $\{\text{10, 20, 30, 40, 50, } \ul{60}\}$\\

        \textsc{Opponent's secret card (gpt-oss 20B)}
            & $\{\text{1e-3, 1e-4, } \ul{1e-5}\}$
            & $\{\ul{0}, \text{1e-3, 1e-5}\}$
            & $\{\text{20, 50, 100, 200, } \ul{400}\}$
            & $\{\text{32, } \ul{128}\}$
            & $\{\text{5, 10, } \ul{15}, \text{20}\}$\\

        \multicolumn{6}{@{}l}{\textbf{The Chameleon}}\\
        
        \textsc{Secret word (Llama 3.1 70B)}
            & N/A & N/A & N/A & N/A
            & $\{\text{10, 20, 30, 40, 50, } \ul{60}\}$\\
        
        \textsc{Secret word (Qwen3 32B)}
            & N/A & N/A & N/A & N/A
            & $\{\text{10, 20, 30, } \ul{40}, \text{50, 60}\}$\\
        
        \textsc{Secret word (gpt-oss 20B)}
            & N/A & N/A & N/A & N/A
            & $\{\ul{5}, \text{10, 15, 20}\}$\\
        
        \textsc{Chameleon identity (Llama 3.1 70B)}
            & $\{\ul{1e-3}, \text{1e-4, 1e-5}\}$
            & $\{\text{0, 1e-3, } \ul{1e-5}\}$
            & $\{\text{20, 50, 100, 200, } \ul{400}, \text{600}\}$
            & $\{\ul{32}, \text{128}\}$
            & $\{\text{10, 20, 30, } \ul{40}, \text{50, 60}\}$\\
        
        \textsc{Chameleon identity (Qwen3 32B)}
            & $\{\text{1e-3, } \ul{1e-4}, \text{1e-5}\}$
            & $\{\text{0, } \ul{1e-3}, \text{1e-5}\}$
            & $\{\text{20, 50, 100, 200, } \ul{400}, \text{600}\}$
            & $\{\ul{32}, \text{128}\}$
            & $\{\text{10, 20, 30, 40, 50, } \ul{60}\}$\\
        
        \textsc{Chameleon identity (gpt-oss 20B)}
            & $\{\text{1e-3, } \ul{1e-4}, \text{1e-5}\}$
            & $\{\ul{0}, \text{1e-3, 1e-5}\}$
            & $\{\text{20, 50, 100, } \ul{200}, \text{400}\}$
            & $\{\ul{32}, \text{128}\}$
            & $\{\text{5, 10, } \ul{15}, \text{20}\}$\\

        \bottomrule
    \end{tabular}}
    \label{tab:appendix:internal_probe_hyperparameters}
\end{table*}
\newpage

\subsection{Bayesian Coherence Coefficient (BCC)}\label{app:bcc}
For evaluating whether the LLM belief updates are internally consistent with Bayes' rule, we use the Bayesian Coherence Coefficient (BCC)~\cite{imran2025are}, extending it to both
(i) verbal beliefs and (ii) internal beliefs.

\paragraph{Generic derivation.}
Consider an interaction indexed by $t=1,\dots,T$. Let $Z$ denote a latent variable of interest (\eg, opponent's type, opponent's secret card), let $O_t$ be the newly observed evidence at step $t$ (\eg, an opponent action, a bet, or a clue word), and let $h_{t-1}$ denote the interaction history prior to observing $O_t=o_t$. Let the agent's belief at time $t$ be
\begin{equation}
b_t(z) \;\coloneq\; P(Z=z \mid h_t), \qquad h_t \coloneq (h_{t-1}, o_t).
\end{equation}
Bayes' rule gives
\begin{equation}
P(z \mid o_t, h_{t-1})
=
\frac{P(o_t \mid z, h_{t-1}) \, P(z \mid h_{t-1})}
     {P(o_t \mid h_{t-1})},
\label{eq:bayes_rule_generic}
\end{equation}
where we use $P(z)$ as a shorthand for $P(Z=z)$.

To express belief updates, we consider log-odds ratios between two hypotheses $z$ and $z'$.
From~\eqref{eq:bayes_rule_generic},
\begin{equation}
\frac{P(z \mid o_t, h_{t-1})}{P(z' \mid o_t, h_{t-1})}
=
\frac{P(o_t \mid z, h_{t-1})}{P(o_t \mid z', h_{t-1})}
\cdot
\frac{P(z \mid h_{t-1})}{P(z' \mid h_{t-1})},
\end{equation}

\begin{equation}
\begin{aligned}
&\underbrace{
\log\frac{P(z \mid o_t, h_{t-1})}{P(z' \mid o_t, h_{t-1})}
-
\log\frac{P(z \mid h_{t-1})}{P(z' \mid h_{t-1})}
}_{\text{Bayes log-odds update}} \\
&=
\log\frac{P(o_t \mid z, h_{t-1})}{P(o_t \mid z', h_{t-1})}.
\end{aligned}
\label{eq:logodds_update_equals_llr}
\end{equation}

Thus, under Bayes-optimal updating, the log-odds update equals the log-likelihood ratio (LLR):
\begin{equation}
\underbrace{\Delta_t(z,z')}_{\text{Bayes log-odds update}}
\;=\;
\underbrace{\Lambda_t(z,z')}_{\text{LLR}}
,\ \ \text{where}
\end{equation}

\begin{equation}
\Delta_t(z,z')
\;\coloneq\;
\log\frac{b_t(z)}{b_t(z')}
-
\log\frac{b_{t-1}(z)}{b_{t-1}(z')}
\end{equation}

\begin{equation}
\Lambda_t(z,z')
\;\coloneq\;
\log\frac{P(o_t \mid z, h_{t-1})}{P(o_t \mid z', h_{t-1})}.
\label{eq:expected_update}
\end{equation}

\paragraph{Observed vs. expected (Bayes-predicted) updates.}
In our experiments, we obtain a belief distribution $\hat b_t$ either from (i) a verbal probe (LLM-produced distribution) or (ii) an internal probe (softmax of a linear model).
We then compute the observed update
\begin{equation}
\widehat{\Delta}_t(z,z')
\;\coloneq\;
\log\frac{\hat b_t(z)}{\hat b_t(z')}
-
\log\frac{\hat b_{t-1}(z)}{\hat b_{t-1}(z')}.
\label{eq:observed_update}
\end{equation}
To compute the corresponding expected (Bayes-predicted) updates $\Lambda_t(z,z')$, we require a likelihood model $P(o_t \mid z, h_{t-1})$, which is specified separately for each game below.

Notice that BCC is a \textit{self-consistency} measure: it evaluates whether the agent's realized belief changes $\widehat{\Delta}_t$ track the Bayes-optimal changes implied by the same evidence under the likelihood model.

\paragraph{Definition of BCC.}
For discrete latent variables, we evaluate updates for randomly sampled pairs
$\mathcal{Z} = \{(z,z') \in \mathrm{dom}(Z) \times \mathrm{dom}(Z)\}$ and form vectors
$\widehat{\boldsymbol{\Delta}}_t \in \mathbb{R}^{|\mathcal{Z}|}$ and
$\boldsymbol{\Lambda}_t \in \mathbb{R}^{|\mathcal{Z}|}$ by stacking
$\widehat{\Delta}_t(z,z')$ and $\Lambda_t(z,z')$ over $(z,z')\in\mathcal{Z}$ obtained from all collected game trials.
Then, we report
\begin{equation}
\mathrm{BCC}_t
\;\coloneq\;
\rho\!\left(
\mathrm{vec}\big(\{\widehat{\boldsymbol{\Delta}}_t\}\big),
\mathrm{vec}\big(\{\boldsymbol{\Lambda}_t\}\big)
\right),
\end{equation}
where $\rho(\cdot,\cdot)$ is the Pearson correlation coefficient and $\mathrm{vec}(\cdot)$ denotes concatenation over trials. We compute BCC separately for each timestep $t$ (\eg, individual rounds in repeated normal-form games) and for internal and verbal beliefs.

\paragraph{Likelihood model for repeated normal-form games.}
In this setting, the latent variable is the opponent's (hidden) type:
\begin{equation}
Z \in \mathrm{dom}(Z) \coloneq \{1,2\}.
\end{equation}
Each type $z \in \mathrm{dom}(Z)$ is associated with a known memoryless stochastic policy $\pi_z(\cdot)$ over actions $\{A,B\}$.

The observation at time $t$ is the opponent's action
\begin{equation}
o_t = a_t \in \{A,B\},
\end{equation}
and the history $h_{t-1}$ contains all previous actions and any publicly revealed side information.

Because the opponent is memoryless conditional on its type, the likelihood term in
\eqref{eq:expected_update} becomes
\begin{equation}
P(o_t \mid z, h_{t-1})
=
P(a_t \mid z, h_{t-1})
=
\pi_z\!\left(a_t\right).
\end{equation}

The Bayes-predicted log-likelihood ratio between the two candidate types $z$ and $z'$ is therefore
\begin{equation}
\Lambda_t(z,z')
=
\log
\frac{\pi_z\!\left(a_t\right)}{\pi_{z'}\!\left(a_t\right)}.
\end{equation}
We correlate these Bayes-predicted updates with the observed log-odds updates computed from the model's beliefs $\hat b_t(z)$ via~\eqref{eq:observed_update} to obtain BCC.

\paragraph{Likelihood model for Generalized Kuhn Poker.}
In this setting, the latent variable is the opponent's private card:
\begin{equation}
Z \in \mathrm{dom}(Z) \coloneq \{1,\dots,D\}.
\end{equation}
Each value $z \in \mathrm{dom}(Z)$ corresponds to a possible card rank dealt to the opponent at the beginning of the hand.

The observation at time $t$ is the opponent's publicly visible action
\begin{equation}
o_t = a_t \in \mathcal{A}(s_t),
\end{equation}
where $\mathcal{A}(s_t)$ denotes the set of legally available actions (\eg, call, fold) in the current game state $s_t$. The history $h_{t-1}$ contains all previous actions and public game information.

For the likelihood term in~\eqref{eq:expected_update}, we use the opponent LLM's own action probabilities. Specifically, for each candidate private card $z$, we condition the opponent LLM on the public history $h_{t-1}$ together with the counterfactual assignment that its private card equals $z$, and extract the model's probability distribution $P_{\mathrm{LLM}}(\cdot \mid z, h_{t-1})$ over the next possible actions.

The Bayes-predicted log-likelihood ratio between two candidate card values $z$ and $z'$ is therefore
\begin{equation}
\Lambda_t(z,z')
=
\log
\frac{P_{\mathrm{LLM}}(o_t \mid z,h_{t-1})}
     {P_{\mathrm{LLM}}(o_t \mid z', h_{t-1})}.
\end{equation}

\paragraph{Likelihood model for The Chameleon.}
In this setting, the latent variable is the hidden identity of the chameleon:
\begin{equation}
Z \in \mathrm{dom}(Z) \coloneq \{1,\dots,N\},
\end{equation}
where each value $z \in \mathrm{dom}(Z)$ corresponds to the hypothesis that player $z$ is the chameleon.

The observation at time $t$ is the publicly visible clue word produced by the current speaker:
\begin{equation}
o_t = w_t \in \mathcal{V},
\end{equation}
where $\mathcal{V}$ denotes the set of possible clue words, and the history $h_{t-1}$ contains all previously generated clues together with the public game information (category and candidate secret words).

For the likelihood term in~\eqref{eq:expected_update}, we again use the LLM's own word probabilities. Specifically, for each candidate chameleon identity $z$, we condition the speaking LLM on the public history $h_{t-1}$ together with the counterfactual assignment that $Z=z$ (\ie, that player $z$ is the chameleon), and extract the model's probability distribution $P_{\mathrm{LLM}}(\cdot \mid z, h_{t-1})$ over the possible clue words.

The Bayes-predicted log-likelihood ratio between two candidate chameleon identities $z$ and $z'$ is therefore
\begin{equation}
\Lambda_t(z,z')
=
\log
\frac{P_{\mathrm{LLM}}(o_t \mid z, h_{t-1})}
     {P_{\mathrm{LLM}}(o_t \mid z', h_{t-1})}.
\end{equation}

\subsection{Round-by-Round Progression of the Bayesian Coherence Coefficient}\label{app:round_by_round_bcc}
\begin{figure*}[t]
    \centering
    \includegraphics[width=\linewidth]{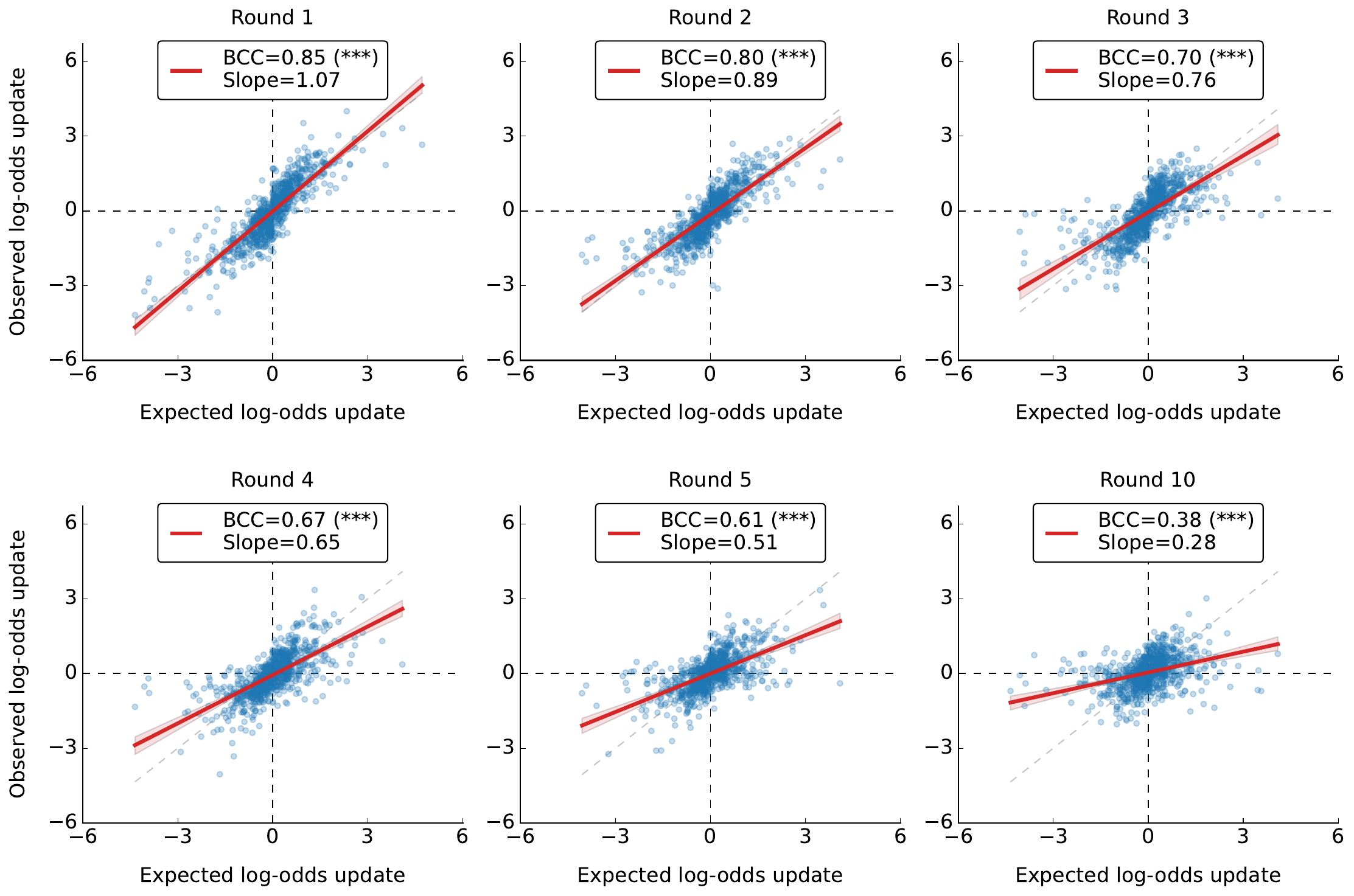}
    \caption{Round-by-round progression of the Bayesian Coherence Coefficient (BCC) of internal beliefs of Llama 3.1 70B about the opponent's type. Data collected from 800 repeated normal-form games. Asterisks in parentheses denote statistical significance (two-sided Wald test with t-distributed test statistic; $H_0$: zero correlation, $H_A$: non-zero correlation; * $p<0.05$; ** $p<0.01$; *** $p<0.001$).}
    \label{fig:bcc_scatter_internal_matrix}
\end{figure*}

The main text reports the Bayesian Coherence Coefficient (BCC) and the slope of its associated regression line as summary statistics per round. \autoref{fig:bcc_scatter_internal_matrix} shows the underlying data for the internal beliefs about the opponent's type in the repeated normal-form games with Llama 3.1 70B, plotting the observed log-odds update against the Bayes-predicted update for each round separately. Early rounds lie close to the identity line, with both a high correlation and a slope near one, which indicates updates of approximately correct direction and magnitude. As the interaction proceeds, the point cloud rotates toward the horizontal and the fitted slope falls below one, so the model increasingly under-reacts to new evidence rather than updating in the wrong direction.

\clearpage

\begin{textbox*}[t]
\begin{promptbox}{Prompt: Generalized Kuhn Poker}
You are an expert poker player playing Generalized Kuhn Poker. You must analyze the situation and decide on your next action.

\vspace{\the\baselineskip}

**Game Rules:**

* The game has 3 players, and the deck consists of 20 cards, numbered 1 to 20.

* All players start with a stack of 100 chips.

* Each player is dealt one secret card. The player with the highest card at the end wins the pot.

* The game proceeds in up to 3 rounds of betting.

* The actions are ``check'' or ``bet''. If a player bets, subsequent players can ``fold'' or ``call''.

* Allowed bet sizes are [1, 3, 5, 10, 15, 20, 50] chips, and players cannot bet more than their remaining stack.

\vspace{\the\baselineskip}

**Current Situation:**

It is your turn to act as Player 1. Your stack is 85 chips.

\vspace{\the\baselineskip}

**Your Card (Secret):**

* 4

\vspace{\the\baselineskip}

**Action History (in order of play):**

Round 1:

* Player 0: bet 15

* Player 1: call

* Player 2: call

Round 2:

* Player 0: check

* Player 1: bet 5

* Player 2: fold

* Player 0: call

Round 3:

* Player 0: check

\vspace{\the\baselineskip}

**Question:**

Based on the rules and the history, what is your next action? Your legal actions are: check, bet. Choose one action. If you bet, specify the amount. Respond with a single line with the chosen action and, if you bet, the amount. Do not say anything else. The format is:
chosen\_action amount\_if\_bet

\vspace{\the\baselineskip}

**Response:**
\end{promptbox}
\caption{The prompt used for asking LLMs for their next action in Generalized Kuhn Poker.}
\label{fig:kuhn_poker_prompt}
\end{textbox*}
\newpage

\begin{textbox*}[t]
\centering
\begin{promptbox}{Prompt: Generating new categories and secret words for The Chameleon game}
We are playing the game Chameleon but do not have enough game cards (categories and possible secret words).
Please suggest \verb|<NUM-OF-NEW-CARDS>| new categories and associated 16 secret words for the game. Here are the existing categories and secret words:
\vspace{1ex}

\verb|<EXISTING-CATEGORIES-AND-SECRET-WORDS>|

\vspace{1ex}
Please provide the \verb|<NUM-OF-NEW-CARDS>| new categories and secret words in a JSON format, where each category is a key and the value is a list of secret words. Do not say anything else, just return the JSON.
\end{promptbox}
\caption{The prompt used for asking GPT-4o for new categories and secret words for The Chameleon game.}
\label{fig:chameleon_new_game_cards_prompt}
\end{textbox*}
\newpage
\begin{textbox*}[t]
\centering
\begin{promptbox}{Prompt: The Chameleon}
\textbf{Game:} You will play a game of Chameleon with 3 other players. The aim in this game is to unmask the Chameleon without giving away the secret word. If you are the Chameleon, your mission is to blend in with the other players, avoid detection, and work out the secret word. One of the players will be randomly selected as the Chameleon. The Chameleon will know its identity but not the secret word. The other players will know the secret word but will not know who the Chameleon is. Each player (including the Chameleon) takes it in turns to say one word related to the secret word. Take a few moments to think of your word. If another player says the word you were going to say, you can repeat it. Choose wisely! If your word is too obvious, the Chameleon might catch on and figure out the secret word. If it's too cryptic, people might start to think that you're the Chameleon. After all players have said their word, you will vote for who you think the Chameleon is. The player with the most votes will be revealed as the Chameleon. If the Chameleon is correctly revealed, the Chameleon has one last chance to guess the secret word. If the Chameleon guesses correctly, the Chameleon wins. If not, the other players win. If the Chameleon is not correctly revealed, the Chameleon wins. The other players lose. Just say yes or no to confirm that you understand the rules of the game. Do not say anything else than yes or no.

\textbf{Player:} Yes

\textbf{Game:} The category for this game is Historical Periods. The possible secret words are elizabethan era, bronze age, renaissance, roman empire, industrial revolution, baroque period, iron age, ancient egypt, cold war, age of enlightenment, byzantine era, great depression, roaring twenties, stone age, victorian era, middle ages. Just say yes or no to confirm that you understand the category and possible secret words. Do not say anything else.

\textbf{Player:} Yes

\textbf{Game:} Let's start the game. You are player number 3 in the running order. Wait for the identities to be chosen. Just say your player number to confirm that you understood your running order. Do not say anything else other than the number.

\textbf{Player:} 3
\end{promptbox}
\caption{The base conversation history used in the gameplay of LLM players in the Chameleon game. ``Game:'' and ``Player:'' prefixes user prompts and LLM player's responses, respectively.}
\label{fig:chameleon_base_prompt}
\end{textbox*}
\newpage
\begin{textbox*}[t]
\centering
\begin{promptbox}{Prompt: The Chameleon -- Non-chameleon player}
\textbf{Game:} You are not the Chameleon. The secret word is: industrial revolution. Wait for your turn to say your word. To confirm that you understand your identity, say yes if you are the Chameleon. Say no if you are not the Chameleon. Do not say anything else.

\textbf{Player:} No

\textbf{Game:} Here are what the previous players said. Player 1 said Factory. Player 2 said Innovation. Player 3 said Steam. Player 4 said Coal.

\textbf{(Game:)} Based on what the other players have said so far, who do you think the Chameleon is? Only give the player number. Do not say anything else.

\textbf{(Player:)} 2
\end{promptbox}
\caption{The continuation of the base conversation history used in the gameplay of non-chameleon LLM players in the Chameleon game. ``Game:'' and ``Player:'' prefix user prompts and LLM player's responses, respectively. The final messages prefixed with ``(Game:)'' and ``(Player:)'' are used only for verbal probing.}
\label{fig:chameleon_nonchameleon_prompt}
\end{textbox*}
\newpage
\begin{textbox*}[t]
\centering
\begin{promptbox}{Prompt: The Chameleon -- Chameleon player}
\textbf{Game:} You are the Chameleon. Wait for your turn to say your word. To confirm that you understand your identity, say yes if you are the Chameleon. Say no if you are not the Chameleon. Do not say anything else.

\textbf{Player:} Yes

\textbf{Game:} Here are what the previous players said. Player 1 said Factory. Player 2 said Steam. It is now your turn to say your word. If your word is too obvious, the Chameleon might catch on and figure out the secret word. If it's too cryptic, people might start to think that you're the Chameleon. Make sure that your response is a single word. What is your word?

\textbf{Player:} Textiles
\end{promptbox}
\caption{The continuation of the base conversation history used in the gameplay of the chameleon LLM player in the Chameleon game. ``Game:'' and ``Player:'' prefix user prompts and LLM player's responses, respectively.}
\label{fig:chameleon_chameleon_prompt}
\end{textbox*}
\newpage
\begin{textbox*}[t]
\centering
\begin{promptbox}{Prompt: Repeated normal-form games with opponent types by strategy}

You are playing a game repeatedly with 1 other player, Player 1. There are 2 possible actions in each round (action A, action B). Players make their actions simultaneously, and the outcome is determined by their actions in the given round. All the players, including you, want to maximize their payoff (number of points).
\vspace{\the\baselineskip}

Here is the history of the game so far:

\hspace{2ex} Round 1: You played action B, and Player 1 played action B.

\hspace{2ex} Round 2: You played action A, and Player 1 played action A.

\hspace{2ex} \dots

\hspace{2ex} Round 8: You played action A, and Player 1 played action A.

\vspace{\the\baselineskip}

Note that Player 1 is one of two types, which determines their strategy and the payoffs of the game. In case they are of the first type, they play action A with probability 0.28 and action B with probability 0.72. In case they are of the second type, they play action A with probability 0.63 and action B with probability 0.37. However, you do not know which type they are.

\vspace{\the\baselineskip}

Given the history and the rules of the game, please provide your action for the next round. Respond with a single line that contains only the letter of your chosen action, do not say anything else.

\end{promptbox}
\caption{The prompt used in repeated normal-form games for specifying the possible opponent's types by strategy.}
\label{fig:matrix_opp1}
\end{textbox*}
\newpage
\begin{textbox*}[t]
\centering
\begin{promptbox}{Prompt: Repeated normal-form games with opponent types by strategy and round type}

You are playing a game repeatedly with 1 other player, Player 1. There are 2 possible actions in each round (action A, action B). Players make their actions simultaneously, and the outcome is determined by their actions in the given round. All the players, including you, want to maximize their payoff (number of points).
\vspace{\the\baselineskip}

Here is the history of the game so far:

\hspace{2ex} Round 1 (red): You played action B, and Player 1 played action B.

\hspace{2ex} Round 2 (blue): You played action A, and Player 1 played action A.

\hspace{2ex} \dots

\hspace{2ex} Round 8 (blue): You played action A, and Player 1 played action A.

\vspace{\the\baselineskip}

Note that Player 1 is one of two types, which determines their strategy and the payoffs of the game. In case they are of the first type, they play action A with probability 0.5 in blue rounds and 0.82 in red rounds, and action B with probability 0.5 in blue rounds and 0.18 in red rounds. In case they are of the second type, they play action A with probability 0.8 in blue rounds and 0.05 in red rounds, and action B with probability 0.2 in blue rounds and 0.95 in red rounds. However, you do not know which type they are.

\vspace{\the\baselineskip}

Given the history and the rules of the game, please provide your action for the next round. The next round will be red. Respond with a single line that contains only the letter of your chosen action, do not say anything else.

\end{promptbox}
\caption{The prompt used in repeated normal-form games for specifying the possible opponent's types by strategy and round type.}
\label{fig:matrix_opp2}
\end{textbox*}
\newpage
\begin{textbox*}[t]
\centering
\begin{promptbox}{Prompt: Repeated normal-form games with opponent types by payoff matrix}

You are playing a game repeatedly with 1 other player, Player 1. There are 2 possible actions in each round (action A, action B). Players make their actions simultaneously, and the outcome is determined by their actions in the given round. All the players, including you, want to maximize their payoff (number of points).
\vspace{\the\baselineskip}

Note that Player 1 is one of two types, which determines their strategy and the payoffs of the game. You do not know which type they are. If they are of the first type, the payoffs are as follows:

\hspace{2ex} If you play action A and the other player plays action A, you

\hspace{2ex} get 3.7 points and they get -3.7 points.

\hspace{2ex} If you play action A and the other player plays action B, you

\hspace{2ex} get 9.1 points and they get -9.1 points.

\hspace{2ex} If you play action B and the other player plays action A, you

\hspace{2ex} get 8.5 points and they get -8.5 points.

\hspace{2ex} If you play action B and the other player plays action B, you

\hspace{2ex} get 4.2 points and they get -4.2 points.

\vspace{\the\baselineskip}
  
If they are of the second type, the payoffs are as follows:

\hspace{2ex} If you play action A and the other player plays action A, you

\hspace{2ex} get 5.0 points and they get -5.0 points.

\hspace{2ex} If you play action A and the other player plays action B, you

\hspace{2ex} get 8.8 points and they get -8.8 points.

\hspace{2ex} If you play action B and the other player plays action A, you

\hspace{2ex} get 8.6 points and they get -8.6 points.

\hspace{2ex} If you play action B and the other player plays action B, you

\hspace{2ex} get 0.4 points and they get -0.4 points.

\vspace{\the\baselineskip}

Given the history and the rules of the game, please provide your action for the next round. Respond with a single line that contains only the letter of your chosen action, do not say anything else.

\end{promptbox}
\caption{The prompt used in repeated normal-form games for specifying the possible opponent's types by payoff matrix.}
\label{fig:matrix_opp3}
\end{textbox*}
\newpage
\begin{textbox*}[t]
\centering
\begin{promptbox}{Verbal probe prompt: Opponent's strategy}
You are playing a game repeatedly with 1 other player, Player 1. There are 2 possible actions in each round (action A, action B). Players make their actions simultaneously, and the outcome is determined by their actions in the given round. All the players, including you, want to maximize their payoff (number of points). Players may adapt their strategies as they play more rounds.
\vspace{\the\baselineskip}

Here are the rules of the game:

\hspace{2ex} If you play action A and the other player plays action A, you get 4.2 points and they get 6.5 points.

\hspace{2ex} If you play action A and the other player plays action B, you get 4.4 points and they get 8.9 points.

\hspace{2ex} If you play action B and the other player plays action A, you get 9.6 points and they get 3.8 points.

\hspace{2ex} If you play action B and the other player plays action B, you get 7.9 points and they get 5.3 points.
  
\vspace{\the\baselineskip}

Here is the history of the game so far:

\hspace{2ex} Round 1: You played action B, and Player 1 played action B. Your payoff for this round was 7.9 points.

\hspace{2ex} Round 2: You played action A, and Player 1 played action A. Your payoff for this round was 4.2 points.

\hspace{2ex} \dots

\hspace{2ex} Round 27: You played action A, and Player 1 played action A. Your payoff for this round was 4.2 points.

\vspace{\the\baselineskip}

Based on the history so far, what is the most likely policy (probability distribution over actions) of Player 1? Respond with a single line that contains only the Player 1's policy in JSON format, do not say anything else. The format is:

\verb|{"<action-letter>": <probability>, ...}|

\end{promptbox}
\caption{The prompt used for verbal probes of opponent's strategy in repeated normal-form games.}
\label{fig:matrix_vp}
\end{textbox*}
\begin{textbox*}[t]
\begin{promptbox}{Verbal probe prompt: Opponent's private card}
You are an expert poker player playing Generalized Kuhn Poker. You must analyze the situation and decide on your next action.

\vspace{\the\baselineskip}

**Game Rules:**

* The game has 3 players, and the deck consists of 20 cards, numbered 1 to 20.

* All players start with a stack of 100 chips.

* Each player is dealt one secret card. The player with the highest card at the end wins the pot.

* The game proceeds in up to 3 rounds of betting.

* The actions are ``check'' or ``bet''. If a player bets, subsequent players can ``fold'' or ``call''.

* Allowed bet sizes are [1, 3, 5, 10, 15, 20, 50] chips, and players cannot bet more than their remaining stack.

\vspace{\the\baselineskip}

**Current Situation:**

It is your turn to act as Player 1. Your stack is 85 chips.

\vspace{\the\baselineskip}

**Your Card (Secret):**

* 4

\vspace{\the\baselineskip}

**Action History (in order of play):**

Round 1:

* Player 0: bet 15

* Player 1: call

* Player 2: call

Round 2:

* Player 0: check

* Player 1: bet 5

* Player 2: fold

* Player 0: call

Round 3:

* Player 0: check

\vspace{\the\baselineskip}

**Question:**

Based on the rules and the history, which card is the Player 0 most likely to have? Respond with a single line that contains only the Player 0 card (1, 2, ..., or 20), do not say anything else.

\vspace{\the\baselineskip}

**Response:**
\end{promptbox}
\caption{The prompt used for verbal probes of opponent's private card in Generalized Kuhn Poker.}
\label{fig:kuhn_poker_vp}
\end{textbox*}
\newpage
\begin{textbox*}[t]
\centering
\begin{promptbox}{Verbal probe prompt: Secret word}
\textbf{Game:} You are the Chameleon. Wait for your turn to say your word. To confirm that you understand your identity, say yes if you are the Chameleon. Say no if you are not the Chameleon. Do not say anything else.

\textbf{Player:} Yes

\textbf{Game:} Here are what the players said: Player 1 said Factory. Player 2 said Steam.
Based on what the other players have said so far and based on the possible secret words, what do you think the secret word is? Remember, all players except you know the secret word. Say the exact secret word that you believe they are concealing. Do not say anything else.

\textbf{Player:} industrial revolution
\end{promptbox}
\caption{The continuation of the base conversation history used for verbal probes of the secret word in the Chameleon game. ``Game:'' and ``Player:'' prefix user prompts and LLM player's responses, respectively.}
\label{fig:chameleon_secret_word_vp}
\end{textbox*}
\newpage
\clearpage

\subsection{Additional Experimental Results}\label{app:additional_experimental_results}
This section reports additional experiments that test the robustness of the findings in the main text and extend them in several directions, including:
\begin{enumerate}
    \item Analysis of the effect of action labels and prompt order (\autoref{app:prompt_action_labels})
    \item Evaluation in games framed as real-life financial scenarios (\autoref{app:prompt_context})
    \item Measuring action optimality as a function of compositional complexity (\autoref{app:action_optimality})
    \item Validation of results in non-quantized and base language models, and analysis of the impact of model size (\autoref{app:quantization_post_training})
    \item Comparison with alternative internal probing techniques (\autoref{app:alternative_probing}) and against additional observation-feature baseline predictor (\autoref{app:observation_feature_baseline})
    \item Tracing the decodability of latent game information and information routing in models, layer by layer (\autoref{app:layerwise_tracing})
    \item Evaluation of frontier API-only LLMs, including reasoning models (\autoref{app:api_only_llms})
\end{enumerate}

\subsubsection{Effect of action labels and prompt order}\label{app:prompt_action_labels}
\begin{figure*}[t]
    \centering
    \includegraphics[width=0.85\linewidth]{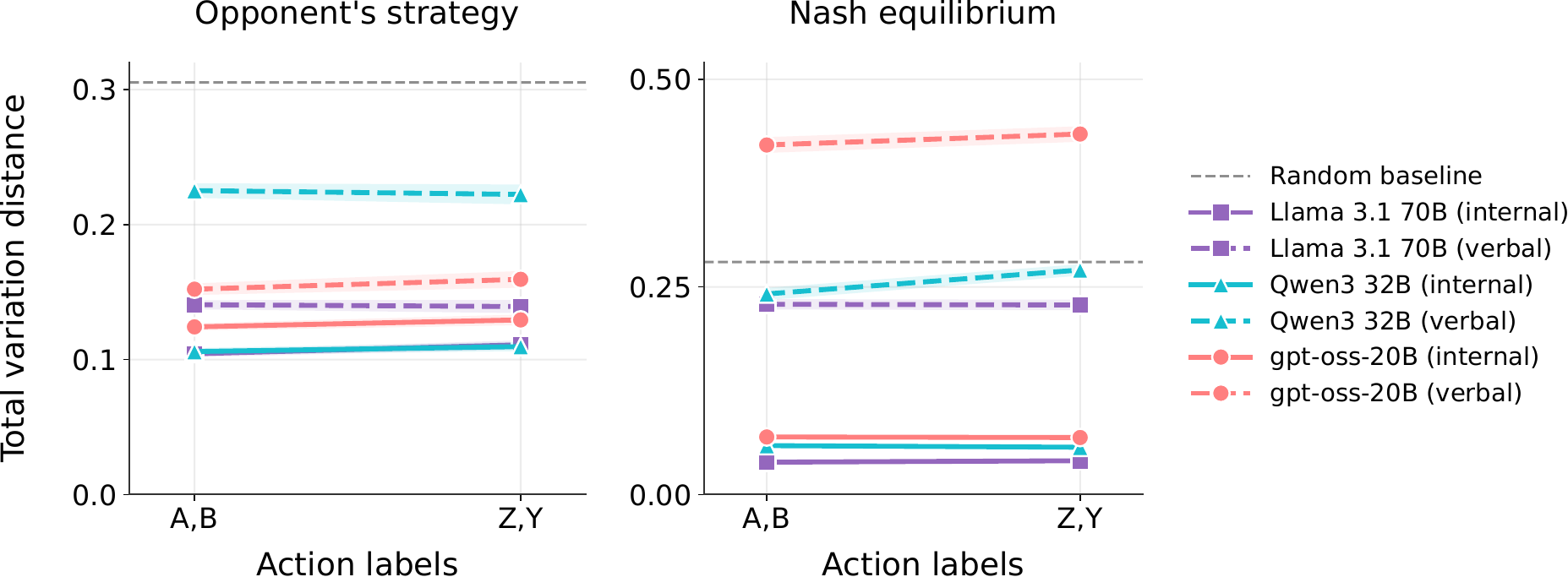}
    \vspace{6pt}
    \caption{Effect of action labels: Inference of opponent's strategy (left) and the mixed-strategy Nash equilibrium (right) in repeated normal-form games. Shown are the means and their standard errors obtained from 600 repeated normal-form games.}
    \label{fig:hidden_info_strategy_by_action_labels}
\end{figure*}

\begin{figure*}[t]
    \centering
    \includegraphics[width=\linewidth]{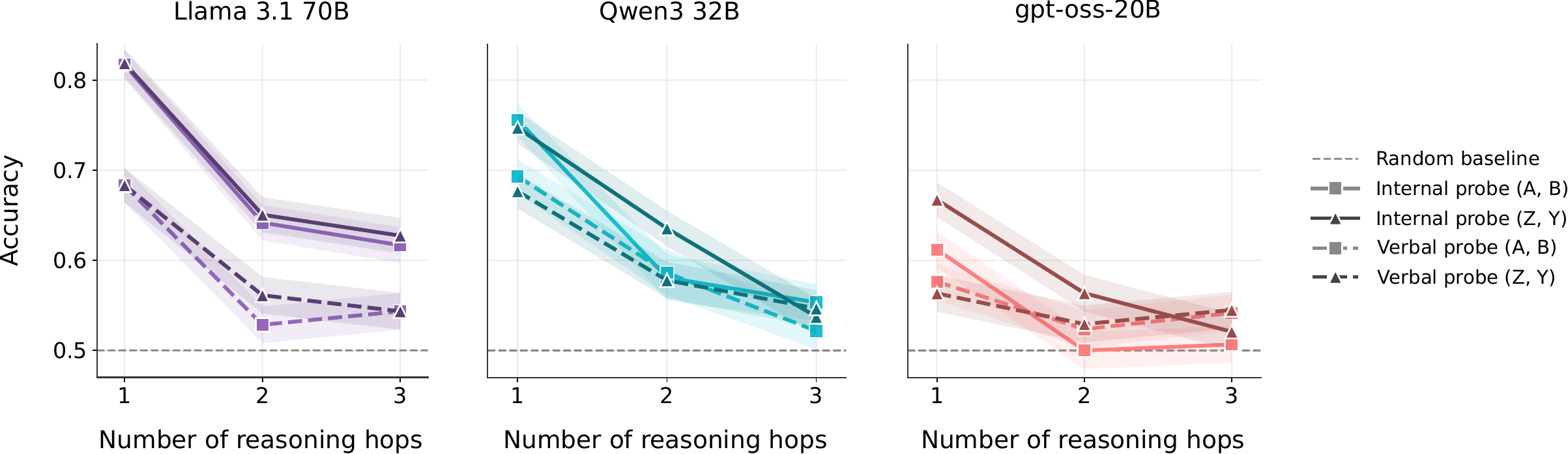}
    \vspace{-6pt}
    \caption{Inference of opponent's type in repeated normal-form games using internal and verbal probes. Number of reasoning hops corresponds to: \textbf{1:} Opponent's type by strategy. \textbf{2:} Opponent's type by strategy and round type. \textbf{3:} Opponent's type by payoff matrix. Letters in the legend parentheses denote the action labels. Shown are the means and their standard errors obtained from 600 repeated normal-form games.}
    \label{fig:hidden_info_types_by_action_labels}
\end{figure*}

The analyses in the main text use the action labels A and B, which raises the question of whether the reported effects depend on the identity or the order of the action tokens rather than on the underlying game quantities. To test this, we re-ran the repeated normal-form game experiments with permuted labels (B and A) and with alternative labels (Z and Y). \autoref{fig:hidden_info_strategy_by_action_labels} shows the inference of the opponent's strategy and the mixed-strategy Nash equilibrium, and \autoref{fig:hidden_info_types_by_action_labels} the inference of the opponent's type under these labelings. The internal and verbal probe accuracies are close across labelings, indicating that belief formation does not depend on the specific action tokens.

\begin{figure*}[h!]
    \centering
    \includegraphics[width=\linewidth]{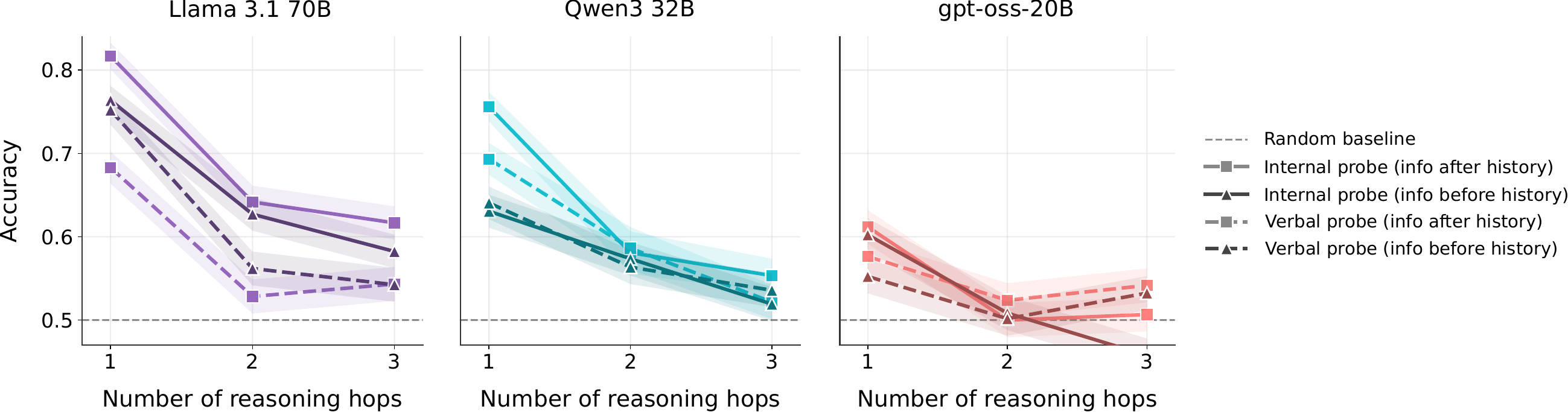}
    \vspace{-6pt}
    \caption{Effect of prompt order: Inference of opponent's type in repeated normal-form games with the side information (candidate opponent's strategies or payoff matrices) placed either before or after the history in the prompt. Number of reasoning hops corresponds to: \textbf{1:} Opponent's type by strategy. \textbf{2:} Opponent's type by strategy and round type. \textbf{3:} Opponent's type by payoff matrix. Shown are the means and their standard errors obtained from 600 repeated normal-form games.}
    \label{fig:hidden_info_opp_type_prompt_order}
\end{figure*}

\begin{figure*}[t]
    \centering
    \begin{subfigure}[t]{0.48\linewidth}
        \centering
        \includegraphics[width=\linewidth]{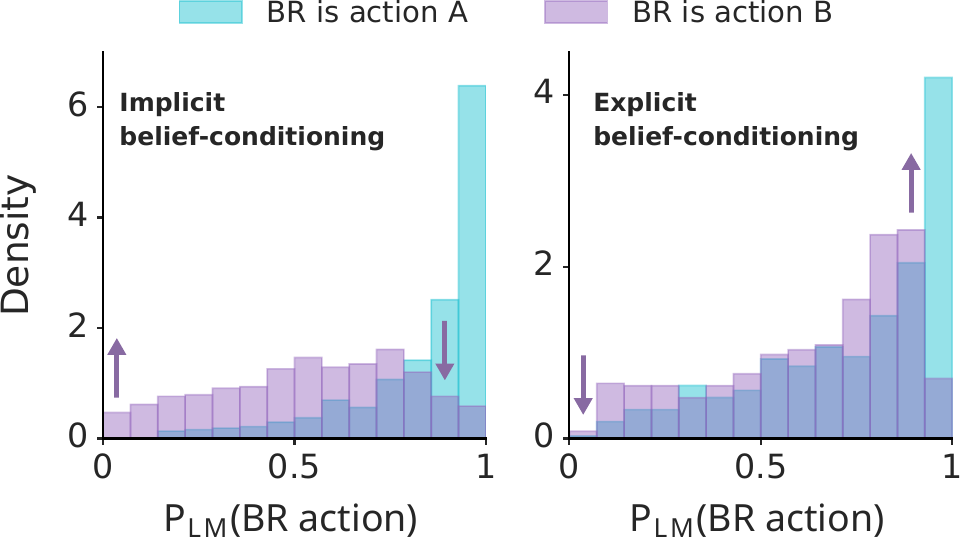}
        \caption{Effect of the action-letter bias on selecting the best response (BR) action by the LLM in games where the actions are switched (B, A). The plot on the right illustrates a scenario where verbalizing internal beliefs about the opponent's strategy partially mitigates this bias. Shown is the histogram of next-action probabilities (x-axis) assigned by the LLM in 1,000 repeated normal-form games.}
        \label{fig:br_bias_b_a_actions}
    \end{subfigure}\hfill
    \begin{subfigure}[t]{0.5\linewidth}
        \centering
        \includegraphics[width=\linewidth]{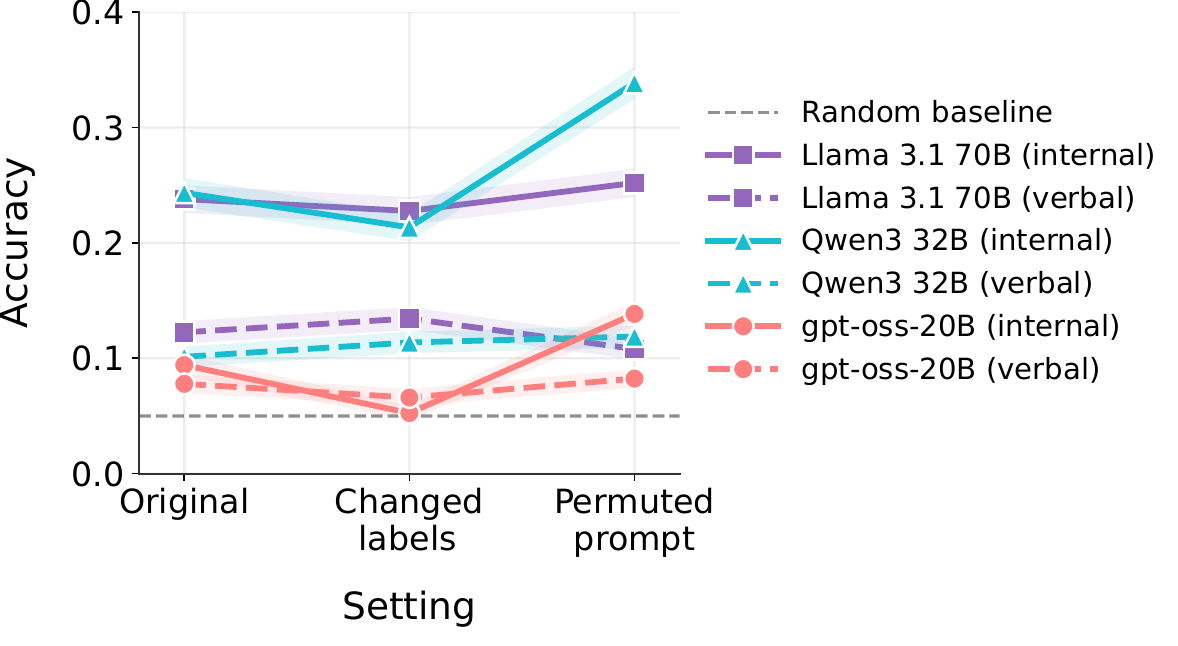}
        \caption{Effect of prompt order, action, and card labels: Inference of the opponent's secret card in Generalized Kuhn Poker. Shown are the means and their standard errors obtained from 600 game trials.}
        \label{fig:kuhn_poker_prompt_changes}
    \end{subfigure}
    \vspace{10pt}
    \caption{Effect of action labels and prompt order.}
\end{figure*}

Further, \autoref{fig:br_bias_b_a_actions} repeats the action-letter bias analysis of Figure 11 in the main text with the action order switched. The inflated probability mass stays on the action letter A under implicit belief conditioning, illustrating a bias toward specific letters. Similar to Figure 11, verbalizing internal beliefs about the opponent's strategy in the prompt mitigates this surface-level bias to a large extent. This experiment led us to state the bias as an action-letter bias rather than a positional bias toward the first option. The two readings make the same prediction when A is listed first, and they differ only when the labels are permuted, in which case the probability mass follows the letter and not the position.

We also test whether the results depend on where the side information about the opponent's type is placed in the prompt. \autoref{fig:hidden_info_opp_type_prompt_order} shows the inference of the opponent's type when the candidate opponent's strategies or payoff matrices are given either before or after the action history. The two orderings give similar probe accuracies, so the models do not rely on a fixed position of the side information in the prompt and are robust to this prompt-order choice.

Additionally, we repeat the Generalized Kuhn Poker experiments with increased bet sizes and changed card labels from numbers to letters (A to T instead of 1 to 20; \autoref{fig:kuhn_poker_prompt_labeling}) and with reordered blocks of game information in the prompt (\autoref{fig:kuhn_poker_prompt_permuted}). As shown in \autoref{fig:kuhn_poker_prompt_changes}, the inference of the opponent's secret card is again stable under both changes. Taken together, these experiments indicate that the reported observation-belief gap is a property of the underlying inference process rather than an artifact of a particular prompt layout or a particular choice of action tokens. The one exception is the action-letter bias itself, which by construction depends on the letters used to name the actions.

\subsubsection{Effect of contextual framing}\label{app:prompt_context}
The games in the main text are stated in abstract terms, which raises the question of whether the reported gaps also appear when the same strategic problem is embedded in a realistic scenario. To test this, we re-ran the experiments on belief formation and belief updating under a realistic framing of the repeated normal-form games and of The Chameleon.
 
For the repeated normal-form games, we place the LLM in a financial decision-making scenario of a CEO, following the framing and the prompts of \cite{lore2024}, who used it to study LLMs in realistic settings modeled as normal-form games. The payoff structure and the action history are the same as in the standard framing, and only the wording of the scenario changes (prompt provided in \autoref{fig:matrix_prompt_realistic}). For The Chameleon, we rephrase the game as a special-situations desk of four analysts who share a channel used to transmit the company's sensitive information. One of the four analysts on the currently active channel is an operative placed by a rival fund. The operative holds a codebook, that is, the full list of candidate passphrases, but not the active passphrase, which rotates per session and is hand-delivered. This preserves the information structure of the original game, where the chameleon knows the candidate words but not the secret word. Prompt for this realistic framing of The Chameleon is shown in \autoref{fig:chameleon_prompt_realistic}. We leave out a realistic rewriting of Generalized Kuhn Poker because the game already describes a realistic poker setting.

\begin{figure*}[t]
    \centering
    \includegraphics[width=0.8\linewidth]{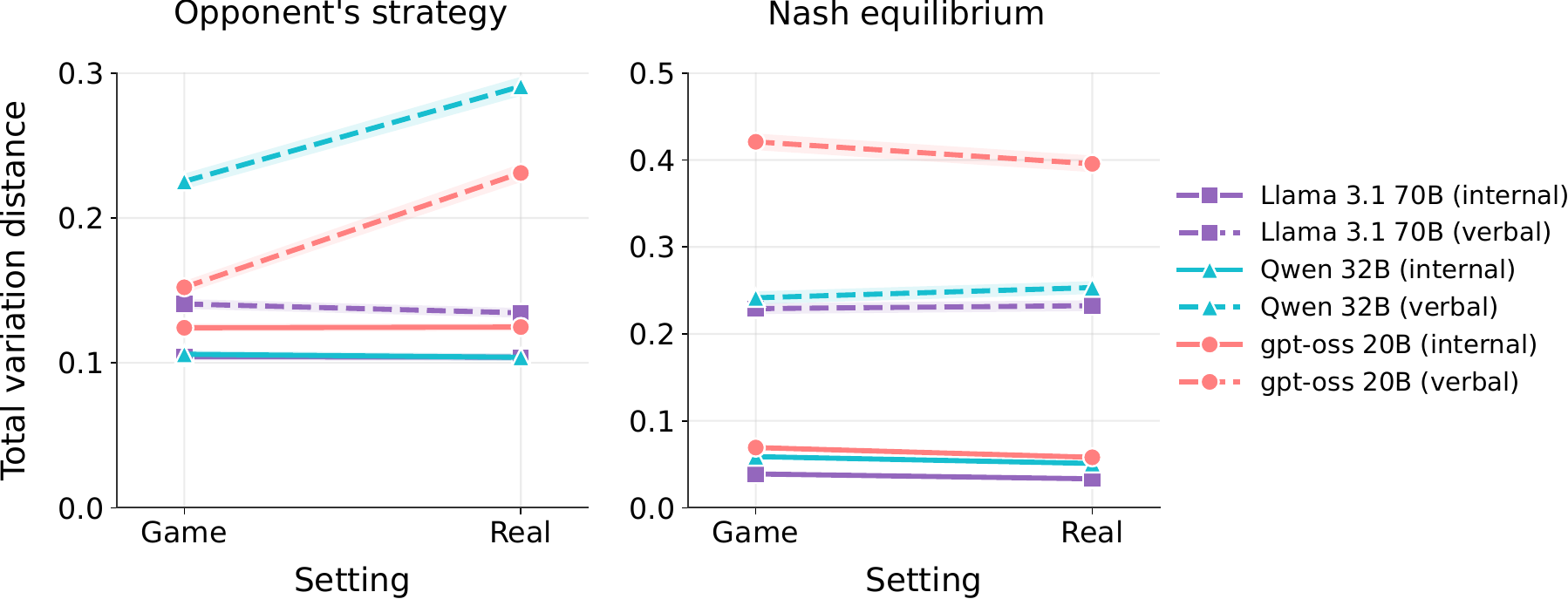}
    \vspace{4pt}
    \caption{Effect of context: Inference of opponent's strategy (left) and the mixed-strategy Nash equilibrium (right) in repeated normal-form games under the standard ``Game'' framing and in a ``Real'' framing of a realistic financial decision-making scenario of a CEO, following the prompts of \cite{lore2024}. Shown are the means and their standard errors obtained from 1,600 repeated normal-form games.}
    \label{fig:hidden_info_strategy_ceo}
\end{figure*}

\begin{figure*}[h!]
    \centering
    \includegraphics[width=\linewidth]{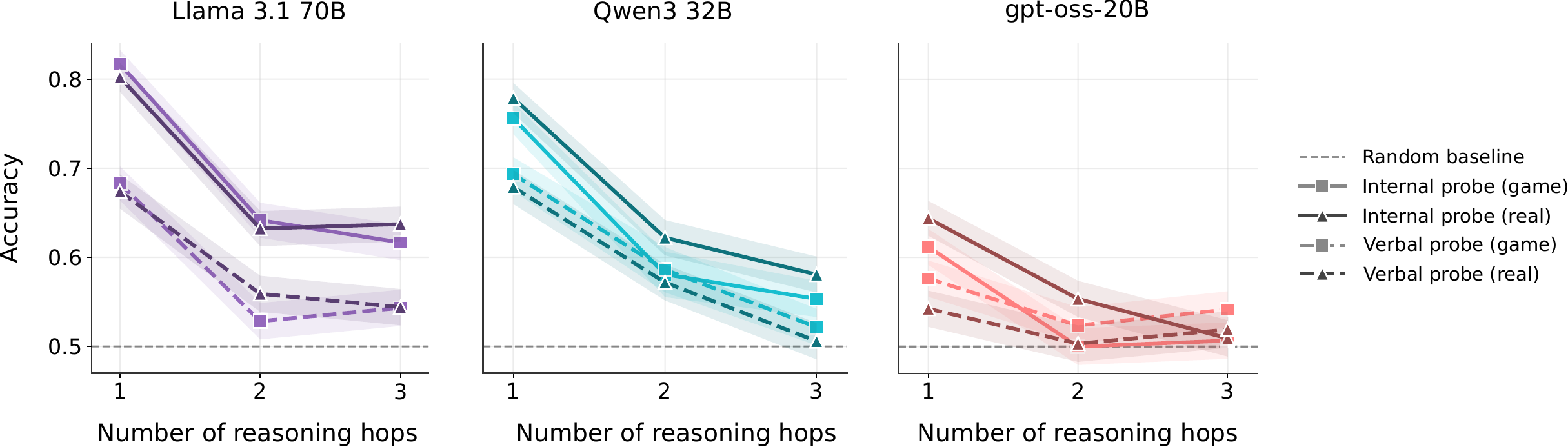}
    \vspace{-4pt}
    \caption{Effect of context: Inference of the opponent's type in repeated normal-form games under the standard ``Game'' framing and in a ``Real'' framing of a realistic financial decision-making scenario of a CEO, following the prompts of \cite{lore2024}. Number of reasoning hops corresponds to: \textbf{1:} Opponent's type by strategy. \textbf{2:} Opponent's type by strategy and round type. \textbf{3:} Opponent's type by payoff matrix. Shown are the means obtained from 600 repeated normal-form games.}
    \label{fig:hidden_info_opp_type_ceo}
\end{figure*}

\begin{figure*}[ht!]
    \centering
    \includegraphics[width=\linewidth]{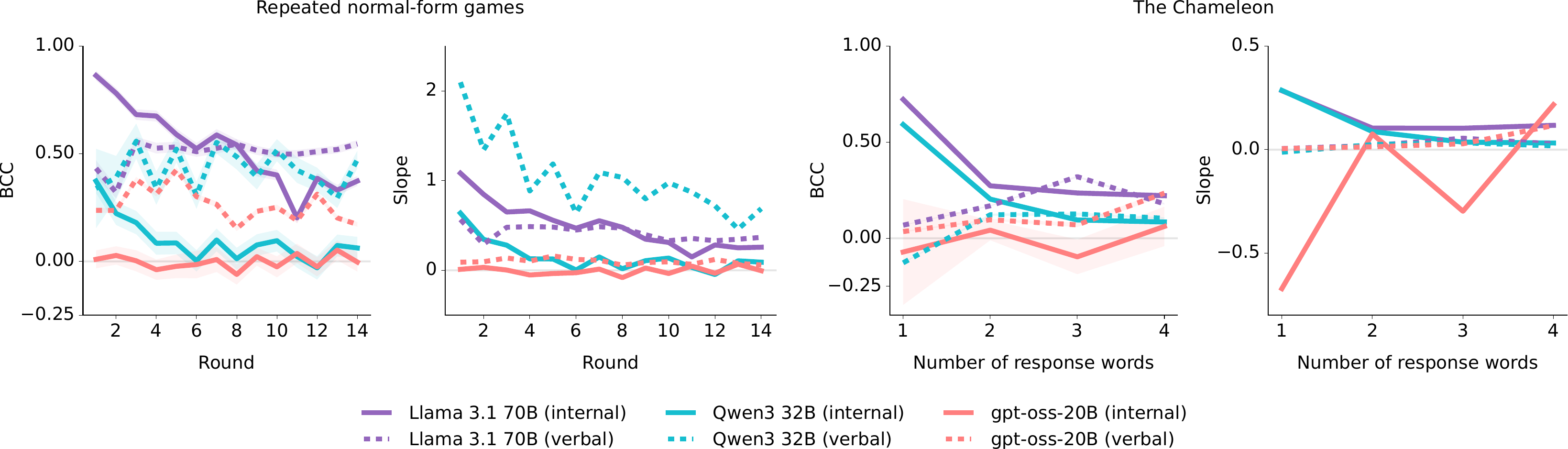}
    \vspace{-6pt}
    \caption{Progression of the Bayesian Coherence Coefficient (BCC) and the slope of its associated regression line with increasing length of interaction. Shown are the results from 1,000 repeated normal-form games framed as a realistic financial decision-making scenario of a CEO, following the prompts of \cite{lore2024} (left); and from 1,168 game trials of The Chameleon framed as a special-situations desk of four analysts (right). BCC is measured with respect to internal and verbal beliefs about the opponent's type (repeated normal-form games), and about the chameleon identity (The Chameleon).}
    \label{fig:bcc_ceo}
\end{figure*}

\begin{figure*}[ht!]
    \centering
    \includegraphics[width=0.82\linewidth]{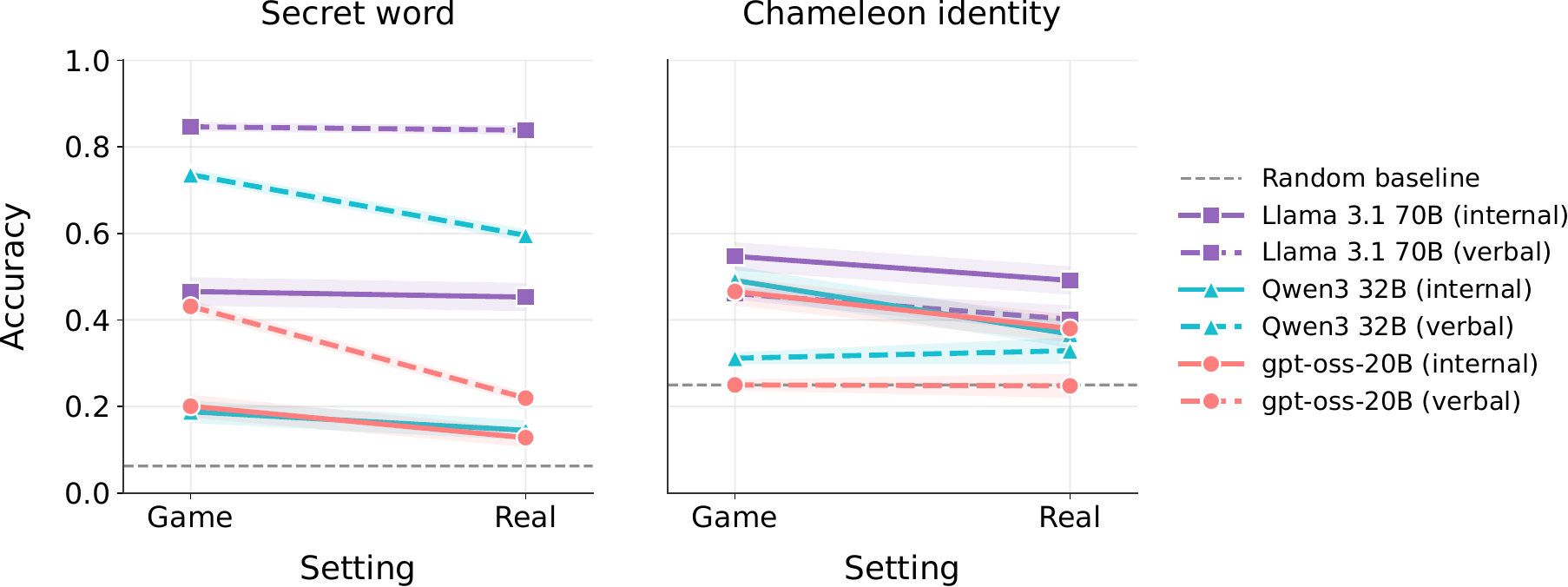}
    \vspace{4pt}
    \caption{Effect of context: Inference of the secret word (left) and the chameleon identity (right) in the Chameleon game under the standard ``Game'' framing and under a ``Real'' framing of a realistic special-situations desk of four analysts with access to a shared channel used to transmit the company's sensitive information. Shown are the means obtained from 230 Chameleon game trials.}
    \label{fig:hidden_info_chameleon_real}
\end{figure*}
 
The results under the realistic framing are shown in \autoref{fig:hidden_info_strategy_ceo} for the inference of the opponent's strategy and the mixed-strategy Nash equilibrium, in \autoref{fig:hidden_info_opp_type_ceo} for the inference of the opponent's type, in \autoref{fig:bcc_ceo} for Bayesian belief updating, and in \autoref{fig:hidden_info_chameleon_real} for the inference of the secret word and of the chameleon identity. The qualitative findings do not change. Internal beliefs remain more accurate than verbal ones, with the same exception of the secret word in The Chameleon that we report in the main text, belief accuracy still declines with the number of required reasoning hops, and Bayesian coherence still deteriorates over the course of the interaction.
 
We also do not observe a consistent degradation of the probes' accuracies relative to the standard framing. The main exception is that the smaller models perform substantially worse at inferring the opponent's strategy in the repeated normal-form games and at inferring the secret word in The Chameleon. A plausible reason is that the realistic prompts carry additional wording and storyline that is not relevant for the inference, which leaves less capacity for the latent game quantities in the smaller models. Overall, we conclude that the two gaps we report are not artifacts of the abstract presentation of the games.

\begin{figure*}[t]
    \centering
    \includegraphics[width=\linewidth]{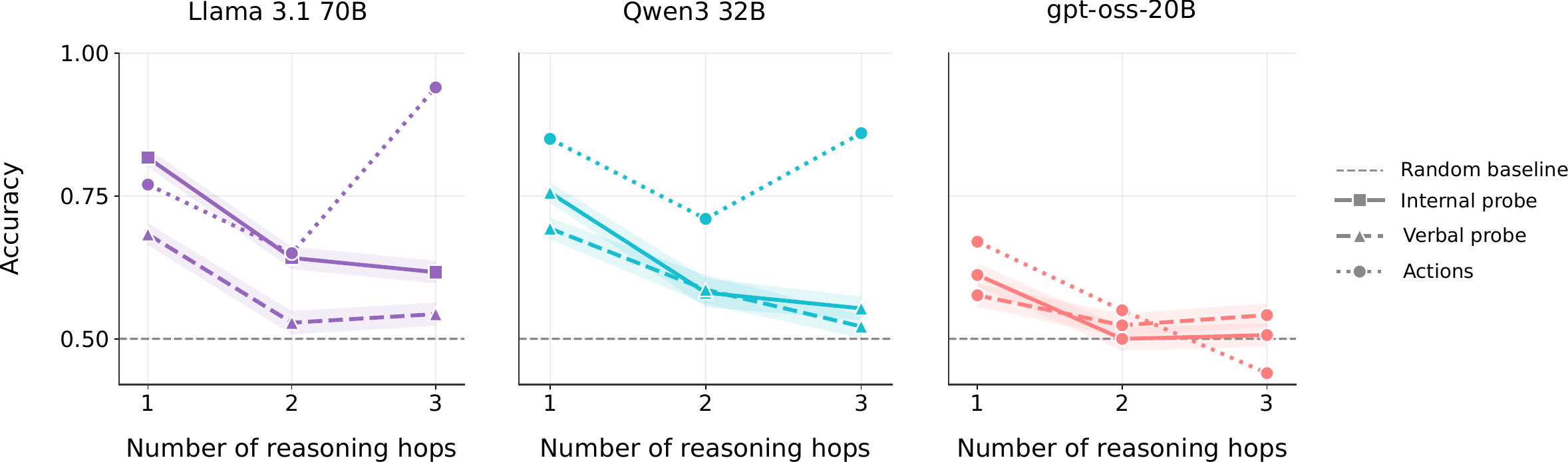}
    \vspace{-6pt}
    \caption{Accuracy of actions with respect to the best response and inference of opponent's type in repeated normal-form games. Number of reasoning hops corresponds to: \textbf{1:} Opponent's type by strategy. \textbf{2:} Opponent's type by strategy and round type. \textbf{3:} Opponent's type by payoff matrix. Shown are the means and their standard errors obtained from 600 repeated normal-form games.}
    \label{fig:hidden_info_opp_type_br_rate}
\end{figure*}

\subsubsection{Action optimality versus inferential difficulty}\label{app:action_optimality}
The experiment in the main text measures how well the opponent's type can be recovered from internal and verbal probes as the inferential difficulty grows, but it does not measure how well the models play under the same manipulation. We therefore report the optimality of the selected actions across the one-hop, two-hop, and three-hop conditions. We measure action optimality as the rate at which the LLM selects the best response with respect to the ground-truth opponent's type. We note that this experiment differs from the corresponding experiment in the main text because here we provide the LLMs with the payoff matrices in all three conditions. In the main text, the payoff matrices are given only in the three-hop setting, so that the models have to rely on the stated strategies in the one-hop and two-hop settings to infer the opponent's type.

As \autoref{fig:hidden_info_opp_type_br_rate} shows, action optimality follows a different trajectory than the accuracy of belief inference. The best-response rate declines from one hop to two hops, which tracks the decline of the internal and verbal belief accuracies, but it then rises sharply at three hops for the larger and more capable models, reaching 0.94 for Llama 3.1 70B and 0.86 for Qwen3 32B. In these cases, the action optimality in the three-hop setting is above its one-hop value, while the decodability of the opponent's type is at its lowest.
 
A likely explanation is the composition of the prompt rather than the inferential difficulty. The three conditions differ not only in the nominal number of hops but also in how much policy-relevant information the prompt states. The one-hop prompt states two candidate strategies and the two associated candidate payoff matrices. The two-hop prompt states four candidate strategies, that is, two strategies for each of the two round types, together with the round-type labels and the candidate payoff matrices. The three-hop prompt states only the two candidate payoff matrices, from which the equilibrium strategies can be derived. Action optimality is thus highest where the prompt carries the least explicit information about strategies and lowest in the two-hop condition, which carries the most. In other words, the ordering of action optimality runs opposite to the amount of strategic side information given in the prompt. One potential explanation is that in the three-hop condition the models' actions are driven directly by the observed action history and the stated payoff matrices, without an intermediate representation of the opponent's type, which would account for action optimality being highest where type decodability is lowest. A second, not mutually exclusive possibility is that the additional candidate strategies stated in the one- and two-hop prompts give action selection more opportunities to be misdirected. We offer both as hypotheses rather than demonstrations, and regard the result itself as an additional finding: supplying more policy-relevant side information need not improve the strategic decision-making of LLMs.

\subsubsection{Effect of quantization, post-training, and scale}\label{app:quantization_post_training}
Our analyses rely on the geometry of the hidden states, on linear decodability, and on activation steering, so it is important to check that the reported effects are not artifacts of the quantization we use for the larger models. It is also possible that instruction tuning shifts the effective objective of the models away from strict payoff maximization or accurate inference of latent information. To address both concerns, we re-ran the belief formation experiment with Qwen3 8B, Qwen3 14B, and Qwen3 32B in both their quantized and their full-precision variants, and with the base variants of the Qwen3 models that have not undergone instruction tuning. We restrict this ablation to a single model family because of the computational cost of running the full-precision variants of the larger models.
 
\begin{figure*}[t]
    \centering
    \includegraphics[width=0.84\linewidth]{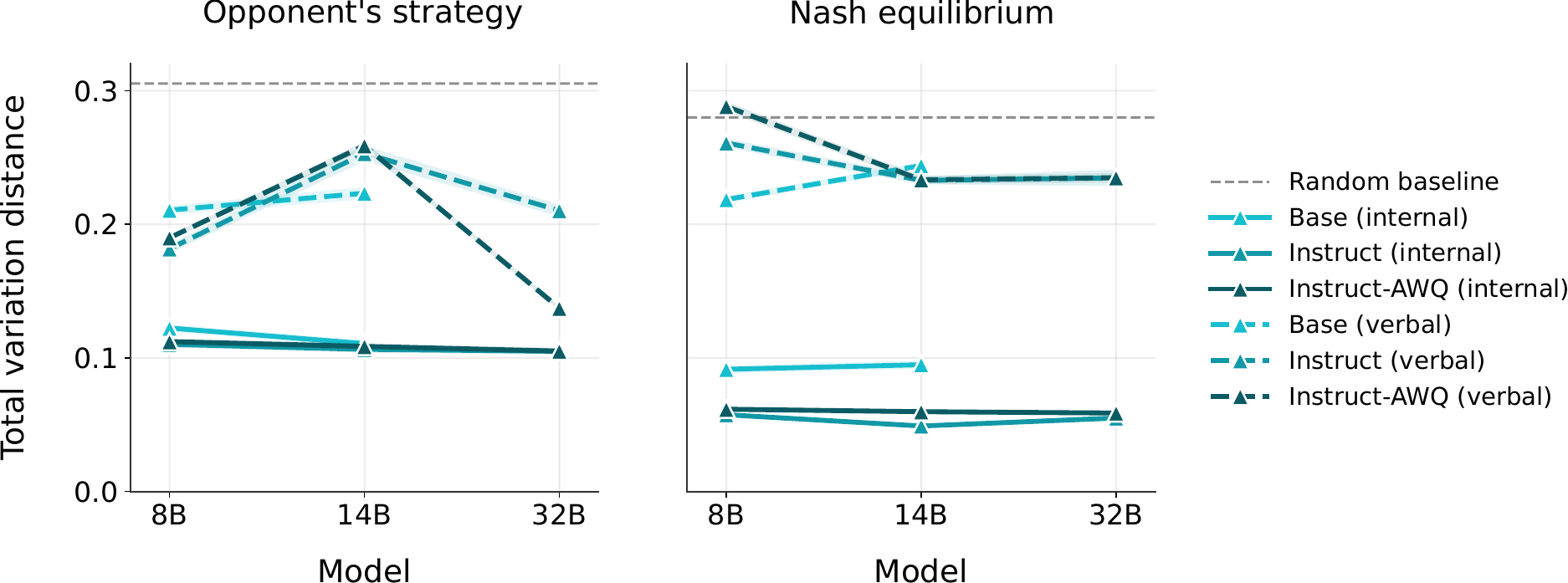}
    \caption{Effect of quantization and post-training: Inference of opponent's strategy (left) and the mixed-strategy Nash equilibrium (right) in repeated normal-form games using internal and verbal probes. Shown are the means and their standard errors obtained from 1,600 (left) and 800 (right) repeated normal-form games. The base model of Qwen3 32B was not released and is therefore omitted from the analysis.}
    \label{fig:hidden_info_strategy_and_zsmne_by_post_and_quant}
\end{figure*}

\begin{figure*}[t]
    \centering
    \includegraphics[width=\linewidth]{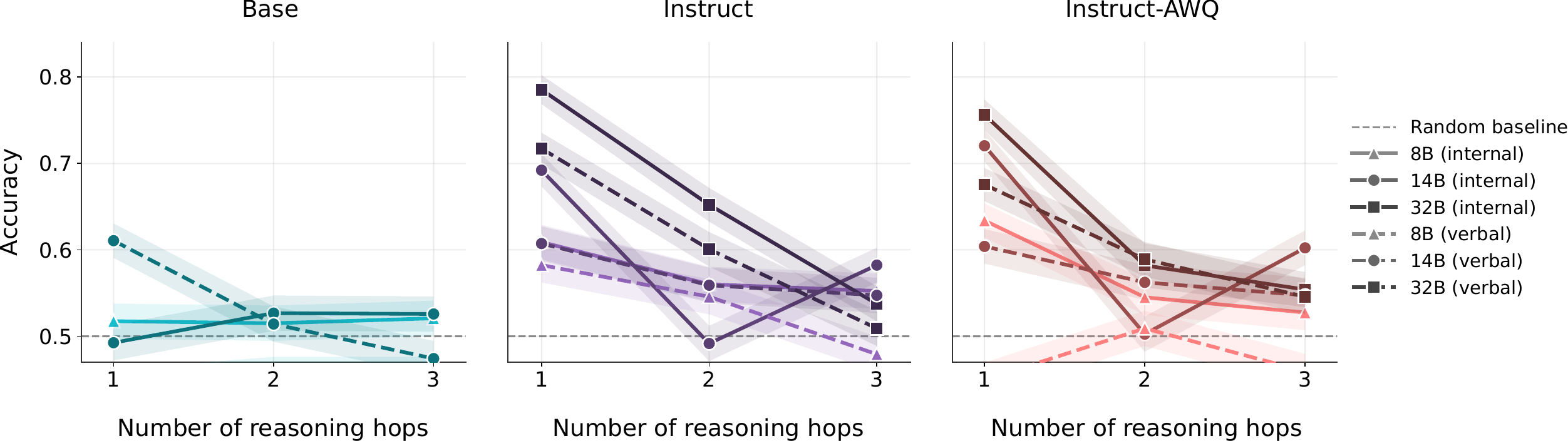}
    \caption{Effect of quantization and post-training on Qwen3 LLMs: Inference of opponent's type in repeated normal-form games using internal and verbal probes. Number of reasoning hops corresponds to: \textbf{1:} Opponent's type by strategy. \textbf{2:} Opponent's type by strategy and round type. \textbf{3:} Opponent's type by payoff matrix. Shown are the means and their standard errors obtained from 1,600 repeated normal-form games. The base model of Qwen3 32B was not released and is therefore omitted from the analysis.}
    \label{fig:hidden_info_opp_type_by_post_and_quant}
\end{figure*}

\autoref{fig:hidden_info_strategy_and_zsmne_by_post_and_quant} reports the inference of the opponent's strategy and of the mixed-strategy Nash equilibrium, and \autoref{fig:hidden_info_opp_type_by_post_and_quant} shows the inference of the opponent's type under these variants. The full-precision results match the quantized results qualitatively, and the quantitative differences between them are small, so the findings reported in the main text are not artifacts of quantization. The base models show the same qualitative patterns as their instruction-tuned counterparts. In the majority of cases, internal beliefs remain more accurate than verbal ones and accuracy still declines with the number of reasoning hops. The gap we identify is therefore not primarily a product of post-training or quantization.

\begin{figure}[h!]
    \centering
    \includegraphics[width=\linewidth]{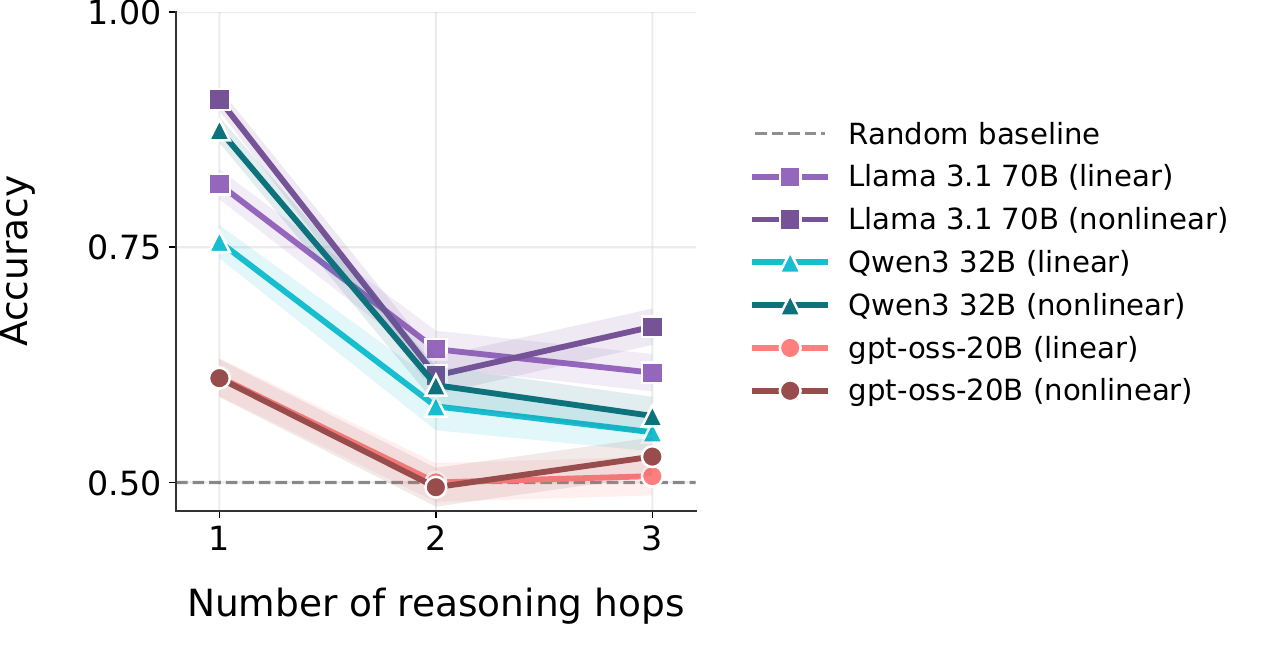}
    \caption{Inference of opponent's type using linear vs. nonlinear two-hidden-layer internal probe in repeated normal-form games. Number of reasoning hops corresponds to: \textbf{1:} Opponent's type by strategy. \textbf{2:} Opponent's type by strategy and round type. \textbf{3:} Opponent's type by payoff matrix. Shown are the means and their standard errors obtained from 800 repeated normal-form games. Internal and verbal correspond to internal and verbal probes, respectively.}
    \label{fig:hidden_info_nonlinear}
\end{figure}

Model scale within the Qwen3 family has a smaller effect than one might expect. The largest 32B model performs best on average across the internal and verbal probing tasks, but the improvement is not monotone in model size, and the 8B and 14B variants trade places depending on the target variable.

\subsubsection{Alternative probing techniques}\label{app:alternative_probing}
We restrict the internal probes in the main text to linear models because the goal is to quantify the inferential capability that is already present in the model rather than to add expressive power to it. More expressive probes risk learning the target function themselves instead of reading out what the model has already computed~\cite{hewitt2019, voita2020information}. Linear accessibility is also the decision-relevant notion in our setting. Up to layer normalization, the components of a transformer read the residual stream through linear projections, and the next-token logits are a linear function of the final hidden state, so at the action-selection step information can influence the action only to the extent that it is linearly exposed.

This restriction, however, raises a question about the interpretation of the multi-hop results. A decline in the accuracy of a linear readout could reflect a genuine failure of composed inference, or it could reflect that the required representation becomes more distributed and less linearly accessible as the task becomes more compositional. To separate these two interpretations, we repeat the experiment of inferring the opponent's type with a nonlinear internal probe, namely a feedforward network with two hidden layers, and compare it against the linear probe across the three hop conditions.
 
As shown in \autoref{fig:hidden_info_nonlinear}, the nonlinear probe declines with hop count in the same way as the linear probe, and the drop from one hop to two hops is again the largest. The degradation is therefore not fully an artifact of the linear restriction. Two further observations support this interpretation. First, the target variable and the probe architecture are identical across the three conditions, so an explanation based on probe inadequacy would have to apply specifically to the composed computation while the structurally identical one-hop target stays decodable. Second, the verbal probe, which reads the target out through the full nonlinear forward pass of the model, declines in parallel.

\begin{figure*}[t]
    \centering
    \includegraphics[width=0.82\linewidth]{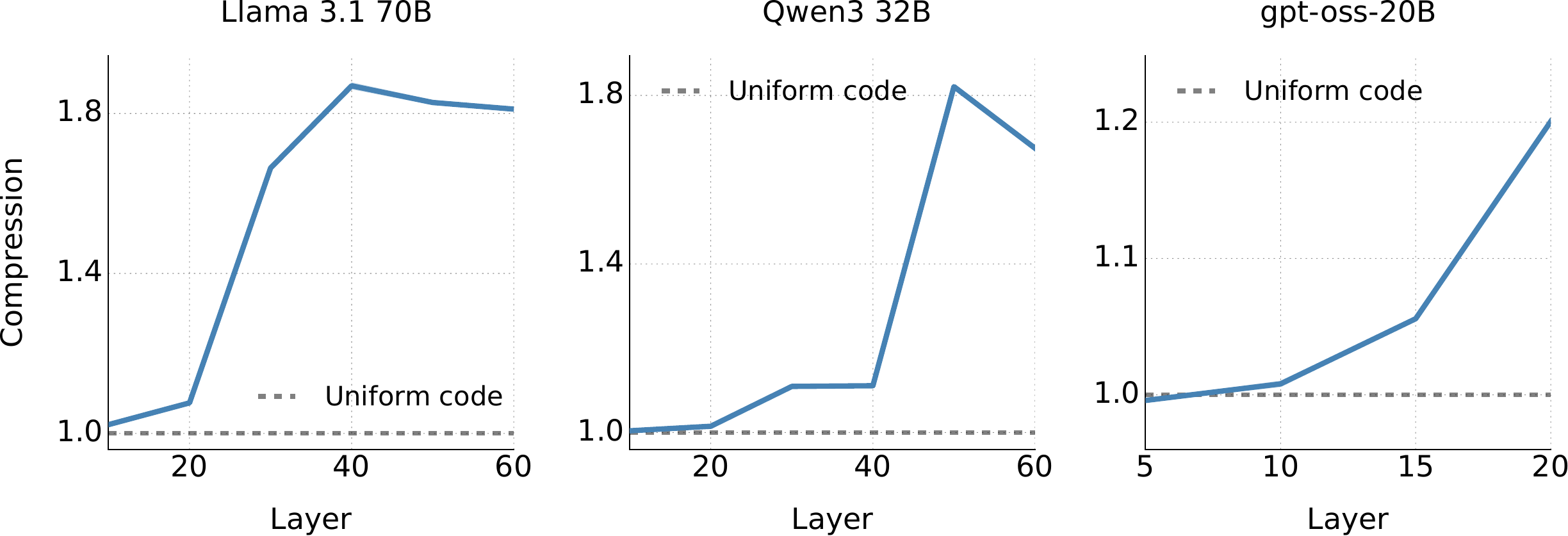}
    \vspace{4pt}
    \caption{Minimum description length (MDL) probing of the opponent's type (specified by the candidate strategies; one reasoning hop) in repeated normal-form games, shown per layer for each model. Compression is the ratio of the uniform codelength to the online (prequential) codelength; the dashed line at 1.0 marks the uniform code, \ie, no saving over transmitting the labels directly. Higher is better. Computed on all 15 rounds in 500 game trials.}
    \label{fig:mdl_probing}
\end{figure*}

A potential concern might be that the accuracy of a probe may reflect what the probe itself learns rather than what the representation already exposes. To address it, we repeated the opponent-type probing analysis in the repeated normal-form games using minimum description length (MDL) probing~\cite{voita2020information}. Rather than reporting the performance of a probe tuned on a fixed training set, MDL measures the number of bits needed to transmit the ground-truth labels to a receiver who already holds the hidden states.

We use online (prequential) coding: the training data is partitioned into blocks of increasing size, the first block is encoded with a uniform code over the label set, and each subsequent block is encoded under the predictive distribution of a probe trained only on the preceding blocks, so that its codelength is the out-of-sample (generalization) cross-entropy of the probe in bits. Because every probe is retrained from the data that has already been transmitted, the probe itself never enters the message. The resulting codelength is therefore a joint measure of how accurately and how cheaply the latent variable can be read out. A probe that reaches high accuracy only after seeing most of the training set pays for that in the bits spent on the early blocks. We report compression, which is the ratio of the uniform codelength to the online codelength. A value of 1.0 indicates that the representation gives no saving over transmitting the labels directly, and higher values indicate that the opponent's type is more extractable from the hidden states.
 
As shown in \autoref{fig:mdl_probing}, compression stays at the level of the uniform code in the early layers of all three models and rises well above it in the later-middle layers. It peaks at 1.85 for Llama 3.1 70B at layer 40, at 1.82 for Qwen3 32B at layer 50, and at 1.20 for gpt-oss-20B at layer 20. Both the ordering of the models and the location of the peak layers agree with the accuracy-based results reported in the main text and in \autoref{tab:appendix:internal_probe_hyperparameters}. That compression exceeds 1 by a substantial margin indicates that the probes recover information that is genuinely present in the representations rather than fitting the target themselves. The conclusions are therefore consistent across probing methods.

\subsubsection{Observation-feature baseline predictor}\label{app:observation_feature_baseline}
An internal probe could in principle recover a latent variable by reconstructing it from the individual observations that the model has encoded, rather than by reading out a representation of the variable that the model itself computed. To test which of the two is the case, we compare the internal probe against a baseline predictor that is trained on the observable features of the game instead of on the hidden states of the LLM. We use the repeated normal-form games and Llama 3.1 70B as a representative setting.
 
The baseline predicts the same latent variables as the internal probe, namely the opponent's strategy, the mixed-strategy Nash equilibrium, and the opponent's type. Its inputs are formed by concatenating (i) the payoffs and the one-hot encoded actions from all previous rounds, (ii) the flattened payoff matrix, (iii) the current round number, and, when the target is the opponent's type, (iv) the candidate opponent's strategies or payoff matrices. The baseline is linear and is trained, tuned, and evaluated on the same splits as the internal probe.
 
\begin{figure*}[t]
    \centering
    \includegraphics[width=\linewidth]{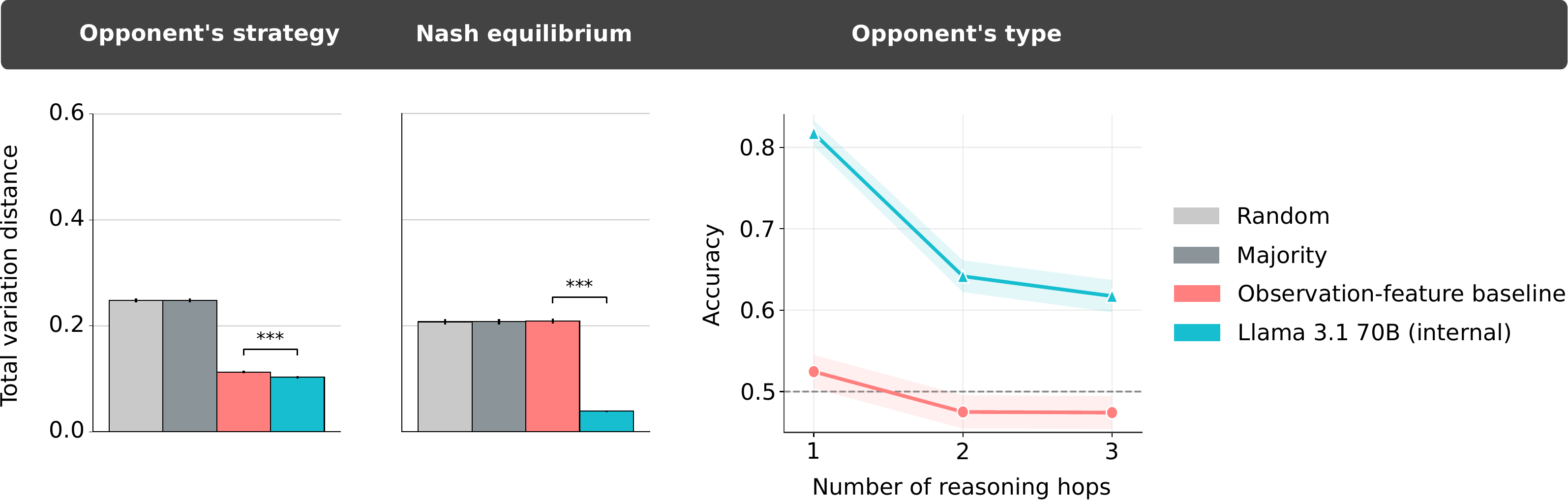}
    \vspace{-6pt}
    \caption{Comparison between observation-feature baseline and internal probes in inferring latent information in repeated normal-form games. Number of reasoning hops corresponds to: \textbf{1:} Opponent's type by strategy. \textbf{2:} Opponent's type by strategy and round type. \textbf{3:} Opponent's type by payoff matrix. Shown are the means and their standard errors. Asterisks denote statistical significance (one-sided Wilcoxon signed-rank test; * $p<0.05$; ** $p<0.01$; *** $p<0.001$). Total variation distance: lower is better; Accuracy: higher is better.}
    \label{fig:obs_feature_baseline}
\end{figure*}

For the opponent's strategy, the baseline performs similarly to the internal probe (\autoref{fig:obs_feature_baseline}), although still worse. This is expected given that the empirical action frequency of the opponent is a simple function of the observed history and is therefore easy for a linear predictor to compute. For the Nash equilibrium and for the opponent's type, in contrast, the baseline remains at the level of random chance while the internal probe stays well above it. Both of these variables are nonlinear functions of the stated game information, so a linear map applied to the raw observations cannot recover them. This gives evidence that the hidden states carry a transformed representation of these quantities rather than a trivial re-encoding of the observations, and that the LLM itself performs the transformation from the observations to the latent state.

\begin{figure*}[h!]
    \centering
    \includegraphics[width=\linewidth]{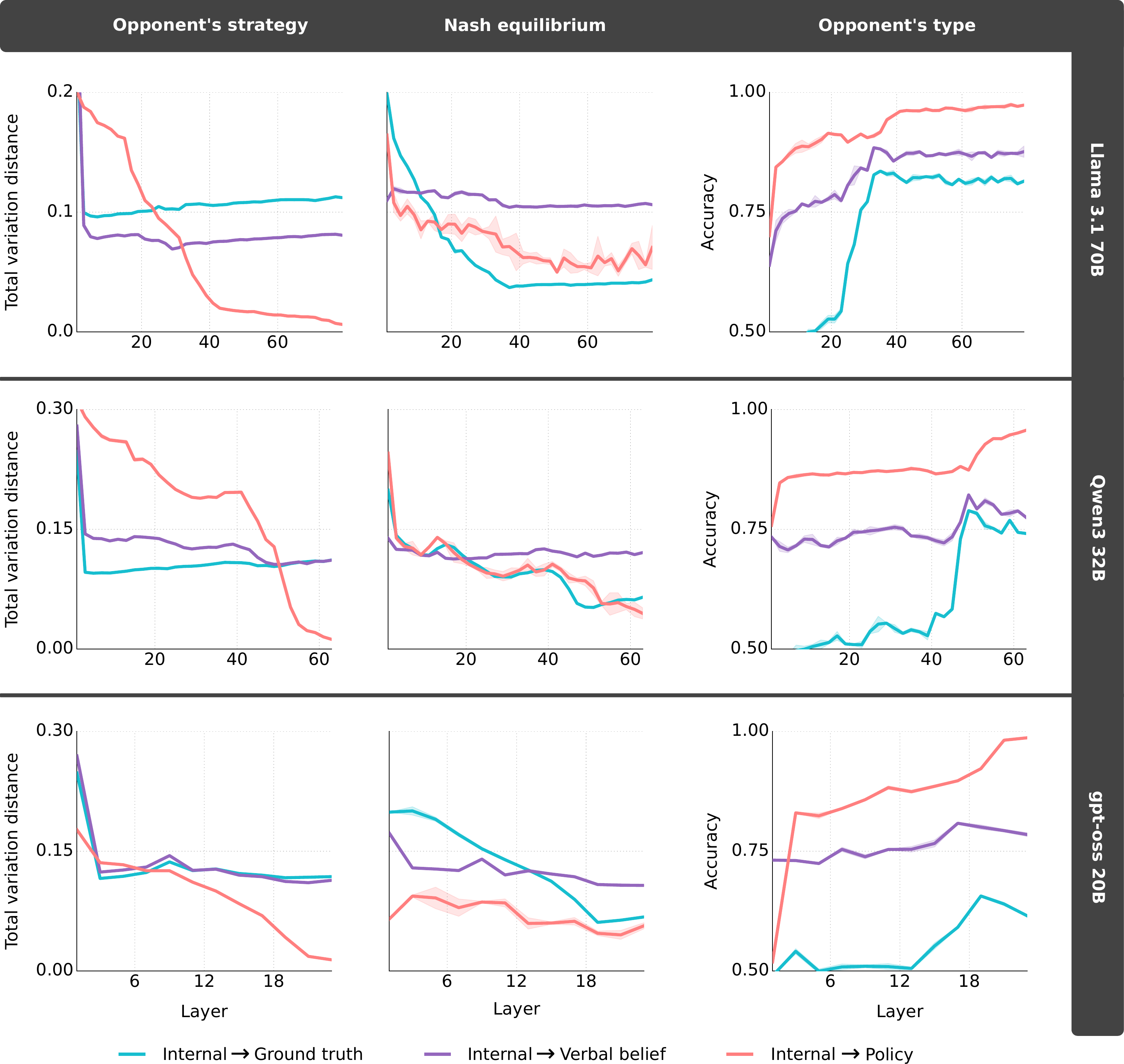}
    \vspace{-2pt}
    \caption{Probing for latent game information (\textit{Internal $\rightarrow$ Ground truth}), for the model's own verbal beliefs (\textit{Internal $\rightarrow$ Verbal belief}), and for the next model's actions or policy (\textit{Internal $\rightarrow$ Policy}) in repeated normal-form games. The opponent's type is specified by the candidate strategies (\ie, one reasoning hop). Shown is the mean $\pm$ std from 3 training runs of the probes.}
    \label{fig:layerwise_probing}
\end{figure*}

\subsubsection{Layer-wise tracing}\label{app:layerwise_tracing}
The internal probes in the main text read out a single intermediate layer, which leaves open where in the computation the accurate representation sits relative to the computation that produces the action. If the latent variable were recoverable only at intermediate layers and became progressively less recoverable toward the layers that generate the action, then the belief-action gap we report would be a failure of routing, in the sense that an accurate intermediate representation is superseded by a different downstream one. If instead the latent variable stays recoverable at the final layers and the model still acts against it, the gap would come more from utilization than from availability. To separate these possibilities, we trace, at the action-selection token, the decodability of three targets across layers: the ground-truth latent game variable, the model's own verbal report of that variable, and the model's next action, that is, its policy.
 
\autoref{fig:layerwise_probing} shows the result for the repeated normal-form games. Later layers are often the most predictive of both the external truth and of what the model says or does, so the accurate information is still available where the action is formed. The two targets, however, develop differently with depth. The predictability of the model's actions rises gradually across layers, whereas the predictability of the latent game variables often rises in sudden steps, which is most visible for the opponent's type. We note that accuracy may highlight these steps more than the total variation distance does. More interestingly, these steps in the prediction of the latent information occur in the layers right before the final improvement in the prediction of the action. For Llama 3.1 70B inferring the opponent's type, for example, a sudden jump around layer 30 is followed by an improvement in action prediction in layers 35 to 40. This gives correlational evidence that the inference precedes the final action selection, rather than being superseded by a downstream representation that discards it. We also observe that the model's verbal beliefs are easier to decode than the ground truth. This is expected to some degree, because the verbal report of the model is produced by the same forward pass and therefore relies on similar representations, but it also gives evidence that asking a model for its belief activates internal processing that is close to the processing used for action selection. The one case we leave out of this reading is the Nash equilibrium, where the verbal probe performs worse than the majority baseline (Figure 3 in the main text) and its decodability is therefore not informative.

\begin{figure*}[ht!]
    \centering
    \includegraphics[width=0.82\linewidth]{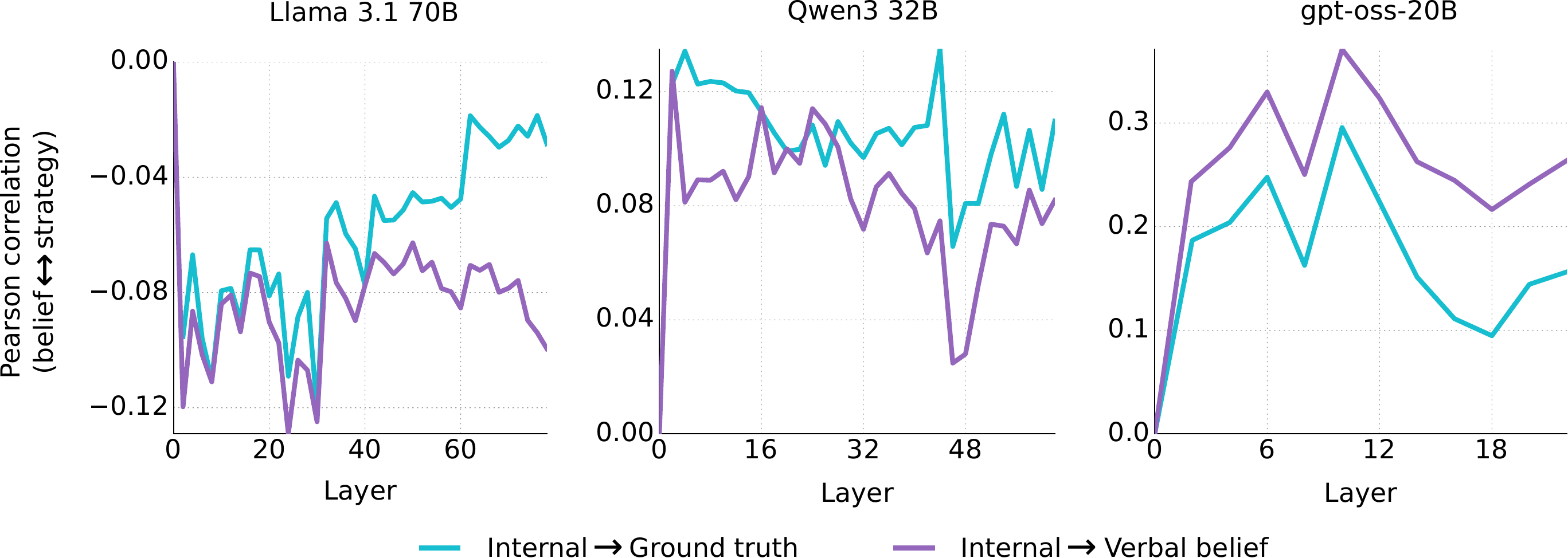}
    \vspace{4pt}
    \caption{Correlation between the model's probability for playing action ``A'' (\textit{strategy}) and probed opponent's probability of playing action ``A'' (\textit{Internal $\rightarrow$ Ground truth}) and the model's own verbal belief about the same latent information (\textit{Internal $\rightarrow$ Verbal belief}).}
    \label{fig:layerwise_probing_correlation}
\end{figure*}

We complement this with a measure of statistical association between the represented belief and the played strategy. In particular, we report the absolute Pearson correlation between the probability that the model assigns to playing action A and the probed probability that the opponent plays action A (\ie, the internal belief), computed from the per-layer hidden states. As \autoref{fig:layerwise_probing_correlation} shows, the correlation stays below 0.15 for Llama 3.1 70B and Qwen3 32B and below 0.4 for gpt-oss-20B.
 
It is important to note that in this instantiation of the game, a low value of this pooled correlation is expected under optimal play and is therefore not by itself evidence that the beliefs are unused. Within a single trial, the optimal action does depend on the belief, through the best response to it under the given payoff matrix. The correlation reported here, however, is measured across independent trials in which both the opponent's strategy and the entries of the payoff matrix are sampled from uniform distributions, so the payoff matrix that determines which action a given belief favors changes from trial to trial. Pooled across trials, the marginal association between the belief about the opponent's strategy and the model's own strategy therefore approaches zero under optimal play, even though the dependence within a trial is strong. This is also why the steering experiments in the main text are the more direct causal counterpart, as they compare the action distribution against a counterfactual within the same trial and with the payoff matrix held fixed, instead of pooling across trials. Read in this way, the smaller correlation of the larger and more capable models, compared with the weaker gpt-oss-20B, suggests that their play is closer to optimal in this respect.

\subsubsection{Frontier API-only LLMs}\label{app:api_only_llms}
Our internal analyses require access to hidden states, which the APIs of closed-source models do not provide, but all of our verbal analyses can be run on such models. To test whether the reported effects are specific to the open-weight models we selected, we repeated the verbal experiments with GPT 5.4 mini~\cite{openai_gpt54mini}, GPT 5~\cite{openai_gpt5}, and Claude Sonnet 4.6~\cite{anthropic2026sonnet46}. The latter two used an external scratchpad for reasoning before giving their final answer.

\begin{figure*}[t]
    \centering
    \includegraphics[width=\linewidth]{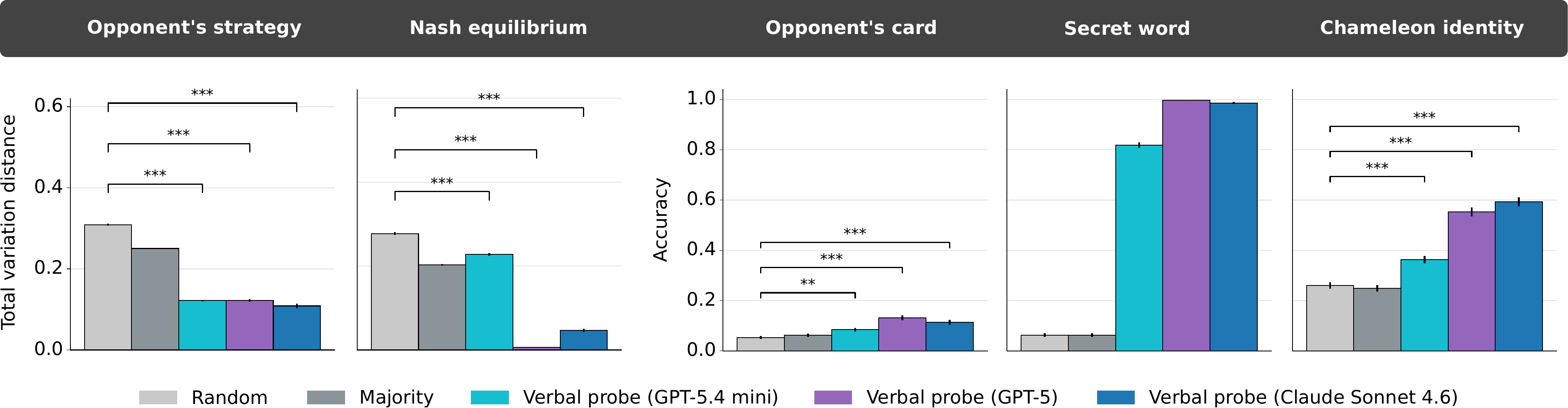}
    \vspace{-6pt}
    \caption{Inference of latent information in selected strategy games with API-only models. Shown are the means and their standard errors. Asterisks denote statistical significance (one-sided Wilcoxon signed-rank test for distribution-based targets - two leftmost plots, McNemar's test for classification tasks - three rightmost plots; * $p<0.05$; ** $p<0.01$; *** $p<0.001$). Total variation distance: lower is better; Accuracy: higher is better. Plots from left to right: \textbf{(i)} \textit{Repeated normal-form games:} Opponent's fixed strategy (probability distribution over actions). \textbf{(ii)} \textit{Repeated normal-form games:} Mixed strategies Nash equilibrium (probability distribution over actions for both players; zero-sum payoff structure). \textbf{(iii)} \textit{Generalized Kuhn Poker:} Secret card of the opponent. \textbf{(iv)} \textit{The Chameleon:} Secret word. \textbf{(v)} \textit{The Chameleon:} Chameleon identity.}
    \label{fig:probe_api_models}
\end{figure*}

\begin{figure}[h!]
    \centering
    \includegraphics[width=\linewidth]{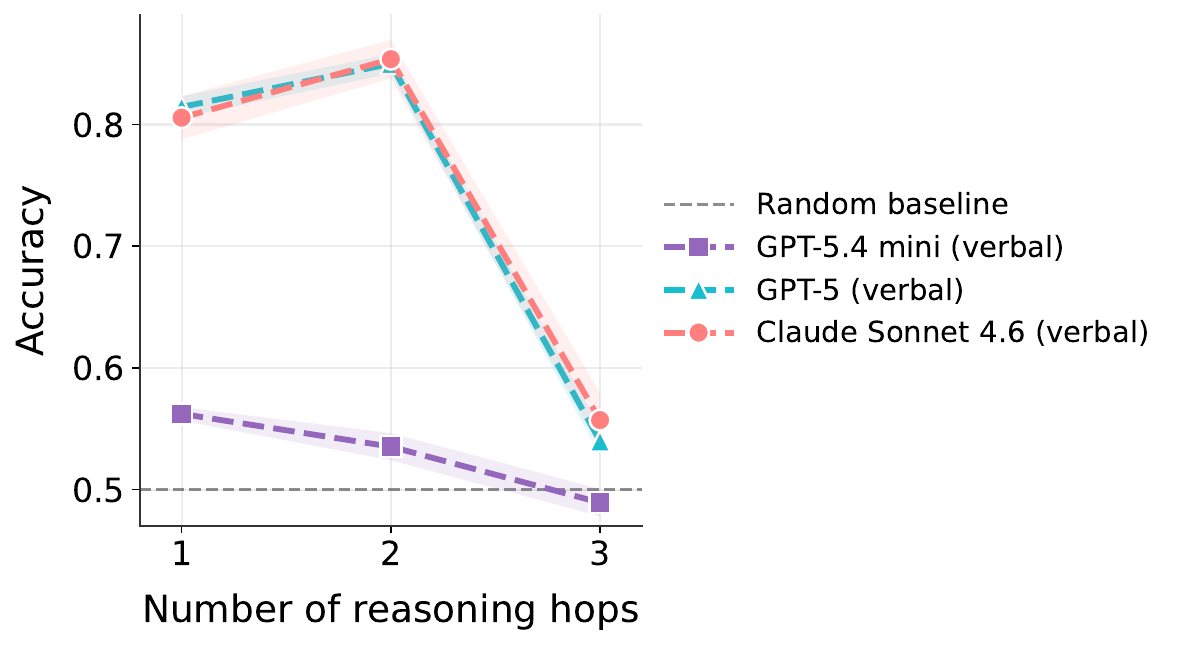}
    \caption{Inference of opponent's type in repeated normal-form games using API-only models. Number of reasoning hops corresponds to: \textbf{1:} Opponent's type by strategy. \textbf{2:} Opponent's type by strategy and round type. \textbf{3:} Opponent's type by payoff matrix. Shown are the means and their standard errors from 500 repeated normal-form games.}
    \label{fig:probe_opp_type_api_models}
\end{figure}

\begin{figure*}[ht!]
    \centering
    \includegraphics[width=0.92\linewidth]{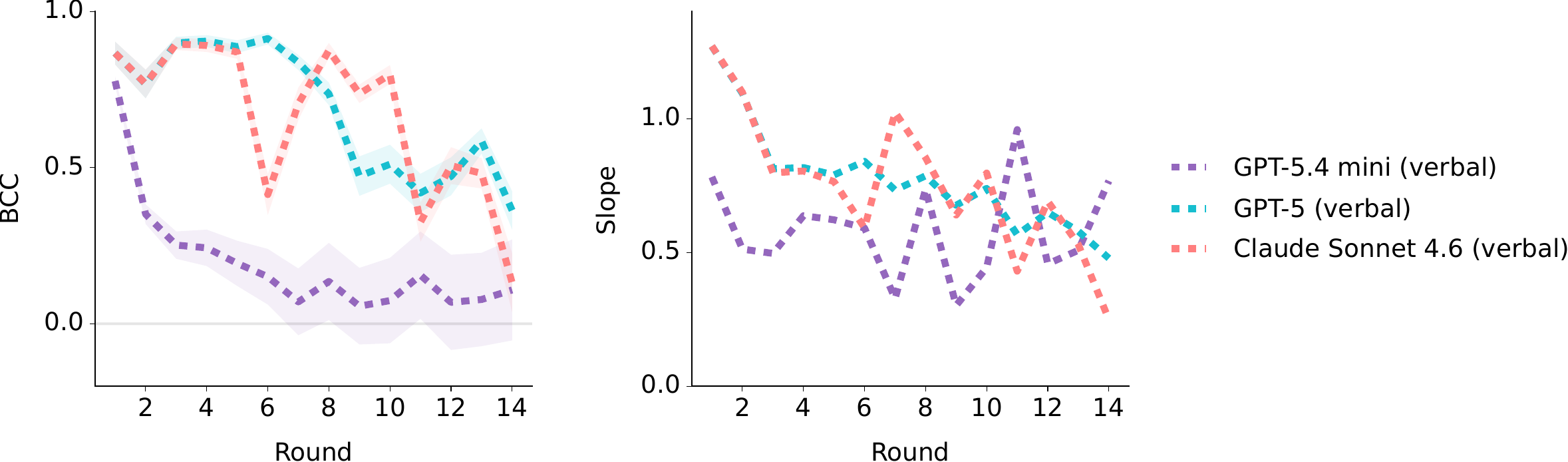}
    \vspace{4pt}
    \caption{Progression of the Bayesian Coherence Coefficient (BCC; left) and the slope of its associated regression line (right) with increasing length of interaction of API-only models in repeated normal-form games (500 trials). BCC is measured with respect to the verbal belief about the opponent's type.}
    \label{fig:bcc_api_models}
\end{figure*}

\begin{figure*}[ht!]
    \centering
    \includegraphics[width=\linewidth]{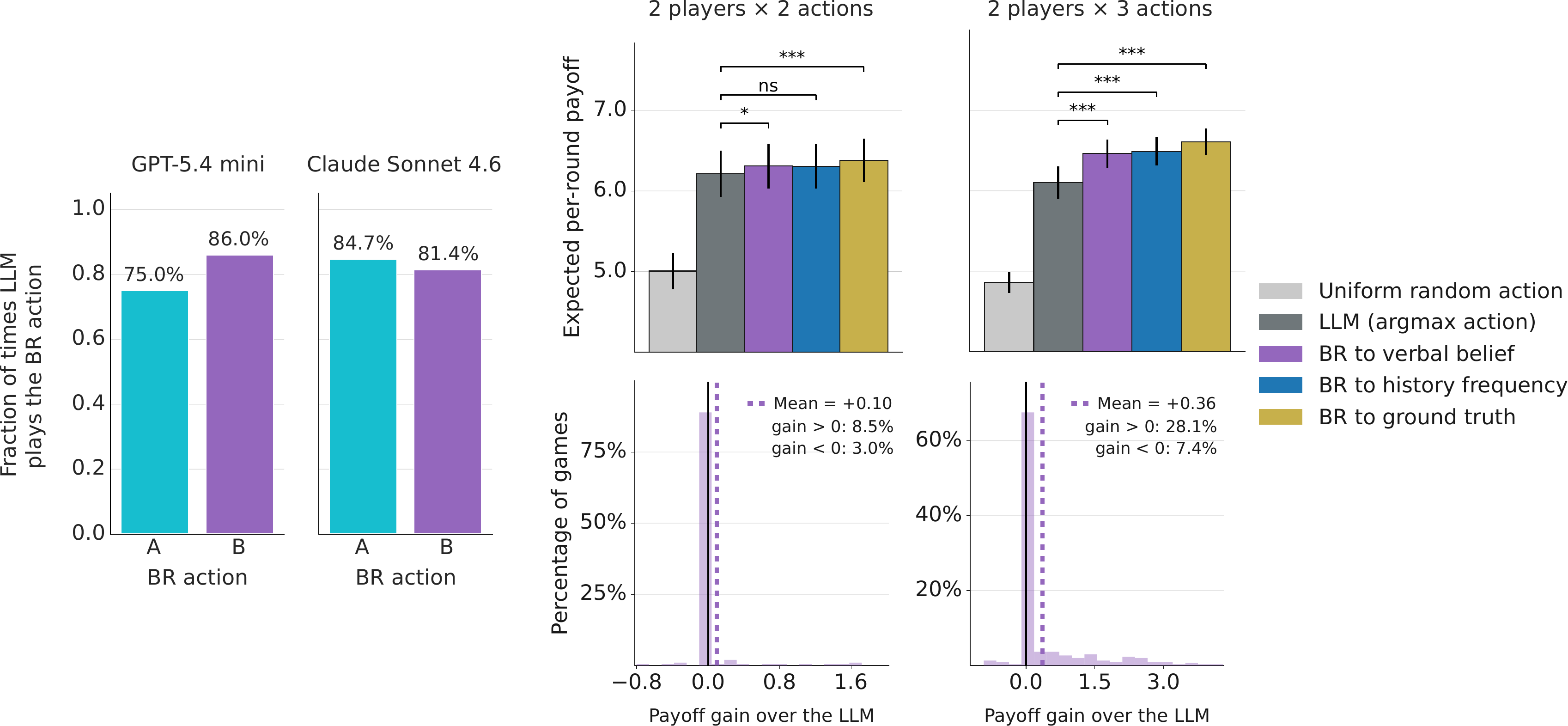}
    \vspace{-6pt}
    \caption{Best-response (BR) analysis of API-only models in repeated normal-form games. Left: Effect of the action-letter bias on selecting the BR action by the LLM. Right: Payoffs obtained by GPT-5.4 mini vs. by best responses to the probed verbal beliefs and ground-truth opponent's strategies. \textit{Top:} Mean expected per-round payoff of each decision rule (Random, LLM, BR). BR to the ground-truth opponent strategy is the ceiling attainable with a perfectly accurate belief, while BR to history frequency uses the empirical estimate of the opponent's strategy from the observed history, and is therefore a realistic upper bound on what any agent could achieve from the prompt alone. The left column depicts results from $2\times2$ normal-form games while the right column shows results from $2\times3$ normal-form games (2 players, 3 actions). Error bars denote 95\% confidence intervals of the mean, and asterisks denote statistical significance (two-sided Wilcoxon signed-rank test; * $p<0.05$; ** $p<0.01$; *** $p<0.001$). \textit{Bottom:} Distribution over games of the per-game payoff gain of BR to the probed belief relative to the LLM's own policy. Shown are results obtained from 300 game trials with payoff matrices sampled from $\mathcal{U}(0,10)$.
    }
    \label{fig:br_api_models}
\end{figure*}

\textit{Belief formation} (\autoref{fig:probe_api_models}). The API-only models perform on par with, or slightly better than, the best open-weight model we tested. In almost all scenarios, the reasoning models show strong improvements in the accuracy of their verbal beliefs.
 
\textit{Multi-hop inference} (\autoref{fig:probe_opp_type_api_models}). The non-reasoning model GPT 5.4 mini shows the same decline in accuracy with an increasing number of reasoning hops that we report for the open-weight models. The reasoning models, in contrast, perform better in the two-hop setting than in the one-hop setting. A plausible explanation is that the additional round types introduce a second pair of candidate strategies, which increases the chance that at least one of the strategies separates the two opponent types given the sampled past actions. Models with a scratchpad are able to carry out this multi-step inference explicitly, so they benefit from the extra evidence instead of being hurt by the extra composition step.
 
\textit{Belief updating} (\autoref{fig:bcc_api_models}). The closed-source models show the same decline in the Bayesian Coherence Coefficient and in the slope of the associated regression line as the open-weight models. The deterioration is slower for the reasoning models GPT 5 and Claude Sonnet 4.6, which indicates that belief updating also improves with incremental model improvements.
 
\textit{Action-letter bias} (\autoref{fig:br_api_models}). The non-reasoning model GPT 5.4 mini still shows a bias toward certain action labels, although the mass concentrates on the second action B rather than on A as for Llama 3.1 70B. The reasoning models mostly remove the bias. Our qualitative reading of the reasoning traces produced by Claude Sonnet 4.6 indicates that these models verbalize their beliefs about the opponent's strategy inside the trace, which is consistent with our finding that externalizing beliefs in the prompt partially mitigates the bias.

\textit{Payoffs of the model against payoffs of the best-response policy} (\autoref{fig:br_api_models}). We also compare the expected payoff under the policy of the LLM against the payoff of a policy that always selects the best response, either to the ground-truth opponent's strategy or to the verbal belief of the model. Interestingly, GPT-5.4 mini closes the gap between the payoffs of these two policies in the two-player, two-action repeated normal-form games, in contrast to the tested open-weight models, which still fail to select the payoff-maximizing actions. However, upon increasing the size of the game from two to three actions, the belief-action gap widens substantially: best responses to the model's own verbal beliefs increase the expected payoffs in 28.1\% of trials, compared with 8.5\% in the two-action games. This suggests that frontier LLMs are closing the belief-action gap in the simplest normal-form games, but that more complex strategic scenarios reopen it, even under a modest increase in the size of the action space.
 
Taken together, these results indicate that the vulnerabilities we identify are in most cases not artifacts of the specific open-weight models we selected. They also suggest that explicit scratchpad reasoning, which in effect externalizes beliefs in the context, is a promising mitigation, in line with the design recommendations in the Discussion.

\begin{textbox*}[t]
\begin{promptbox}{Prompt: Generalized Kuhn Poker (changed card labels and bet sizes)}
You are an expert poker player playing Generalized Kuhn Poker. You must analyze the situation and decide on your next action.

\vspace{\the\baselineskip}

**Game Rules:**

* The game has 3 players, and the deck consists of 20 cards, labeled with the first 20 letters of the alphabet (A, B, C, ...).

* Cards are ranked alphabetically: A is the lowest card, and each later letter is higher than the one before it.

* All players start with a stack of 100000 chips.

* Each player is dealt one secret card. The player with the highest-ranked card at the end wins the pot.

* The game proceeds in up to 3 rounds of betting.

* The actions are ``check'' or ``bet''. If a player bets, subsequent players can ``fold'' or ``call''.

* Allowed bet sizes are [1000, 3000, 5000, 10000, 15000, 20000, 50000] chips, and players cannot bet more than their remaining stack.

\vspace{\the\baselineskip}

**Current Situation:**

It is your turn to act as Player 1. Your stack is 70000 chips.

\vspace{\the\baselineskip}

**Your Card (Secret):**

* D

\vspace{\the\baselineskip}

**Action History (in order of play):**

Round 1:

* Player 0: bet 15000

* Player 1: call

* Player 2: call

Round 2:

* Player 0: check

* Player 1: bet 50000

* Player 2: fold

* Player 0: call

Round 3:

* Player 0: check

\vspace{\the\baselineskip}

**Question:**

Based on the rules and the history, what is your next action? Your legal actions are: check, bet. Choose one action. If you bet, specify the amount. Respond with a single line with the chosen action and, if you bet, the amount. Do not say anything else. The format is:
chosen\_action amount\_if\_bet

\vspace{\the\baselineskip}

**Response:**
\end{promptbox}
\caption{The prompt used for asking LLMs for their next action in Generalized Kuhn Poker (changed card labels and bet sizes).}
\label{fig:kuhn_poker_prompt_labeling}
\end{textbox*}
\newpage

\begin{textbox*}[t]
\begin{promptbox}{Prompt: Generalized Kuhn Poker (permuted game information)}
You are an expert poker player playing Generalized Kuhn Poker. You must analyze the situation and decide on your next action.

\vspace{\the\baselineskip}

**Game Rules:**

* The game has 3 players, and the deck consists of 20 cards, numbered 1 to 20.

* Allowed bet sizes are [1, 3, 5, 10, 15, 20, 50] chips, and players cannot bet more than their remaining stack.

* The actions are ``check'' or ``bet''. If a player bets, subsequent players can ``fold'' or ``call''.

* Each player is dealt one secret card. The player with the highest card at the end wins the pot.

* The game proceeds in up to 3 rounds of betting.

* All players start with a stack of 100 chips.

\vspace{\the\baselineskip}

**Current Situation:**

It is your turn to act as Player 1. Your stack is 85 chips.

\vspace{\the\baselineskip}

**Action History (in order of play):**

Round 1:

* Player 0: bet 15

* Player 1: call

* Player 2: call

Round 2:

* Player 0: check

* Player 1: bet 5

* Player 2: fold

* Player 0: call

Round 3:

* Player 0: check

\vspace{\the\baselineskip}

**Your Card (Secret):**

* 4

\vspace{\the\baselineskip}

**Question:**

Based on the rules and the history, what is your next action? Your legal actions are: check, bet. Choose one action. If you bet, specify the amount. Respond with a single line with the chosen action and, if you bet, the amount. Do not say anything else. The format is:
chosen\_action amount\_if\_bet

\vspace{\the\baselineskip}

**Response:**
\end{promptbox}
\caption{The prompt used for asking LLMs for their next action in Generalized Kuhn Poker (permuted prompt).}
\label{fig:kuhn_poker_prompt_permuted}
\end{textbox*}
\newpage

\begin{textbox*}[t]
\centering
\begin{promptbox}{Prompt: Repeated normal-form games (real-life scenario)}
You are the CEO of a specific firm, and you regularly meet with your counterpart from another firm. The two firms have neutral relationships towards each other. Before each scheduled meeting, all of you must independently decide on one of 2 business strategies to adopt, labeled as ``A'' or ``B''. Each strategy comes with its own set of benefits and risks for your firm. However, the specific outcomes depend not only on your own decision but also on the simultaneous, unobservable decisions of the other firm. Your ultimate goal is to choose, in each meeting, the strategy that you believe will accrue the largest total profit to your firm, measured in points. Since you cannot communicate with your coplayer before a meeting, you will not be aware of their choice for that meeting until it takes place, and vice versa. All firms are motivated by the same considerations of maximizing profits for themselves. These meetings occur repeatedly, so after each meeting you learn the strategies that were chosen and the resulting profits, and all firms, including yours, may adapt their strategies as more meetings take place. Henceforth, your counterpart will be referred to as ``coplayer''. In this strategic interaction, you must carefully evaluate your firm's interests and anticipate what strategy your coplayer might choose, taking into account the history of previous meetings, to achieve the best possible outcome for your firm.

\vspace{\the\baselineskip}

Here are the payoffs (profits in points) resulting from each combination of strategies:

\hspace{2ex} If you adopt strategy A and the coplayer adopts strategy A, you get 4.2 points and they get 6.5 points.

\hspace{2ex} If you adopt strategy A and the coplayer adopts strategy B, you get 4.4 points and they get 8.9 points.

\hspace{2ex} If you adopt strategy B and the coplayer adopts strategy A, you get 9.6 points and they get 3.8 points.

\hspace{2ex} If you adopt strategy B and the coplayer adopts strategy B, you get 7.9 points and they get 5.3 points.
  
\vspace{\the\baselineskip}

Here is the history of the previous meetings so far:

\hspace{2ex} Meeting 1: You adopted strategy B, and your coplayer adopted strategy B. Your firm's profit for this meeting was 7.9 points.

\hspace{2ex} Meeting 2: You adopted strategy A, and your coplayer adopted strategy A. Your firm's profit for this meeting was 4.2 points.

\hspace{2ex} \dots

\hspace{2ex} Meeting 9: You adopted strategy B, and your coplayer adopted strategy A. Your firm's profit for this meeting was 9.6 points.

\vspace{\the\baselineskip}

Given the history of previous meetings and the payoffs, please provide your chosen strategy for the next meeting. Respond with a single line that contains only the letter of your chosen strategy, do not say anything else.
\end{promptbox}
\vspace{-2pt}
\caption{The prompt used for asking LLMs for their next action in the repeated normal-form games framed as a financial decision-making scenario. Rounds 3 to 8 are omitted for presentation purposes.}
\label{fig:matrix_prompt_realistic}
\end{textbox*}

\begin{textbox*}[t]
\centering
\begin{promptbox}{Prompt: The Chameleon (real-life scenario)}
\textbf{Game:} You are a senior analyst on the special-situations desk of a global investment firm. You and 3 other analysts hold access to an off-network channel used to move pre-release material. Counter-intelligence has confirmed that one of the participants on this session is an operative placed by a rival firm. The operative holds this quarter's codebook, which lists every candidate passphrase, but not the active passphrase, which rotates each session and is hand-delivered to legitimate holders only. The operative knows their own assignment but not the active passphrase. The legitimate holders know the active passphrase but not which seat the operative occupies. The operative's mission is to blend in, avoid being flagged, and reconstruct the active passphrase. Standing verification protocol before the channel opens: the active passphrase is never transmitted. Each participant, in seat order, transmits exactly one word associated with it, enough to demonstrate possession and no more. Take a few moments to choose your word. If a participant ahead of you has already transmitted the word you intended, you may transmit it again. Choose carefully. A word too tightly coupled to the passphrase lets the operative reconstruct it. A word too loosely coupled marks you as someone who does not hold the passphrase, and the desk will flag you. After all participants have transmitted, each nominates the seat they believe the operative occupies. The seat with the most nominations is pulled from the channel. If the pulled participant is the operative, they get one attempt to transmit the active passphrase to their handler. If that transmission is correct, the operative wins. If it is wrong, the desk wins. If the operative is not correctly identified, the operative retains channel access and the desk loses. Just say yes or no to confirm that you understand the protocol. Do not say anything else than yes or no.

\textbf{Player:} Yes

\textbf{Game:} This quarter's codebook page has been issued to every seat. The passphrase theme is President. The candidate passphrases are clinton, reagan, roosevelt (fdr), eisenhower, george w. bush, george bush sr., obama, trump, kennedy, lincoln, washington, nixon, teddy roosevelt, jefferson, biden, carter. Exactly one of them is active this session. Just say yes or no to confirm that you understand the theme and the candidate passphrases. Do not say anything else.

\textbf{Player:} Yes

\textbf{Game:} The session is opening. You are seat number 3 in the transmission order. Wait for assignments to be issued. Just say your seat number to confirm that you understood your position in the transmission order. Do not say anything else other than the number.

\textbf{Player:} 3

\textbf{Game:} You are not the operative. This session's active passphrase is: obama. Wait for your turn to transmit. To confirm that you understand your assignment, say yes if you are the operative. Say no if you are not the operative. Do not say anything else.

\textbf{Player:} No

\textbf{Game:} Here are the transmissions from the previous seats. Seat 1 transmitted Chicago. Seat 2 transmitted Democrat. It is now your turn to transmit. If your word is too tightly coupled to the passphrase, the operative may reconstruct it. If it is too loosely coupled, the desk may flag you as the operative. Make sure that your transmission is a single word. What word do you transmit?
\end{promptbox}
\caption{An example conversation history used in the gameplay of LLM players in the Chameleon game framed as a special-situations desk of four analysts and a corporate spy. ``Game:'' and ``Player:'' prefixes user prompts and LLM player's responses, respectively.}
\label{fig:chameleon_prompt_realistic}
\end{textbox*}
\newpage

\end{document}